\documentclass[11pt]{imsart}
\usepackage{times}

\usepackage[round,colon,authoryear]{natbib}
\usepackage{longtable}

\RequirePackage[OT1]{fontenc}
\RequirePackage{amsthm,amsmath,amsfonts,amssymb}

\usepackage{color}

\RequirePackage{latexsym}
\RequirePackage{psfrag}
\usepackage{fullpage}

\usepackage[caption=false]{subfig}
\usepackage{graphics,graphicx}
\usepackage{enumerate}
\usepackage{bbm} 
\usepackage{booktabs}
\usepackage{multirow}
\usepackage{siunitx} 
\usepackage{float}
\usepackage[dvipsnames]{xcolor}

\theoremstyle{plain}

\newtheorem{theo}{Theorem}[section]

\newtheorem{lem}{Lemma}[section]
\newtheorem{prop}{Proposition}[section]
\newtheorem{cor}{Corollary}[section]

\theoremstyle{definition} 

\newtheorem{nota}{Notation}[section]
\newtheorem{de}{Definition}[section]
\newtheorem{exa}{Example}[section]
\newtheorem{as}{Assumption}[section]
\newtheorem{alg}{Algorithm}[section]
\newtheorem{rem}{Remark}[section]

\newcommand{\btheo}{\begin{theo}}
\newcommand{\bde}{\begin{de}}
\newcommand{\ble}{\begin{lem}}
\newcommand{\bpr}{\begin{prop}}
\newcommand{\bno}{\begin{nota}}
\newcommand{\bex}{\begin{exa}}
\newcommand{\bcor}{\begin{cor}}
\newcommand{\spro}{\begin{proof}}
\newcommand{\bas}{\begin{as}}
\newcommand{\balg}{\begin{alg}}
\newcommand{\brem}{\begin{rem}}

\newcommand{\etheo}{\end{theo}}
\newcommand{\ede}{\end{de}}
\newcommand{\ele}{\end{lem}}
\newcommand{\epr}{\end{prop}}
\newcommand{\eno}{\end{nota}}
\newcommand{\eex}{\end{exa}}
\newcommand{\ecor}{\end{cor}}
\newcommand{\fpro}{\end{proof}}
\newcommand{\eas}{\end{as}}
\newcommand{\ealg}{\end{alg}}
\newcommand{\erem}{\end{rem}}

\theoremstyle{plain}

\newtheorem{theos}{Theorem}
\newtheorem{props}{Proposition}
\newtheorem{lems}{Lemma}
\newtheorem{cors}{Corollary}

\theoremstyle{definition}
\newtheorem{exas}{Example}
\newtheorem{algs}{Algorithm}
\newtheorem{asss}{Asumption}
\newtheorem{defns}{Definition}

\newcommand{\btheos}{\begin{theos}}
\newcommand{\etheos}{\end{theos}}
\newcommand{\bprops}{\begin{props}}
\newcommand{\eprops}{\end{props}}
\newcommand{\bdes}{\begin{defns}}
\newcommand{\edes}{\end{defns}}
\newcommand{\blems}{\begin{lems}}
\newcommand{\elems}{\end{lems}}
\newcommand{\bcors}{\begin{cors}}
\newcommand{\ecors}{\end{cors}}
\newcommand{\bexs}{\begin{exas}}
\newcommand{\eexs}{\end{exas}}
\newcommand{\balgs}{\begin{algs}}
\newcommand{\ealgs}{\end{algs}}
\newcommand{\bass}{\begin{asss}}
\newcommand{\eass}{\end{asss}}

\def\cN{\mathcal{N}}

\def\reals{\mathbb{R}}

\def\prob{\mathbb{P}}
\def\expect{\mathbb{E}}

\def\zero{\mathbf{0}}
\def\hSigma{\hat \Sigma}

\def\inci{\Gamma}
\def\Pen{\Gamma}
\def\hbeta{\hat \beta}
\def\btrue{\beta^\ast}
\def\bstar{\btrue}

\def\argmin{\mathop{\!\arg \min}}
\def\ltv{\lambda_{\rm TV}}
\def\lrid{\lambda_S}
\def\lone{\lambda_1}

\def\tX{\tilde X}
\def\ty{\tilde y}
\def\tPen{\tilde \Pen}
\def\ptPen{\tilde \Pen^\dagger}
\def\ie{{\em i.e.},~}
\def\eg{{\em e.g.},~}

\DeclareMathOperator*{\supp}{supp}
\DeclareMathOperator*{\sign}{sign}

\def\ith{$i^{\rm th}$~}

\def\ones{\mathbbm{1}}

\usepackage{hyperref}
\definecolor{orcidlogocol}{HTML}{A6CE39}

\begin{document}

\begin{frontmatter}
\title{Graph-based regularization for regression problems with alignment and highly-correlated designs}
\runtitle{Graph-based regularization for  highly-correlated designs}

\begin{aug}
\thankstext{t0}{These authors contributed equally to the manuscript.}
\thankstext{t1}{Department of Statistics, University of Wisconsin-Madison. Raskutti and Li were supported by NSF DMS 1407028. Song was supported by NIH R01 GM131381-01.}
\thankstext{t2}{Department of Mathematics, University of Wisconsin-Madison. Mark was supported by NSF Awards 0353079 and 1447449.}
\thankstext{t3}{Departments of Statistics and Computer Science, University of Chicago.  Willett was supported by NIH Award 1 U54 AI117924-01 and NSF Awards 0353079, 1447449, 1740707, and 1839338.}
\author{\fnms{Yuan} \snm{Li}\thanksref{t0}\thanksref{t1}\ead[label=e1]{first@somewhere.com}},
\author{\fnms{Benjamin} \snm{Mark}\thanksref{t0}\thanksref{t2}\ead[label=e2]{bmark2@wisc.edu}},
\author{\fnms{Garvesh} \snm{Raskutti}\thanksref{t1}\ead[label=e3]{garvesh@gmail.com}}
,
\author{\fnms{Rebecca} \snm{Willett}\thanksref{t3}\ead[label=e4]{willett@uchicago.edu}}
,
\author{\fnms{Hyebin} \snm{Song}\thanksref{t1}\ead[label=e5]{hb.song@wisc.edu}}
\and
\author{\fnms{David} \snm{Neiman}\thanksref{t1}\ead[label=e6]{dneiman@wisc.edu}}

\runauthor{Li, Mark, Raskutti,Willett, Song, & Neiman}

\end{aug}

\begin{abstract}
Sparse models for high-dimensional linear regression and machine learning have received substantial attention over the past two decades. Model selection, or determining which
features or covariates are the best explanatory variables, is critical to the
interpretability of a learned model. Much of the current literature assumes that
covariates are only mildly correlated. However, in many modern applications covariates are highly correlated and do not exhibit key properties (such as the restricted eigenvalue condition, restricted isometry property, or other related assumptions). 
This work considers a high-dimensional regression setting in which a graph governs both correlations among the covariates and the similarity among regression coefficients
-- meaning there is \emph{alignment} between the covariates and regression coefficients. Using side information about the strength of correlations
among features, we form a graph with edge weights corresponding to
pairwise covariances. This graph is used to define a graph total
variation regularizer that promotes similar weights for 
correlated features.

This work shows how the proposed graph-based regularization yields mean-squared error guarantees for a broad range of covariance graph structures.  These guarantees are optimal for many specific covariance graphs, including block and lattice graphs.  Our proposed approach outperforms other methods for highly-correlated design in a variety of experiments on synthetic data and real biochemistry data.

\end{abstract}

\end{frontmatter}

\section{Introduction}
\label{sc:intro}

High-dimensional linear regression and inverse problems have received
substantial attention over the past two decades (see \citet{hastie2015statistical} for an overview). While there has been considerable theoretical and methodological development, applying these methods in real-world settings is more nuanced since variables or features are often highly correlated, while much of the existing theory and methodology is applicable when features are independent or satisfy weak correlation assumptions such as
the
restricted eigenvalue and other related conditions (see \citet{CandesTao06,BiRiTsy08,GeerBuhl09}). In this paper we develop an approach for parameter estimation in high-dimensional linear regression with highly-correlated designs.

More specifically, we consider observations of the form
\begin{equation}
\label{eq:obs}
y = X\bstar + \epsilon
\end{equation}
where $y \in \reals^n$ is the response variable,
$X \in \reals^{n\times p}$ is the observation or {\em design} matrix,
and $\epsilon\sim \mathcal{N}(\textbf{0},\sigma^2I_{n\times n})$ is Gaussian
noise. Our goal is to estimate $\btrue$ based on $(X,y)$ when $X$ potentially has highly-correlated columns and does not satisfy standard regularity assumptions. Specifically, we define $\Sigma := \frac{1}{n}\expect[X^\top X]$ and consider settings where the minimum eigenvalue of $\Sigma$ may be zero-valued or arbitrarily close to zero.  We consider a Gaussian linear model for simplicity of exposition but our ideas and results can be extended to other settings.  In Appendix~\ref{sec:logistic} we discuss an extension to logistic regression.

Highly-correlated or dependent features arise in many modern scientific problems, including the study of enzyme thermostability (detailed in Section~\ref{sec:applications}), genome wide association studies (GWAS)~\citep{wu2009genome,gene_regulatory}, neuroscience \citep{genfused18}, climate data \citep{barnston1996specification,geisler1985sensitivity,delsole2017statistical,mamalakis2018new}, and topic modeling.

As we discuss and expand upon in Section~\ref{sec:disc}, there is a large body of work addressing the problem of high-dimensional regression under highly correlated design 
(\eg \cite{buhlmann2013correlated,Zou05}). The key challenge associated with highly-correlated columns is that estimates of $\btrue$ become very sensitive to noise and, if columns are perfectly correlated, $\btrue$ may not be identifiable, which means additional assumptions are required on $\btrue$.

On the other hand, for many applications such as those mentioned above, there is known structure among $\btrue$ since groups of covariates often exhibit similar influence on the response. There is also a large body of work studying the high-dimensional linear model under additional assumptions on $\btrue$ including group structure (\eg \cite{shen2010grouping,ZhaoRochaYu}), graph structure (\eg \cite{sharpnack2012sparsistency,hallac2015network,MarialYu,wang2016trend}), and others. 

In this work, we consider a case of highly correlated designs with additional structure on $\btrue$.  We use side information to generate a covariance graph and then use an \emph{alignment} condition to ensure a corresponding graph structure on $\bstar$. The alignment condition resolves the lack of identifiability by incorporating side information about the covariance. Importantly, we develop novel theoretical guarantees for our procedure under this alignment condition.

\subsection{Motivating application: Biochemistry}
\label{sec:applications}

In this section we apply the proposed GTV methodology to an application in biochemistry, specifically protein analysis. In particular we focus on a specific protein of great interest, the cytochrome P450 enzyme, which is an important protein in a number of environments. More specifically, cytochrome P450 proteins are versatile biocatalysts which have been heavily employed for production of pharmaceutical products and synthesis of other useful compounds \citep{Guengerich2002-fz}. Additionally, thermostable proteins have great industrial importance since they can withstand tough industrial process conditions \citep{Niehaus1999-un}. We aim to understand how 3-D structural properties of proteins are related to the thermostability of the proteins.

The dataset we use is a P450 chimeric protein dataset generated by the Romero Lab at UW-Madison\footnote{Raw data is available at \mbox{https://github.com/Jerry-Duan/Structural-features}}. The dataset contains thermostability measurements and features encoding the amino acid sequences and describing structural properties of $242$ chimeric P450 proteins. The chimeric proteins in the dataset are created by recombining fragments of the genes of the three wild-type P450s (parent proteins) for eight blocks ~\citep{Li2007-qo}. Since the amino acid sequences for the parent proteins are known, the amino acid sequence for a chimeric protein can be obtained from the recombination information for each block which parent the gene fragment is inherited from. From the amino acid sequence information, $50$ features describing the structural properties of each protein were estimated by modeling 3-D structures of the chimeric enzymes via the Rosetta biomolecular modeling suite \citep{Alford2017-pt}. A full description of the $50$ structural features is provided in Table \ref{tab:biochem_table} in the Appendix. As our goal is to understand the relationship between the structure and thermostability of the proteins, we use a linear model where the design matrix ${X} \in \mathbb{R}^{n \times p}$ consists of the structure features and the response variable ${y} \in \mathbb{R}^{n}$ contains the thermostability measurements for $n=242$ and $p=50$. 

Importantly, many of the structural features are known to be highly correlated and we use side information to estimate the covariance structure between the structural features. 
The side information consists of the amino acid sequences for the P450 chimeric proteins. We use the sequence, structure, and function paradigm for protein design in which a protein sequence determines the structure of the protein and the structure determines the function of the protein. In particular, we exploit the sequence-structure relationship to obtain a good estimate of the covariance matrix of the structural features.
The combination of highly correlated features and side information to estimate the covariance matrix makes this problem a natural fit for out GTV methodology. More details on the estimation of the covariance and the application are provided in Section~\ref{sec:biochem}.

\subsection{Problem formulation and proposed estimator}
\label{sec:estimator}

First we define our model based on the standard linear model where data $(X^{(i)}, y^{(i)})_{i=1}^n \in \mathbb{R}^p \times \mathbb{R}$ are drawn i.i.d.~according to
\begin{eqnarray*}
  y^{(i)}={X^{(i)}}^{\top}\btrue+\epsilon^{(i)}, \mbox{ where } X^{(i)}\sim \cN(\textbf{0},\Sigma_{p\times p}) \mbox{ and } \epsilon^{(i)}\sim \cN(0, \sigma^2).
\end{eqnarray*}
Let $y = (y^{(1)}, y^{(2)},...,y^{(n)})^{\top} \in \mathbb{R}^n$, $X = [X^{(1)}, X^{(2)},...,X^{(n)}]^{\top} \in \mathbb{R}^{n \times p}$ and $\epsilon = (\epsilon^{(1)}, \epsilon^{(2)},...,\epsilon^{(n)})^{\top}\in \mathbb{R}^n$. Hence the linear model can be expressed in the standard matrix-vector form:
\begin{eqnarray*}
  y = X \btrue+\epsilon.
\end{eqnarray*}
Our goal is to estimates $\btrue$.  We are particularly interested in a setting where the columns of $X$ may be highly correlated (\ie $\lambda_{\min}(\Sigma) \approx 0$), but $\btrue$ is well-aligned with the covariance structure (\ie correlated features have similar weights in $\btrue$).

We assume $\Sigma$ is unknown and is estimated using either $X$ or side information; let $\hSigma$ denote the estimate of the covariance matrix. Define $\hat s_{j,k} := \sign(\hSigma_{j,k})$.  Based on the estimated covariance matrix $\hSigma$, we consider the following
estimator for $\bstar$:
\begin{align}
\hat \beta = &~\argmin_\beta \frac{1}{n} \|y-X\beta\|_2^2 + \lrid
\sum_{j,k} |\hSigma_{j,k}| (\beta_j - \hat s_{j,k} \beta_k)^2\nonumber
\\
& \qquad + \lone (\ltv
\sum_{j,k} |\hSigma_{j,k}|^{1/2} |\beta_j - \hat s_{j,k} \beta_k|+\|\beta\|_1),\label{eq:est1}
\end{align}
where $\lrid,~\lone$ and $\ltv$ are regularization parameters.  

This estimator can be interpreted from a graph/network perspective by defining the \emph{covariance graph} based on the covariance matrix $\hat \Sigma$. Let $G = (V,E,W)$ be an undirected weighted graph where $V = \{1,2,...,p\}$ with edge weight
$w_{j,k}~(1\leq j\neq k\leq p)$ associated with edge $(j,k)\in E$. The edge weights corresponding to $W = (w_{j,k})$ may be negative. Now we define our covariance graph. Let $w_{j,k} = \hSigma_{j,k}$, which denotes the $(j,k)$ entry of the covariance matrix $\hSigma$. Then $E := \{(j,k)\;:\;w_{j,k} \neq 0,~ j \neq k\}$ and the entries of the weight matrix $W \in \mathbb{R}^{p \times p}$ are $W_{j,k} = w_{j,k}$. Given this graph, the regularization term $\sum_{j,k} |\hSigma_{j,k}|^{1/2} |\beta_j - \hat s_{j,k} \beta_k|$ is a measure of the {\em graph total variation} of the signal $\beta$ with respect to the graph $G$ and $\sum_{j,k} |\hSigma_{j,k}| (\beta_j - \hat s_{j,k} \beta_k)^2$ corresponds to a {\em graph Laplacian regularizer} with respect to $G$. 

Further let $\inci$ be the {\em weighted edge incidence matrix} associated with the graph $G$. Specifically, we denote the set of edges in our graph as $(j_\ell,k_\ell)$ for $\ell=1,\ldots,m$ where
$m := |E|$ is the size of the edge set.
Let 
\begin{eqnarray}
\label{eq:Gamma}
\inci = \sum_{\ell=1}^m \inci_\ell, \qquad \mbox{where} \qquad \inci_\ell := |\hSigma_{j_\ell,k_\ell}|^{1/2}u_\ell \left[e_{j_\ell} - \sign(\hSigma_{j_\ell,k_\ell})
  e_{k_\ell}\right]^\top \in \reals^{m \times p},
\end{eqnarray}
where $u_\ell\in \reals^m$ and $e_\ell \in \reals^p$ are the $\ell^{\text{th}}$ canonical basis vectors (all zeros except for a one in the $\ell^{\text{th}}$ element). 
Then the $\ell^{\rm th}$ row of
$\inci$ is
$$|\hSigma_{j_\ell,k_\ell}|^{1/2} \left[e_{j_\ell} -
  \sign(\hSigma_{j_\ell,k_\ell})e_{k_\ell}\right]^\top.$$

Next suppose $\lone>0$ and $\lambda_{TV},\lambda_S\geq0$. We define
$$\tX = \tX_{\lrid} := \begin{bmatrix} X \\ \sqrt{n\lrid}\Pen \end{bmatrix}\in \mathbb{R}^{(n+m)\times p},~\ty := \begin{bmatrix} y \\ \zero_{m\times1} \end{bmatrix}\in \mathbb{R}^{n+m}, \mbox{ and } \tPen:= \begin{bmatrix} \ltv\Pen \\
 I_{p\times p} \end{bmatrix}\in \mathbb{R}^{(m+p)\times p}.$$

Using these definitions, we may write the estimator \eqref{eq:est1}
equivalently as
\begin{align}
\label{eq:est}
\hbeta =&\; \argmin_\beta \frac{1}{n}\|y-X\beta\|_2^2 + \lrid \|\Pen \beta\|_2^2 + \lone (\ltv \|\Pen \beta\|_1+\|\beta\|_1) \\
\label{eq:est2}
 =&\; \argmin_\beta \frac{1}{n}\|\ty-\tX\beta\|_2^2 + \lone \|\tPen \beta\|_1.
\end{align}

The three regularizers play the following roles:
\begin{itemize}
\item We refer to $\|\Gamma\beta\|_2^2=\sum_{j,k} |\hSigma_{j,k}| (\beta_j - \hat s_{j,k}
\beta_k)^2$ as the {\bf Laplacian smoothing penalty}; \citet{hebiri2011smooth} studied
a variant of this regularizer with $\hSigma_{j,k}$ replaced
with arbitrary non-negative weights. Because each term is squared, it
helps to reduce the ill-conditionedness of $X$ when columns are highly correlated, as reflected in our
analysis.
\item We refer to
  $\|\Gamma\beta\|_1=\sum_{j,k} |\hSigma_{j,k}|^{1/2} |\beta_j - \hat s_{j,k}
  \beta_k|$
  as the {\bf total variation penalty}, as do
  \citet{shuman2013emerging,wang2016trend,sadhanala2016total,hutter2016optimal}; it is
  closely related to the edge LASSO penalty
  \citep{sharpnack2012sparsistency}. Note that these prior works
  consider general weighted graphs (instead of graphs defined by a
  covariance matrix $\hSigma$, as we do). This regularizer
  promotes estimates $\hat \beta$ that are \emph{well-aligned} with the graph
  structure; for instance, a group of nodes with large edge weights connecting them
  (\ie a group of columns of $X$ that are highly correlated) are more
  likely to be associated with coefficient estimates with similar
  values.
\item We refer to $\|\beta\|_1$ as the {\bf sparsity regularizer}. The
  combination of the sparsity regularizer and total variation penalty
  amount to the fused LASSO
  \citep{tibshirani2005sparsity,tibshirani2011}. 
\end{itemize}
The combined effect of the three regularization terms is to find
estimates of $\btrue$ which are both a good fit to the data when the columns
of $X$ are highly correlated and 
well-aligned with
the underlying graph.  This alignment structure may be desirable in a number of settings, including the neural decoding problem considered in the introduction.

\subsection{Contributions} {\em This paper addresses the question of how to estimate $\btrue$ from observations in \eqref{eq:obs} when $X$ has highly-correlated columns.} We propose a regularized regression approach in which {\em  the regularization function depends upon the covariance of $X$}. For a {\em fixed} graph $G$, the proposed estimator is closely related to the previously-proposed fused
LASSO \citep{tibshirani2005sparsity}, generalized LASSO
\citep{tibshirani2011}, edge LASSO
\citep{sharpnack2012sparsistency}, network LASSO
\citep{hallac2015network}, trend filtering \citep{wang2016trend}, and
total-variation regularization
\citep{shuman2013emerging,hutter2016optimal}. In contrast to these
past efforts, {\em our focus is on settings in which columns of $X$ are
highly correlated and these correlations inform the choice of graph
$G$.}

On the other hand there is a large body of work on highly dependent features; in Section~\ref{sec:disc} we provide a thorough comparison of our method with other related approaches. In this paper we make the following contributions:
\begin{itemize}
\item A novel estimator with corresponding finite-sample theoretical
  guarantees for highly-correlated design matrices $X$. General theoretical guarantees for both mean-squared error and variable selection consistency provide insight into the impact of the alignment of $\btrue$ with the covariance graph, and properties of the covariance graph structure such as smallest and largest block-sizes and smallest non-zero eigenvalue. 
  
\item New mean-squared error guarantees for three specific covariance graph structures, a block complete graph, a chain graph, and a lattice graph. Our error bounds match the optimal rates in the independent case where $\Sigma$ is a diagonal matrix, and also match the optimal rates for the block and lattice covariance graphs.  
\item A simulation study which shows that our method out-performs
  state-of-the-art alternatives such as the  Cluster Representative LASSO (CRL, \citet{buhlmann2013correlated}) and Ordered Weighted LASSO (OWL, \citet{Candes_OWL}) in terms of mean squared error in a variety of settings.
\item A validation of our method on real biochemistry data that demonstrates the adavantages of GTV. 
\end{itemize}

The remainder of this paper is organized as follows: 
In Section~\ref{sec:disc} we discuss existing work and results for this problem and its relationship to our estimator; in Section~\ref{sec:main} we present our main theoretical results for both mean-squared error and variable selection consistency; in Section~\ref{sec:Sim} we carry out a simulation study by comparing our methods to other state-of-the-art methods; in Section~\ref{sec:biochem} we apply our method to a real biochemistry dataset with comparison to other methods; we state our conclusions in Section \ref{sec:conclusion}; proofs are provided in the Appendix.

\subsection{Prior work}
\label{sec:disc}
There is a large body of work related to our proposed estimator. 
Significant effort has been devoted to understanding estimators like
\eqref{eq:est} in the special case where $X = I$ -- that is, in a
``denoising'' setting in which observations are direct measurements of
the signal of interest, $\beta$. Variants of these estimators are
often referred to as the edge or network LASSO
\citep{sharpnack2012sparsistency,hallac2015network}, a special case of
graph trend filtering \citep{wang2016trend} or graph total variation
estimation \citep{shuman2013emerging}.  \citet{wang2016trend} consider a generalization of graph total
variation to higher-order measures of variation of signals for
denoising piecewise-polynomial signals on graphs and derive squared error
bounds for the estimates. \citet{hutter2016optimal} also develop sharp oracle inequalities for the edge
LASSO, with an emphasis on a $2$d lattice graph used in image
processing applications. 

In the high-dimensional regression setting, our approach may be viewed as a generalization
of the classical {\em fused LASSO}
\citep{tibshirani2005sparsity}, where instead of promoting alignment between features with adjacent indices, we instead promote alignment of  features that are neighbors in a graph.  Specifically, the {\em generalized LASSO} of \citet{tibshirani2011,liu2013dictionary}
consider the estimators of the form
\begin{equation}
\hat \beta = \argmin_\beta \frac{1}{n}\|y-X\beta\|_2^2 +
\lambda \|\Gamma \beta\|_1
\label{eq:GL}
\end{equation}
for general $X$ and $\Gamma$; note that both the fused LASSO and the
estimator in \eqref{eq:est} can be written in this
form. 

The works \cite{genfused18} and \cite{gene_regulatory} use the generalized LASSO to mitigate correlation effects similar to the approach described in this work, but {\em without theoretical support}.  \cite{genfused18} aims to predict Alzheimer’s disease outcomes using MRI measures as features.  The authors use prior knowledge of correlations between MRI features to construct a regularizer which promotes alignment between correlated features.  \cite{gene_regulatory} seeks to predict outcomes in cancer patients based on gene expression data.  The authors leverage side information of gene regulatory networks and promote alignment between adjacent vertices in the network. This work provides theoretical justification for the approaches described in those papers. 

A related approach is the {\em clustered LASSO} \citep{she2010sparse}, which takes the form
$$
\hat \beta = \argmin_\beta \frac{1}{n}\|y-X\beta\|_2^2 + \ltv
\sum_{1\leq j<k \leq p} |\beta_j-\beta_k| + \lone\|\beta\|_1.
$$
In contrast to the fused LASSO, the clustered LASSO considers {\em
  all} pairwise differences of elements of $\beta$. \citet{she2010sparse} conducts a
classical asymptotic analysis (fixed $p$ and $n \rightarrow \infty$)
of the clustered LASSO and its generalization \eqref{eq:GL}
and establishes consistency results that depend upon $\Sigma^{-1}$.

Related work by \citet{needell2013stable,needell2013near} consider the special case of the
generalized LASSO of total variation regularization on a grid for
image reconstruction problems. That analysis, while elegant, relies
heavily upon the grid-like graph structure associated with pixels in
images and does not generalize to the setting of this paper. 

A key focus of our work is the setting in which columns of $X$ may be
highly correlated. There are several approaches developed to deal with the high-dimensional linear regression problem with some highly correlated covariates. The {\em Elastic Net} estimator proposed by \citet{Zou05} is
\begin{eqnarray}
\hat{\beta}=\argmin_{\beta}\|y-X\beta\|_2^2+\lone\|\beta\|_1+\lrid\|\beta\|_2^2,
\label{eq:elastic}
\end{eqnarray}
which encourages a grouping effect, in which  strongly-correlated
predictors tend to be in or out of the support of the estimate
together. \citet{witten2014} propose a \emph{Cluster Elastic Net} estimator which 
incorporates clustering information inferred from data to perform more accurate regression.  The \emph{Elastic Corr-net} proposed by \cite{anbari_14} proposes combining an $l_1$ penalty with a correlation based quadratic penalty from \cite{tutz_09}.

An alternative approach explored
by \citet{buhlmann2013correlated}, called {\em Cluster Representative
LASSO (CRL)}, clusters highly correlated columns of $X$, chooses a
single representative for each cluster, and regresses over the cluster
centers. \citet{buhlmann2013correlated} also
considered a {\em Cluster Group LASSO (CGL)} in which a group
sparsity regularizer was used with the original design matrix $X$ and
the group structure was determined by a clustering of the columns of
$X$. These two-stage approaches (first cluster, then regress based on
estimated clusters) admitted encouraging statistical guarantees and
empirical performance. However, (i) they depend heavily upon our
ability to find a good clustering of the columns of $X$, where
clusters must be disjoint or non-overlapping; (ii) clustering decisions
are ``hard'' and do not reflect varying degrees of correlation among
columns, and (iii) clusters are formed independently of the observed
responses ($y$). We examine the performance of CRL in this paper. 
{\em Grouping pursuit} \citep{shen2010grouping} explores
clustering columns of $X$ while leveraging $y$ by using a non-convex
variant of the fused LASSO.

Early work on the adaptive LASSO by \citet{Zou06}
illustrated the impact of adaptivity in the correlated design setting.
Recent work on the {\em Ordered Weighted LASSO (OWL)} estimator
\citep{Candes_OWL} proposed an alternative weighted LASSO
regularizer in which the weights depend on the order statistics of
$\beta$; specifically,
$$
	\hat{\beta}=\argmin_{\beta}\|y-X\beta\|_2^2+\lone\sum_{j=1}^p w_j|\beta|_{[j]},
$$
where $w_1\geq w_2\geq...\geq w_p\geq0$ and $|\beta|_{[j]}$ is the $j^{th}$ largest element in $\{|\beta_1|,|\beta_2|,...,|\beta_p|\}$, their paper shows that this family of regularizers can be used for sparse linear regression with strongly correlated covariates. A special case of OWL is the {\em OSCAR} estimator
\citep{bondell2008simultaneous}. \citet{figueiredo2016ordered} demonstrated that when two
columns of $X$ were {\em identical}, then OWL would assign the
corresponding elements of $\beta$ equal values. 
OWL {adaptively} groups highly correlated columns of $X$ by assigning them equal
weights whenever their correlation exceeds a critical value --  the grouping does not need to be
pre-computed and will depend on the value of $y$.

An estimator called \emph{Pairwise Absolute Clustering and Sparsity (PACS)} estimator is proposed by \citet{sharma2013}.
\citet{hebiri2011smooth} consider smooth {\em S-LASSO} estimators of
the form 
$$
\hat \beta = \argmin_\beta \frac{1}{n}\|y-X\beta\|_2^2 +
\lrid \|\Gamma \beta\|_2^2 + \lone \|\beta\|_1.
$$
The first regularization term, unlike the total variation term in
\eqref{eq:est}, is a quadratic penalty similar to what appears in the
elastic net \eqref{eq:elastic} \citep{Zou05}.  The analyses by \citet{she2010sparse}, \citet{sharma2013} and 
\citet{hebiri2011smooth} do not consider settings in which $X$ and
$\Gamma$ in \eqref{eq:GL} are related.
A similar approach to \citet{hebiri2011smooth} is the {\em weighted fusion estimator} proposed by \citet{daye2009shrinkage}. 
\citet{daye2009shrinkage} focus their analysis on
grouping effects, sign consistency, and limiting distributions, but
do not consider finite sample error bounds of the type developed in
this paper. The \emph{Sparse Laplacian Shrinkage (SLS)} estimator proposed by \citet{huang2011} uses a \emph{minimum concave penalty (MCP)} to replace the LASSO penalty in a weighted fusion estimator to reduce bias.

\section{Assumptions and Main Results}
\label{sec:main}
We first introduce a set of assumptions needed for our main results.
Throughout we use the induced matrix norm notation
$$\|A\|_{p,q} = \sup_{x \neq 0} \frac{\|Ax\|_q}{\|x\|_p}.$$
Specifically, note that $\|A\|_{1,2}$ is the maximum column norm of $A$ and $\|A\|_{op}= \|A\|_{2,2}$. For a symmetric positive semi-definite matrix $A$, let $\lambda_{\min}(A)$ denote its minimum eigenvalue and $\lambda_{\max}(A)$ denote its maximum eigenvalue.

The notation $X_n=O_P(a_n)$ means that the set of values $\frac{X_n}{a_n}$ is stochastically bounded. That is, for any $\epsilon>0$, there exists a finite $M>0$ and a finite $N>0$ such that
$$\prob\left(\left|\frac{X_n}{a_n}\right|>M\right)<\epsilon,~\forall~n>N.$$

\bas
\label{as:tvlambdasigma}
We assume that there exists an absolute constant $c_u>0$ such that 
\begin{eqnarray*}
\lambda_{\max}(\Sigma)\leq c_u.
\end{eqnarray*}
\eas
\brem
This statement assumes that $\Sigma$ is normalized such that the largest eigenvalue of $\Sigma$ can be upper bounded by a positive constant. 
\erem

\bas
\label{as:sigmanormalization}
There exists an absolute constant $c_{\ell}>0$ such that:
\begin{eqnarray*}
c_{\ell}\leq \min_{1 \leq j \leq p}\sum_{k=1}^p|\Sigma_{j,k}|.
\end{eqnarray*}
\eas

\brem
Assumption~\ref{as:sigmanormalization} ensures the $\ell_1$ norm for each row/column is lower bounded by a constant. This assumption is much milder than assuming the minimum eigenvalue of $\Sigma$ is bounded away from $0$.  As an example, Assumption~\ref{as:sigmanormalization} is satisfied when every diagonal entry of $\Sigma$ is bounded below by $c_{\ell}$.  Note that Assumption \ref{as:tvlambdasigma} automatically holds for appropriately normalized features.  However the assumption is nontrivial when considered jointly with Assumption \ref{as:sigmanormalization}, because normalization can potentially cause a violation of Assumption \ref{as:sigmanormalization}.  We show that both Assumptions \ref{as:tvlambdasigma} and \ref{as:sigmanormalization} hold in the examples considered in Section \ref{sec:specificgraph}.

\erem

\bas
\label{as:estimatecov}
The estimated covariance matrix $\hat{\Sigma}$ that is used to construct the matrix $\Gamma$ satisfies
\begin{eqnarray*}
\|\hat{\Sigma}-\Sigma\|_{1,1} = \max_{1 \leq j \leq p}\sum_{k=1}^p|\hSigma_{j,k}-\Sigma_{j,k}|\leq\frac{c_{\ell}}{4},
\end{eqnarray*}
where $c_{\ell}$ is defined in Assumption \ref{as:sigmanormalization}.
\eas
\brem
Assumption~\ref{as:estimatecov} states that we need a sufficiently accurate estimator $\hSigma$ for $\Sigma$. If Assumption~\ref{as:estimatecov} is satisfied then we can use $\hSigma$ to construct $\Gamma$ for our optimization problem stated in (\ref{eq:est2}). 
We estimate $\Sigma$ using side information that is not necessarily based on $(X^{(i)})_{i=1}^n$. For instance, in the cytochrome P450 enzyme setting described in Section~\ref{sec:applications}, we can leverage the recombination information of each chimeric protein to help estimate $\Sigma$.  We elaborate on this in Section \ref{sec:biochem_esti}.  In an MRI context, one can leverage prior knowledge of correlations between MRI features \cite{genfused18}.  In climate forecasting settings, physics-based simulations can be used to generate accurate covariance estimates. 

In some settings, our source of side information may not directly yield an estimate of $\Sigma$, but rather  a collection of $m$ i.i.d.\ unlabeled feature vectors $(\check X^{(i)})_{i=1}^m$ that are potentially independent of the design features $(X^{(i)})_{i=1}^n$ with $\check X^{(i)} \sim \cN(\textbf{0},\Sigma_{p \times p})$. 
In this case, we need to estimate $\Sigma$ based on $(\check X^{(i)})_{i=1}^m$, and there is a large literature on high-dimensional covariance estimation in high dimensions under different structural assumptions (see \citet{BicLev06,BicLev07,cai2011adaptive,cai2016,donoho2013optimal,BaiSil06}). As an example, we consider estimators based on thresholding the sample covariance matrix under block structural assumptions developed by~\citet{BicLev07}.  We show that when the covariance matrix is block structured with $K$ blocks, and $m=O(K^2\log p)$,  Assumption \ref{as:estimatecov} is satisfied.
See Appendix~\ref{app:cov} for more details.
\erem

The performance of our estimator also depends upon the following two properties of the augmented edge incidence matrix $\tPen$ appearing in our regularizer:
\bde[Compatibility factor $k_T$, \citet{hutter2016optimal}] 
\label{def:compatibilityfactor}
We define the compatibility factor $k_T$ of matrix $\tilde{\Gamma}$ for a set $T\subset\{1,2,...,p,p+1,...,p+m\}$ as:
\begin{eqnarray*}
k_{\emptyset}:=1,~k_T:=\inf_{\beta\in\mathbb{R}^p}\frac{\sqrt{|T|}\|\beta\|_2}{\|(\tPen\beta)_T\|_1}~\text{for}~T\neq\emptyset.
\end{eqnarray*}
\ede
This compatibility factor $k_T$ reflects the degree of compatibility of the $\ell_1$-regularizer $\|(\tilde{\Gamma}\beta)_T\|_1$ and the $\ell_2$-error norm $\|\beta\|_2$ for a set $T$. This compatibility factor appears explicitly in
the bounds of our main theorem.

\bde[Inverse scaling factor $\rho$, \citet{hutter2016optimal}]
\label{def:pseudoinverse}
Let $S:=\tPen^{\dagger}=[s_1,...,s_{m+p}]$, where
$\tPen^{\dagger}$ is the Moore-Penrose pseudoinverse of the matrix
$\tPen$, and  define the inverse scaling factor as:
$$\rho:=\|\tPen^{\dagger}\|_{1,2}=\max_{j=1,2,...,m+p}\|s_j\|_2.$$

\ede

\brem Definitions \ref{def:compatibilityfactor} and
\ref{def:pseudoinverse} are first proposed in
\citet{hutter2016optimal}, though the definition of $\rho$ is based on $\tPen$ rather than $\Pen$. Later we will see that $\rho$ and $k_T$ are crucial for our main results.  The quantity $\frac{\rho}{k_T}$ is similar in flavour to the condition number of the matrix $\tilde{\Gamma}$.

\erem

Finally, we define the \emph{estimated graph Laplacian} $L:= \Gamma^{\top} \Gamma$. Recall that $\Gamma$, and therefore $L$, are constructed using the estimated covariance matrix $\hSigma$ rather than $\Sigma$.  Spectral properties of $L$ will play a crucial role in the mean-squared error bounds we derive.

\btheos
\label{theo:tvMain}
Suppose $\lambda_1>0$ and Assumptions \ref{as:tvlambdasigma} to \ref{as:estimatecov} are
satisfied. If
$$\lone\geq\max\left\{48 \sqrt{\frac{{c_u}\rho^2 \sigma^2\log p}{n}},8\lrid \|L \bstar\|_{\infty} \right\},$$ 
then there exist absolute positive constants $C_u$ and $C_1$ such that with probability at least $1-\frac{C_1}{p}$ we have
$$
\|\hat{\beta}-\bstar\|_2^2\leq C_u\min_{T}\max\left\{\frac{\lone^2|T|}{k_T^2\lambda_{\min}^2(\Sigma+\lrid L)},\frac{\lone \|(\tilde{\Gamma}\bstar)_{T^c}\|_1+\lone^2 \|(\tilde{\Gamma}\bstar)_{T^c}\|^2_1}{\lambda_{\min}(\Sigma+\lrid L)}\right\}
$$
provided $\frac{\lone^2|T|}{k_T^2\lambda^2_{\min}(\Sigma+\lrid L)} \rightarrow 0$ (\ie that the estimator is consistent).
\etheos

\brem
Here $\lambda_{\min}(\Sigma+\lrid L)$ plays the
  role of the restricted eigenvalue constant (see \citet{BiRiTsy08} for
  more details about this condition). Recall that from the definition of $L$, if we define the diagonal matrix $D\in\mathbb{R}^{p\times p}$ where each diagonal entry is $D_{jj}=\sum_{k=1}^p|\hSigma_{j,k}|,~1\leq j\leq p$, then 
  $$
  \Sigma + L := \Sigma - \hSigma + D.
  $$
  Hence if $\Sigma$ and $\hSigma$ are ``close'' as is specified by Assumption~\ref{as:estimatecov}, then $\Sigma+L$ is ``close'' to a diagonal matrix which ensures that $\lambda_{\min}(\Sigma+\lrid L)$ may be bounded away from $0$, even if $\lambda_{\min}(\Sigma) = 0$. The following Lemma makes this statement precise:
    \blems
\label{lemma:eigenvalue}
Suppose that Assumption \ref{as:sigmanormalization} and \ref{as:estimatecov} are satisfied and $0 \leq \lrid \leq 1$. Then
\begin{eqnarray*}
\lambda_{\min}(\Sigma+\lrid L)\geq (1 -\lrid)\lambda_{\min}(\Sigma)  + \lrid\frac{c_{\ell}}{2}.
\end{eqnarray*}
\elems
Thus even if $\lambda_{\min}(\Sigma) = 0$, choosing $\lrid$ bounded away from $0$ results in a well-posed inverse problem. On the other hand, in the classical LASSO analysis where $\lambda_{\min}(\Sigma) > 0$, we can choose $\lrid = 0$. 
\erem 
  
\brem
$\|L \bstar \|_{\infty}$ can be seen as a measure of the
\emph{misalignment} of the signal $\bstar$ and the graph
represented by the matrix $\Gamma$. Note that we require
$\lone\geq8\lambda_S\|L \bstar\|_{\infty}$. Hence there is a clear trade-off in the choice of $\lambda_S$. Choosing $\lrid$ close to $1$ ensures $\lambda_{\min}(\Sigma+\lrid L)$ is bounded away from $0$ but incurs a cost that scales with $\|L \bstar\|_{\infty}$. 
   
In general, if $\lambda_{\min}(\Sigma) = 0$, indicating high correlations, we require $\|L \bstar\|_{\infty} \approx 0$ (\ie $\bstar$ is well-aligned with $L$) in order to obtain consistent mean-squared error bounds. Note that analysis of
OWL~\citep{figueiredo2016ordered} assumes $L \bstar = \textbf{0}$ (perfect alignment). If $\lambda_{\min}(\Sigma) = 0$ and $\|L \bstar\|_{\infty}$ is bounded far away from $0$, we encounter identifiability challenges which leads to an inconsistent estimator of $\bstar$, just like the classical LASSO.
\erem

\brem \label{rem:smooth}
A natural question to consider is how  the mean-squared error bound would change if the graph Laplacian penalty $\lrid \|\Gamma \beta\|_2^2$ were replaced by $\lrid \|\beta\|_2^2$ as is used in the 
\citep{Zou05}. Going through the analysis, $\lambda_{\min}(\Sigma+\lrid L)$ would be replaced by $\lambda_{\min}(\Sigma+\lrid I_{p \times p})$ and hence pre-conditioning is still achieved. However the important difference and why we prefer the graph Laplacian penalty is because using our analysis the condition $\lone\geq8\lambda_S\|L \bstar\|_{\infty}$ would be replaced by $\lone\geq8\lambda_S\|\bstar\|_{\infty}$. 
Hence if we were in the strictly sparse case and $\ltv = 0$ we would recover the mean-squared error bound:
\begin{eqnarray*}
\|\hat{\beta}-\beta^{\ast}\|_2^2\preceq \frac{(\frac{\log p}{n}+\lambda_S^2\|\beta^{\ast}\|^2_{\infty})\|\beta^{\ast}\|_0}{\lambda^2_{\min}(\Sigma+\lambda_SI_{p\times p})}.
\end{eqnarray*}
Note that this exactly matches the mean-squared error bound in (11) in \cite{hebiri2011smooth} by replacing $\|\beta^*\|_2^2$ with the bound $\|\beta^{\ast}\|_0 \|\beta^*\|_\infty^2$. 
(The estimator of \citet{hebiri2011smooth} is a generalization of Elastic Net from \citet{Zou05}.)
In general we can not expect $\|\bstar\|_{\infty}$ to be close to zero, but in the case where $\bstar$ is well-aligned with $L$, we would expect $\|L \bstar\|_{\infty}$ to be close to zero which would achieve sharper bounds.
\erem

Now we turn our attention to quantifying $k_T$ and $\rho$ to provide a more interpretable bound. We first have the following lemma to bound $k_T$:
\blems
\label{lemma:kT}
Suppose $T=T_1\cup T_2$ with $T_1\subset\{p+1,p+2,...,p+m\}$ and $T_2\subset\{1,2,...,p\}$. Then we have
\begin{eqnarray*}
k_T^{-1}\leq\frac{\ltv\sqrt{2\|\hSigma\|_{1,1}|T_1|}+\sqrt{|T_2|}}{\sqrt{|T_1|+|T_2|}}.
\end{eqnarray*}
\elems
\noindent The proof for this lemma can be found in  Appendix~\ref{sec:kT}.
\brem
The compatibility factor $k_T$ depends on the choice of support $T$. Usually $T$ will be chosen as $T=\text{Supp}(\tilde{\Gamma}\beta)$ for some $\beta$; then $T_1=\text{Supp}({\Gamma}\beta)$ and $T_2=\text{Supp}(\beta)$ and Lemma \ref{lemma:kT} can be reduced to
\begin{eqnarray*}
k_T^{-1}\leq\frac{\ltv\sqrt{2\|\hSigma\|_{1,1}\|\Gamma\beta\|_0}+\sqrt{\|\beta\|_0}}{\sqrt{\|\Gamma\beta\|_0+\|\beta\|_0}}.
\end{eqnarray*}

\erem
To provide an upper bound for $\rho$ we first define the following graph-based quantities. If $G$ has $K$ connected components where $1 \leq K \leq p$, $L$ is block-diagonal with $K$ blocks. Let $L_k$ denote the $k^\text{th}$ block of $L$, $B_k \subset \{1,2,...,p\}$ denote the nodes corresponding to the $k^\text{th}$ block, and $\mu_k$ denote the smallest non-zero eigenvalue of $L_k$.
\blems
Let $G$ denote the graph associated with $\hSigma$. Then 
\label{lemma:rho}
$$\rho^2\leq\max_{1\leq k\leq K}\left\{\frac{1}{|B_k|}+\frac{2}{1+\mu_k\ltv^2}\right\},$$
where $K$ is the number of connected components in  $G$; $|B_k|$ is the corresponding number of nodes in $B_k$; and $\mu_k$ is the smallest nonzero eigenvalue of the
weighted Laplacian matrix for the $k^\text{th}$ connected component. 
\elems                                                  

By combining results from Lemmas \ref{lemma:kT} and \ref{lemma:rho} we have the following theorem:
\btheos
\label{theo:tvMain2}
Suppose Assumptions \ref{as:tvlambdasigma} to \ref{as:estimatecov} are satisfied and we choose
$$
\lambda_1\geq48\sqrt{\frac{\sigma^2 {c_u} \log p}{n} \max_{1 \leq k \leq K}\biggr(\frac{1}{|B_k|} + \frac{2}{1+ \mu_k\ltv^2}}\biggr) + 8 \lambda_S\|L \beta^{\ast}\|_{\infty}.
$$
Then there exist absolute positive constants $C_1$ and $C$ such that 

\begin{eqnarray*}
\|\hat{\beta}-\bstar\|_2^2 \leq C \frac{\lambda_1^2\|\beta^{\ast}\|_0 + \min(\lambda_1^2 \lambda_{TV}^2\|\hSigma\|_{1,1}\|\Gamma\beta^{\ast}\|_0,\lambda_1\lambda_{TV}\|\Gamma\beta^{\ast}\|_1)}{\min(\lambda_{\min}^2(\Sigma+\lrid L),\lambda_{\min}(\Sigma+\lrid L))},
\end{eqnarray*}
with probability at least $1-\frac{C_1}{p}$ provided $\frac{\lambda_1^2\|\beta^{\ast}\|_0 + \lambda_1^2 \lambda_{TV}^2\|\hSigma\|_{1,1}\|\Gamma\beta^{\ast}\|_0}{\lambda^2_{\min}(\Sigma+\lrid L)} \rightarrow 0$ and $\lambda_1\lambda_{TV}\|\Gamma\beta^{\ast}\|_1\leq1.$
\etheos

The proof of Theorem \ref{theo:tvMain2} is provided in Section \ref{sec:Proof}. The upper bound involves a minimum where one term depends on $\|\Gamma\beta^{\ast}\|_0$ and the other depends on $\|\Gamma\beta^{\ast}\|_1$ by using different choices of $T$. This minimum of two terms also appears in~\citet{hutter2016optimal}. Theorem~\ref{theo:tvMain2} captures the role of $\lambda_{TV}$ and its impact on the mean-squared error (MSE) bounds. {\em As $\lambda_{TV}$ increases,  $\|\beta^{\ast}\|_0$ contributes less to the MSE, while $\|\Gamma\beta^{\ast}\|_0$ or $\|\Gamma\beta^{\ast}\|_1$ contributes more.} To see this, note that  the lower bound on $\lambda_1$ decreases with $\lambda_{TV}$ and the first term in the MSE scales as $\lambda_1^2\|\beta^{\ast}\|_0$. On the other hand the second term of the MSE scales as $\lambda_1^2 \lambda_{TV}^2\|\hSigma\|_{1,1}\|\Gamma\beta^{\ast}\|_0$ or $\lambda_1 \lambda_{TV}\|\Gamma\beta^{\ast}\|_1$ and the lower bound on $\lambda_1 \lambda_{TV}$ increases as $\lambda_{TV}$ increases.  Determining optimal error rates is in general a challenging problem.  However, in the special cases of the block and lattice graphs considered in Section \ref{sec:specificgraph} our bounds are consistent with known optimal rates.  It is straightforward to extend the proofs of Theorems \ref{theo:tvMain} and \ref{theo:tvMain2} in order to derive prediction error bounds on $||X\hat{\beta}-X\beta^\ast||_2^2$.  This is discussed in more detail in Appendix~\ref{sec:prediction_error}.

\subsection{Discussion of main results}

If we are in the setting where $\lambda_{\min}(\Sigma) > C > 0$, which corresponds to the classical LASSO setting, we can set $\lambda_S=\ltv=0$. From Theorem \ref{theo:tvMain2} we can see that 
\begin{eqnarray}
    \|\hat{\beta}-\bstar\|_2^2 \preceq \frac{\sigma^2{c_u}\log p}{n}\|\beta^{\ast}\|_0,
    \label{eq:bndLASSO}
    \end{eqnarray}
which is consistent with classical LASSO results.
On the other hand if $\lambda_{\min}(\Sigma) \approx  0$ (columns are highly correlated) and $\|L \beta^{\ast}\|_{\infty} \approx 0$ ($ \beta^{\ast}$ is well-aligned with $L$), we can set $0 < \lambda_S  \leq 1$ and $\ltv = C \max_{1 \leq k \leq K}\sqrt{\frac{|B_k|}{\mu_k}}$; then we obtain the bound 
\begin{eqnarray*}
    \|\hat{\beta}-\bstar\|_2^2 \preceq \lambda_1^2 \|\beta^{\ast}\|_0 + \min(\lambda_1^2 \lambda_{TV}^2 \|\hSigma\|_{1,1}\|\Gamma \beta^{\ast}\|_0, \lambda_1 \lambda_{TV}\|\Gamma\beta^{\ast}\|_1)
\end{eqnarray*}
where $\lambda_1^2 = O(\max_{1\leq k\leq K}\frac{\sigma^2 c_u \log p}{n|B_k|})$ and $\lambda_1^2 \ltv^2 = O(\max_{1 \leq k \leq K}\frac{|B_k|}{\mu_k}\max_{1\leq k\leq K}\frac{\sigma^2 c_u \log p}{n|B_k|})$. The upper bound may be well below the classical LASSO bound in~\eqref{eq:bndLASSO} when $\min_k |B_k| \gg 1$ and $\Gamma\beta^{\ast} \approx \textbf{0}$. 
    
As mentioned earlier, if $\lambda_{\min}(\Sigma) \approx 0$ (columns are highly correlated) but $\|L \beta^{\ast}\|_{\infty}>C>0$ (bad alignment), our method cannot guarantee a consistent estimator for $\beta^{\ast}$; Cluster Representative LASSO  and Ordered Weighted LASSO 
will also fail in this case. Identifiability assumptions arise, since if two columns of $X$ are nearly identical but the corresponding elements of $\bstar$ are substantially different, no method will be able to accurately estimate parameter values in the absence of additional structure.

We now discuss the roles of the various parameters associated with the MSE bound.

    \paragraph*{Role of $\lambda_S$}The smoothing penalty plays the role of a pre-conditioner where the trade-off is the addition of another term $\lambda_S \|L \beta^{\ast}\|_{\infty}$. This can also be seen in the optimization problem (\ref{eq:est2}) where $X$ is transformed to $\tilde{X}$, so even if the restricted eigenvalue condition is not satisfied for $X$, it is satisfied for $\tilde{X}$. What distinguishes our results from previous work using pre-conditioners for the LASSO~\citep{jia2015preconditioning,NIPS2013_5104} is that prior work does not address the case where $\lambda_{\min}(\Sigma) = 0$, which is where the total variation penalty is important. See also Remark~\ref{rem:smooth}.

\paragraph*{Role of $\lambda_{TV}$}As mentioned earlier, the total variation penalty promotes estimates well-aligned with the graph. As $\lambda_{TV}$ increases, the sparsity parameter $\lambda_1$ decreases while $\lambda_1 \lambda_{TV}$ increases. By increasing $\lambda_{TV}$ we can also adapt to settings where $\beta^{\ast}$ is not sparse provided that $\Gamma \beta^{\ast}$ is sparse (see the examples of specific graph structures below).

\paragraph*{Graph-based quantities}Two important parameters of the covariance graph are $\mu_k$ (the smallest non-zero eigenvalue of a block) and $|B_k|$ (the block size). Clearly the larger $\mu_k$ and $|B_k|$, the lower the bound on $\lambda_1$ which potentially suggests lower mean-squared error. On the other hand, as we illustrate with specific examples later, larger $\mu_k$ typically indicates higher correlation between more covariates and larger $|B_k|$ corresponds to nodes being correlated, which means $\lambda_{\min}(\Sigma)$ is smaller.

\subsection{Specific covariance graph structures}
\label{sec:specificgraph}
In this section we explore three specific graph structures and discuss suitable choices of $\lrid, \lone$ and $\ltv$. For each graph structure we assume  
$$\Sigma_{jj}=a>0 \mbox{ for } 1\leq j\leq p \qquad \mbox{ and } \qquad \Sigma_{jk} = ar \; \forall (j,k) \in E \mbox{ for some } 0 < r \leq 1;$$ we refer to $r$ as the correlation coefficient. Note that here $a$ is a normalization parameter that we set to ensure such that Assumptions \ref{as:tvlambdasigma} and \ref{as:sigmanormalization} are satisfied. We will talk about the specific choices of $a$ for each graph structure below. 
Our general results allow us to quantify the impact of misspecification of $\Sigma$, but for interpretability and simplicity of exposition, we will assume in this section that $\hSigma = \Sigma$ -- that is, that we have perfect side information about the correlation graph.

\subsubsection{Block covariance graph}
\label{sec:blockcomplete}
We first consider a  block complete graph $G$ that has $K$ connected components and each connected component is a complete graph with $\frac{p}{K}$ nodes. The corresponding covariance matrix $\Sigma$ (potentially after a suitable permutation of rows and columns) is block diagonal with $K$ blocks  of  size $\frac{p}{K}\times\frac{p}{K}$. Each of these blocks can be written as
$$ar \ones_{p/K}\ones_{p/K}^\top + a(1-r)I_{p/K},$$
where $\ones_{p/K}$ is the vector of $p/K$ ones.

We set  $a=\frac{K}{p}$ to ensure that Assumptions \ref{as:tvlambdasigma} and \ref{as:sigmanormalization} are satisfied. In the extreme case where $K = p$, we are in the independent case and the estimator reduces to the standard LASSO estimator; whereas for $K = 1$, we are in the fully-connected graph case.

The following lemma provides specific bounds on $\max_{1 \leq k \leq K} \frac{1}{|B_k|}, \mu_k, \rho, \lambda_{\min}(\Sigma+\lrid L)$:
\blems
\label{lemma:blockcomplete}
For a block complete graph with details described above, suppose that $\hSigma=\Sigma$.  Then we have 
\begin{gather*}
\max_{1 \leq k \leq K} \frac{1}{|B_k|} = \frac{K}{p},\\ \mu_k = r,\;\;\mbox{for all}\;\;k\\
\rho \leq \sqrt{\frac{K}{p}+\frac{2}{1+r\ltv^2}},\\\lambda_{\min}(\Sigma+\lrid L)\geq (1-\lambda_S)(1-r)\frac{K}{p}+\lambda_Sr.
\end{gather*}
\elems
The proof of Lemma~\ref{lemma:blockcomplete} is deferred to  Appendix~\ref{sec:blockCompleteProof}. Note that if $r = 1$ then $\lambda_{\min}(\Sigma) = 0$ but $\lambda_{\min}(\Sigma+\lrid L) \geq \lrid$. Using Lemma~\ref{lemma:blockcomplete}, we have the following mean-squared error bound for the block complete graph:
\bcors
For a block complete graph with details described above, suppose that $\hSigma=\Sigma$.  If 
$$
\lone\geq48\sqrt{\frac{\sigma^2 c_u \log p}{n}\left(\frac{K}{p}+\frac{2}{1+r\lambda^2_{TV}}\right)}+8\lrid \|L \bstar\|_\infty$$
and $\lambda_1\lambda_{TV}\|\Gamma\beta^{\ast}\|_1\leq1$. Then with probability at least $1-\frac{C_1}{p}$
\begin{eqnarray*}
\|\hat{\beta}-\bstar\|_2^2 \leq \frac{C\big(\lambda_1^2 \|\beta^{\ast}\|_0 + \min\{\lambda_1^2 \lambda_{TV}^2\|\Gamma\beta^{\ast}\|_0,\lambda_1\lambda_{TV}\|\Gamma\beta^{\ast}\|_1\}\big) }{\min\{[(1-\lambda_S)(1-r)\frac{K}{p}+\lambda_Sr],[(1-\lambda_S)(1-r)\frac{K}{p}+\lambda_Sr]^2\}}
\end{eqnarray*}
given the estimator is consistent, where $C_1$, $C$ are absolute positive constants. \label{cor:block}
\ecors

Consider a setting where $r \approx 1$ and $\Gamma  \beta^{\ast} \approx \textbf{0}$ (near-perfect alignment which corresponds to the parameters in each block having the same values).  Let $K_1 \leq K$ denote the number of blocks which have features that are active in $\btrue$.  If we choose $\lambda_S \asymp 1$, $\lambda_{TV}^2 \asymp \frac{p}{K}$, and $\lambda_1^2 \asymp \frac{K\log p}{pn}$, then
$$\|\hbeta-\btrue\|_2^2 \preceq \frac{K_1 \log p}{n};$$ that is, the MSE is not determined by the number of nonzeros in $\beta^{\ast}$, but rather by $K_1$, the number of clusters of active nodes.  In the case of perfect correlation between the blocks this matches the minimax optimal rate up to log factors (\cite{RasWaiYu11}).  A similar scaling was derived in~\citet{figueiredo2016ordered} also under the assumption that $\Pen\bstar\approx\textbf{0}$.

\subsubsection{Chain covariance graph}
\label{sec:chain}
The covariance matrix correspnding to the chain graph satisfies $\Sigma_{jj}=1$ for all $j$ and $\Sigma_{jk} = r$ for all $(j,k) \in E$ where $E = \{(1,2), (2,3),...,(p-1, p)\}$. Assumptions \ref{as:tvlambdasigma} and \ref{as:sigmanormalization} are clearly satisfied and requiring $r\in(0,\frac{1}{2})$ ensures $\Sigma$ is positive semi-definite.  Note that the chain graph is fully connected so $K=1$ and $B_1=\{1,2,...,p\}$.

The following lemma provides bounds on $\rho$ and  $\lambda_{\min}(\Sigma+\lrid L)$ for the chain covariance graph:
\blems
\label{lemma:chain}
For a chain graph with details described above, suppose that $\hSigma=\Sigma$. Then
\begin{gather*}
\rho \leq \sqrt{\frac{1}{p}+\frac{2\pi}{r\lambda_{TV}+1}},\\
\lambda_{\min}(\Sigma+\lrid L) \geq (1-\lambda_S)(1-2r)+\lambda_S.
\end{gather*} 
\elems
Using Lemma \ref{lemma:chain} we have the following corollary for the chain graph: 
\bcors
For a chain graph with details described above, suppose that $\hSigma=\Sigma$.  If we choose
\begin{eqnarray*}
\lone > 48\sqrt{\frac{\sigma^2 c_u \log p}{n}\left(\frac{1}{p}+\frac{2\pi}{r\lambda_{TV}+1}\right)}+8\lrid \|L\bstar\|_\infty
\end{eqnarray*}
and $\lambda_1\lambda_{TV}\|\Gamma\beta^{\ast}\|_1\leq1$, then with probability at least $1-\frac{C_1}{p}$ we have 
\begin{eqnarray*}
\|\hat{\beta}-\bstar\|_2^2 \leq \frac{C\big(\lambda_1^2 \|\beta^{\ast}\|_0 + \min\{\lambda_1^2\lambda_{TV}^2\|\Gamma\beta^{\ast}\|_0,\lambda_1\lambda_{TV}\|\Gamma\beta^{\ast}\|_1\}\big)}{\min\{[(1-\lambda_S)(1-2r)+\lambda_S],[(1-\lambda_S)(1-2r)+\lambda_S]^2\}}
\end{eqnarray*}
given the estimator is consistent, where $C_1$, $C$ are absolute positive constants. \label{cor:chain}
\ecors

We consider an example where the alignment between the chain graph and $\beta^\ast$ is strong but imperfect.  Suppose that within $\beta^\ast$ there are $O(1)$ blocks which are active, and within each active block all the coefficients have identical magnitude.  Further, suppose $n \preceq p$.  In this setting, $\|\Gamma \beta^\ast\|_0, \|\Gamma \beta^\ast\|_1 \approx 1$. 

If we set $\lambda_{TV}\approx \sqrt{\|\beta^\ast\|_0}$ and $\lambda_S \approx 0$ then Corollary \ref{cor:chain} says $$\text{MSE}_{\text{GTV}} \preceq \frac{\sqrt{\|\beta^\ast\|_0}\log p}{n}$$ which is stronger than the LASSO guarantee of $$\text{MSE}_{\text{LASSO}} \preceq \frac{\|\beta^\ast\|_0\log p}{n}.$$

\subsubsection{Lattice covariance graph}
\label{sec:lattice}
We next consider a covariance structure corresponding to a lattice graph with $p$ nodes (here $p$ must be a perfect square).  Both sides of such a lattice have length $\sqrt{p}$ and the corresponding covariance matrix satisfies
$$\Sigma_{j,k}=\begin{cases}
1, & \text{if}~j=k,\\
r, & \text{if}~|j-k|=1 \text{ and } \min(j,k) \not =0 \bmod \sqrt{p} ,\\
r, & \text{if}~|j-k|=\sqrt{p} \\
0, & \text{else}.
\end{cases}
$$
We require $r \in (0,\frac{1}{4})$ so that $\Sigma$ is positive semi-definite.  Clearly Assumptions \ref{as:tvlambdasigma} and \ref{as:sigmanormalization} are satisfied for any $r\in(0,\frac{1}{4})$, and we note that the lattice graph is fully connected, so $K=1$ and $B_1=\{1,2,...,p\}$.
The following lemma gives bounds on $\rho$ and  $\lambda_{\min}(\Sigma+\lrid L)$:
\blems
\label{lemma:lattice}
For a lattice graph with details described above, suppose that $\hSigma=\Sigma$. Then
\begin{gather*}
\rho \leq \sqrt{\frac{1}{p}+\frac{5\pi\log(2+r\lambda_{TV})}{r^2\lambda_{TV}^2+1}+\frac{10\pi}{r\lambda_{TV}\sqrt{p}+1}},\\
\lambda_{\min}(\Sigma+\lrid L) \geq (1-\lambda_S)(1-4r)+\lambda_S.
\end{gather*} 
\elems
Using Lemma \ref{lemma:lattice} we have the following corollary for the lattice graph: 
\bcors
For a lattice graph with details described above, suppose that $\hSigma=\Sigma$. If we choose
\begin{eqnarray*}
\lone > 48\sqrt{\frac{\sigma^2 c_u \log p}{n}\left(\sqrt{\frac{1}{p}+\frac{5\pi\log(2+r\lambda_{TV})}{r^2\lambda_{TV}^2+1}+\frac{10\pi}{r\lambda_{TV}\sqrt{p}+1}}\right)}+8\lrid \|L\bstar\|_\infty
\end{eqnarray*}
and $\lambda_1\lambda_{TV}\|\Gamma\beta^{\ast}\|_1\leq1$, then with probability at least $1-\frac{C_1}{p}$ we have 
\begin{eqnarray*}
\|\hat{\beta}-\bstar\|_2^2 \leq \frac{C\big(\lambda_1^2 \|\beta^{\ast}\|_0 + \min\{\lambda_1^2\lambda_{TV}^2\|\Gamma\beta^{\ast}\|_0,\lambda_1\lambda_{TV}\|\Gamma\beta^{\ast}\|_1\}\big)}{\min\{[(1-\lambda_S)(1-4r)+\lambda_S],[(1-\lambda_S)(1-4r)+\lambda_S]^2\}}
\end{eqnarray*}
given the estimator is consistent, where $C_1$, $C$ are absolute positive constants. \label{cor:lattice}
\ecors

We again consider an example where the alignment between the graph and $\beta^\ast$ is strong but imperfect.  Suppose that all the active nodes within a $\sqrt{p} \times \sqrt{p}$ lattice are contained in a $\sqrt{\|\beta^\ast\|_0} \times \sqrt{\|\beta^\ast\|_0}$  sublattice, and suppose all active nodes have equal magnitude.  Then $\|\Gamma \beta^\ast\|_0, \|\Gamma \beta^\ast\|_1 \approx \sqrt{\|\beta^\ast\|_0}$.

We assume $n \preceq p$ and we set $\lambda_{TV} \approx \sqrt{n}$, $\lambda_S \approx 0$ and $\lambda_1 \approx \frac{\log p}{n}$.  Corollary \ref{cor:lattice} says

$$\text{MSE}_{\text{GTV}} \preceq \lambda_1^2 \|\beta^\ast\|_0+\lambda_1^2\lambda_{TV}^2 \|\Gamma \beta\|_0 \approx \frac{\|\beta^\ast\|_0\log p}{n^2}+\frac{\sqrt{\|\beta^\ast\|_0}\log p}{n} \approx \frac{\sqrt{\|\beta^\ast\|_0}\log p}{n}$$ which is stronger than the LASSO guarantee of $$\text{MSE}_{\text{LASSO}} \preceq \frac{\|\beta^\ast\|_0 \log p}{n}.$$ 

Note that the $\text{MSE}_{\text{GTV}}$ bound from this example is identitcal to the $\text{MSE}_{\text{GTV}}$ bound from the example considered in the chain graph section.  On one hand, our bound on $\rho$ is stronger in the lattice graph case.  This is consistent with \cite{hutter2016optimal} even though we study the inverse scaling factor of a somewhat different matrix.   However, this phenomenon is counterbalanced by the fact that it is easier to construct near perfect alignment between the chain graph and $\beta^\ast$ than between the lattice graph and $\beta^\ast$.  With the chain graph, for any value of $||\beta^\ast||_0$ we can have $||\Gamma \beta^\ast||_0 \approx 1$.  However, for the lattice graph it is impossible to give a general bound on $||\Gamma \beta^\ast||_0$ which is independent of $||\beta^\ast||_0$.  The best possible alignment yields $||\Gamma \beta^\ast||_0 \approx \sqrt{||\beta^\ast||_0}$.  Our overall rate matches the optimal rates derived in the lattice graph denoising setting considered in \cite{hutter2016optimal}.

\section{Simulation study}
\label{sec:Sim}
In this section we compare our proposed graph-based regularization method with other methods on the block, chain and lattice graphs considered in the corollaries above.  Specific details on how the covariance matrix $\Sigma$ is constructed for each graph structure is discussed in Section \ref{sec:sim_details} in the Appendix.  The data is generated according to $y=X\bstar+\epsilon$ with $X\in\mathbb{R}^{n\times p}$ and $y\in\mathbb{R}^n$. Each row of $X$ is independently generated from ${\cal N}(\textbf{0},\Sigma_{p\times p})$ and $\epsilon$ is generated from ${\cal N}(\textbf{0},\sigma^2I_{n\times n})$ with $\sigma=0.01$. Additionally, we generate $X_{\text{ind}}\in\mathbb{R}^{1000\times p}$ with each row of $X_{\text{ind}}$  independently generated from ${\cal N}(\textbf{0},\Sigma_{p\times p})$. This $X_{\text{ind}}$ provides side information that can be used to improve estimates of  $\Sigma$. This $X_{\text{ind}}$ can be used for covariance estimation (GTV) or clustering (CRL) before parameter estimation.

We show how our proposed graph-based regularization scheme compares to existing state-of-the-art methods in terms of mean-squared error ($\text{MSE}=\|\hat{\beta}-\beta^{\ast}\|_2^2$). For all methods, tuning parameters are chosen based on five-fold cross-validation (in the case of GTV, we perform a three-dimensional search to find $\lambda_1,\lambda_{TV}$ and $\lambda_S$).
We consider the following estimation procedures:
\paragraph*{GTV-Esti (Our method)}Graph-based total variation (GTV) method using 
original design matrix $X\in\reals^{n\times p}$ for both covariance matrix estimation and parameter estimation. To implement GTV-Esti, we first use  $X$ to compute the estimated covariance matrix, $\hSigma$, using hard thresholding of the sample covariance matrix with a threshold is chosen by cross validation (see \citet{BicLev07} for more details). We construct the edge incidence matrix $\Gamma$ based on $\hSigma$ and then estimate $\hat{\beta}$ using \eqref{eq:est2}. 
\paragraph*{GTV-Indep (Our method)}This approach is equivalent to GTV-Esti (above), except that the side information $X_{\text{ind}}$ is used to compute the estimated covariance matrix $\hSigma$.
\paragraph*{CRL-Esti}Cluster Representative LASSO (CRL) method of \citet{buhlmann2013correlated} using $X$ for both covariate clustering and parameter estimation. To implement CRL-Esti, we first use $X$ for covariate clustering using canonical correlations in $X$ (see \citet[Algorithm 1]{buhlmann2013correlated} for more details), then the Cluster Representative LASSO is implemented based on the clusters. 
\paragraph*{CRL-Indep}This approach is equivalent to CRL-Esti (above), except that the side information $X_{\text{ind}}$ is used to improve clustering of the covariates. That is, we run CRL as before, but based on the canonical correlations computed from $X_{\text{ind}}$.
\paragraph*{LASSO}Standard LASSO \citep{Tibshirani96}.
\paragraph*{Elastic Net}Method from \citep{Zou05} which includes both an $l_1$ and an $l_2$ penalty term in order to encourage grouping strongly correlated predictors.
\paragraph*{OWL}Ordered Weighted LASSO \citep{Candes_OWL}. We set the weights for OWL corresponding to the OSCAR regularizer \citep{bondell2008simultaneous}, \ie $w_i=\lambda_1+\lambda_2(p-i)$ with $1\leq i\leq p$ and $\lambda_1,\lambda_2\geq0.$ 

\begin{figure}
	\centering     
	\subfloat[ Block Graph ($n=100$, $s = 84$, $r = 0.8$)]{\label{fig:an}\includegraphics[width=.32\linewidth]{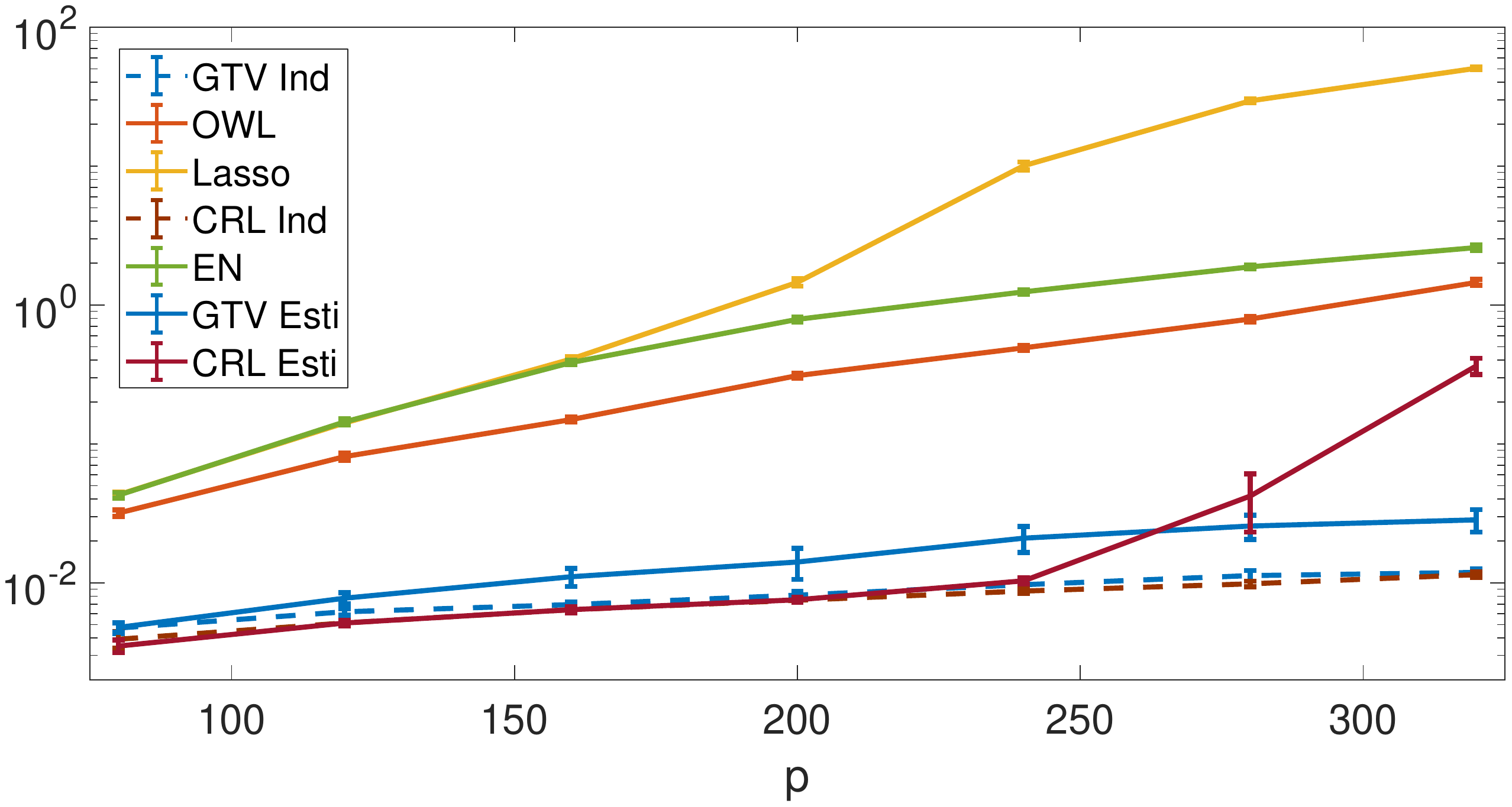}} ~
	\subfloat[ Chain Graph ($n = 100$, $s = 80$, $r = 0.4$)]{\label{fig:ap}\includegraphics[width=.32\linewidth]{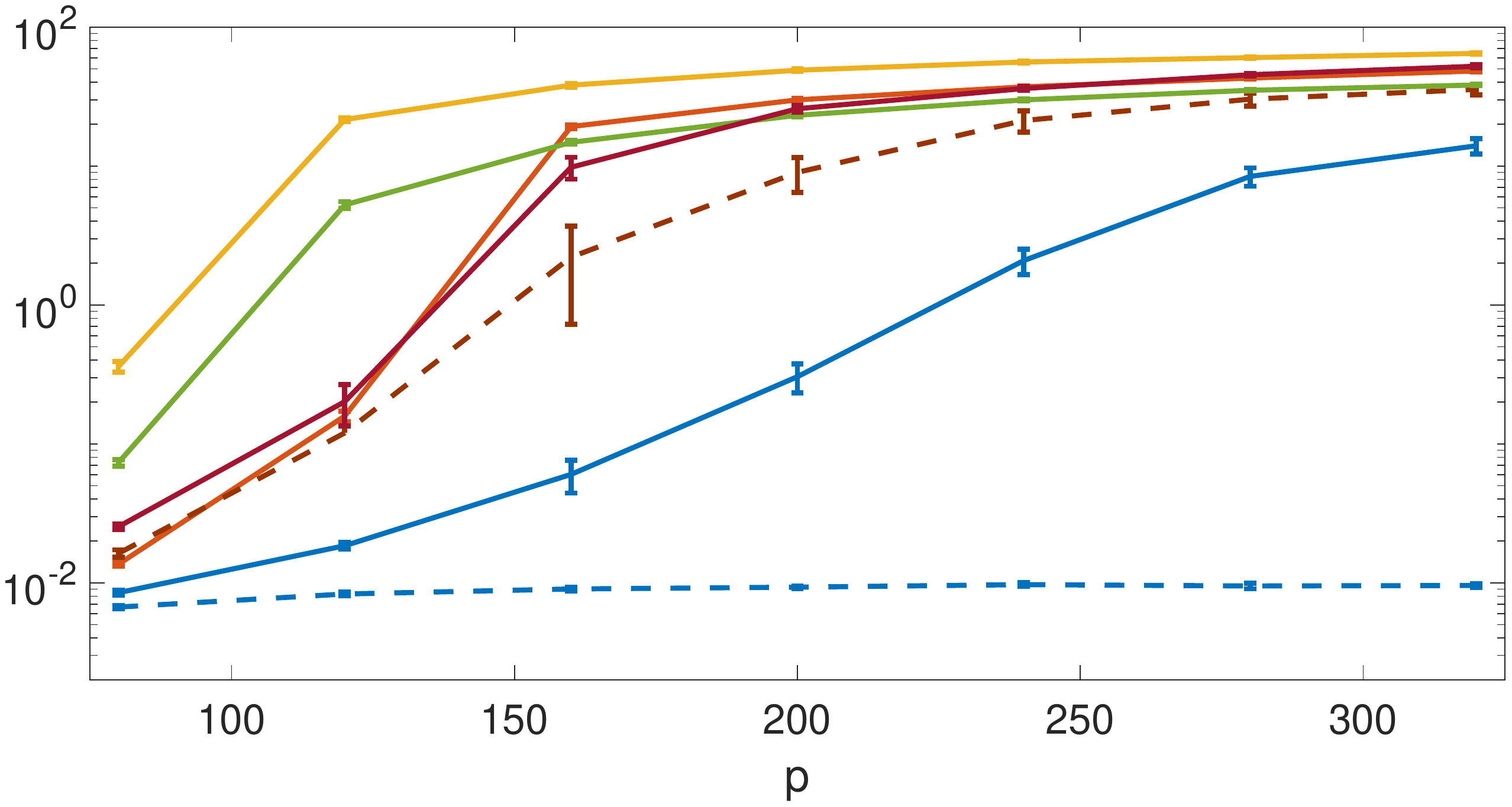}}
	~
	\subfloat[ Lattice Graph ($n = 250$, $s = 81$, $r = 0.2$)]{\label{fig:latp}\includegraphics[width=.32\linewidth]{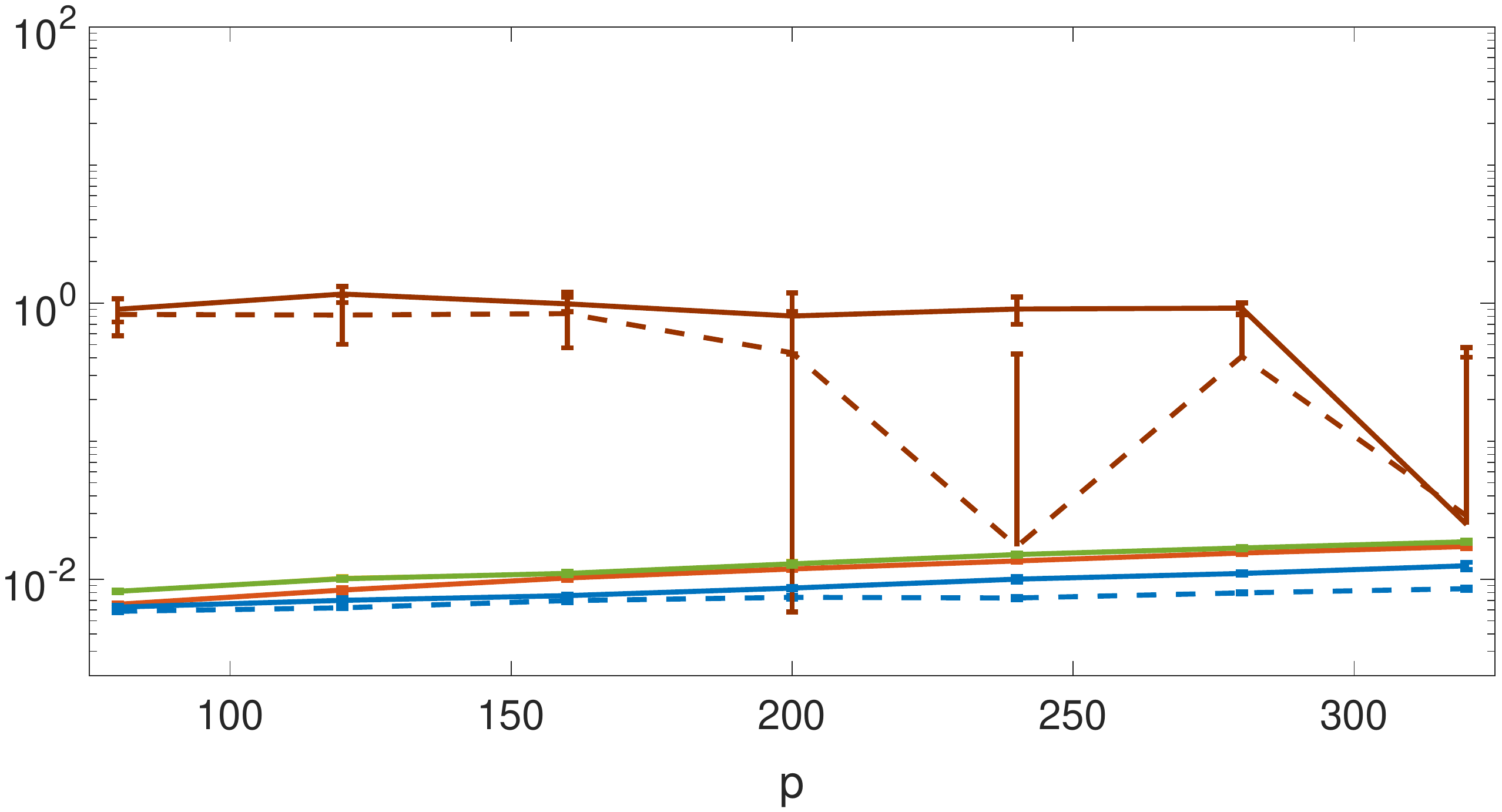}}
	\\
	\subfloat[ Block Graph ($p = 280$, $s = 84$, $r = 0.8$)]{\label{fig:bn}\includegraphics[width=.32\linewidth]{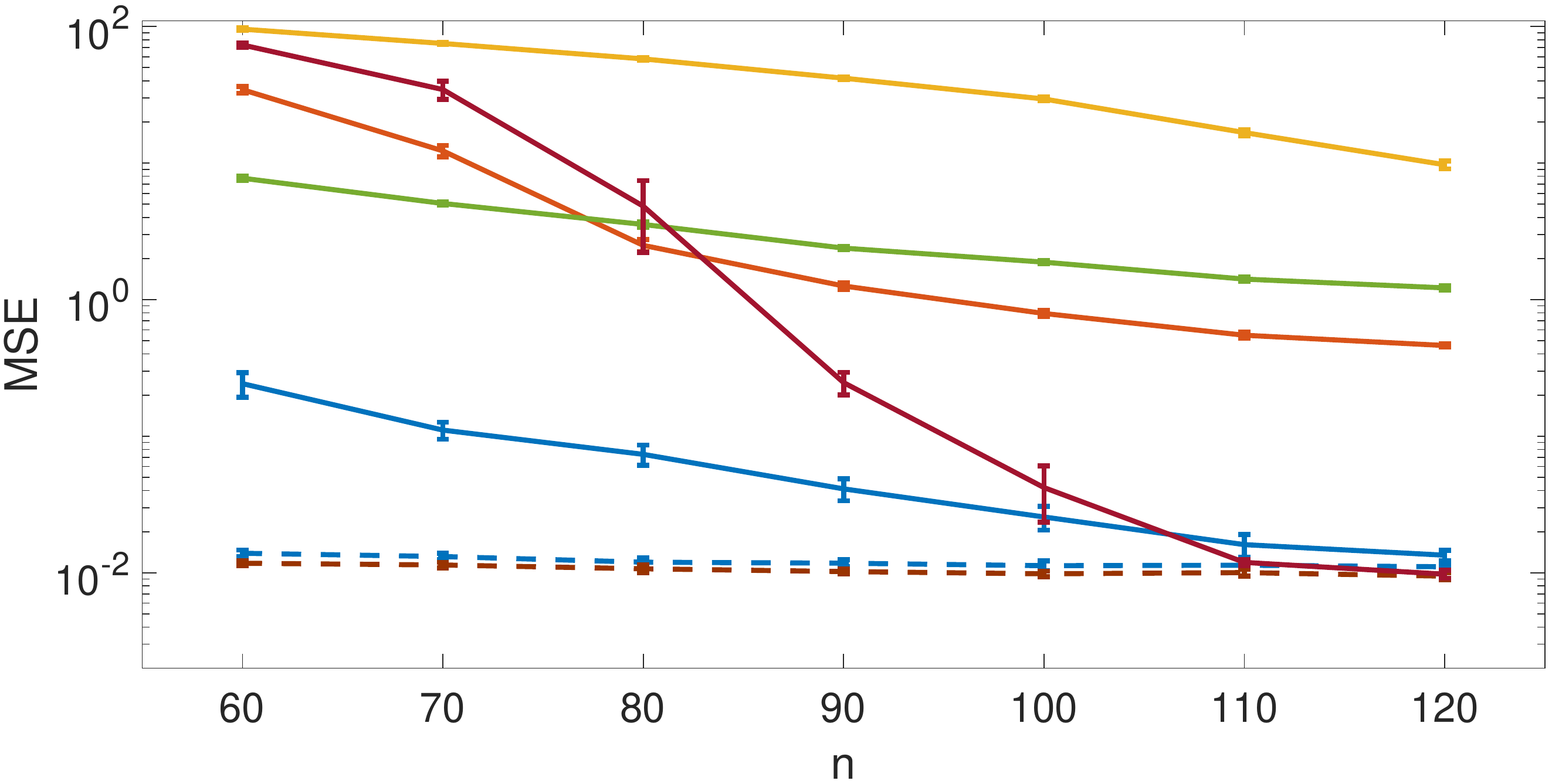}}
	~
	\subfloat[ Chain Graph ($p = 280$, s = $80$, $r = 0.4$)]{\label{fig:bp}\includegraphics[width=.32\linewidth]{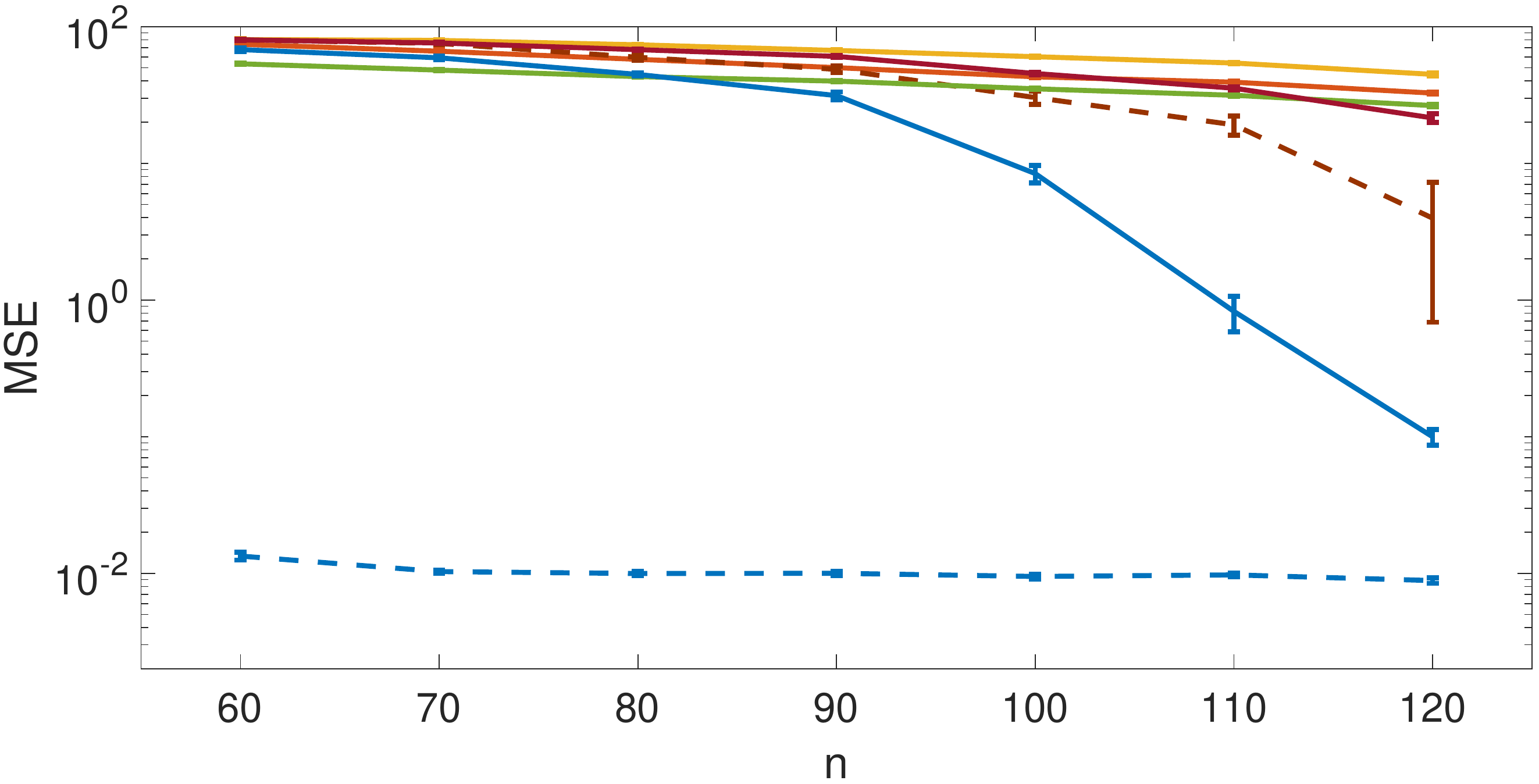}}
	~
	\subfloat[ Lattice Graph ($p = 289$, s = $81$, $r = 0.2$)]{\label{fig:latn}\includegraphics[width=.32\linewidth]{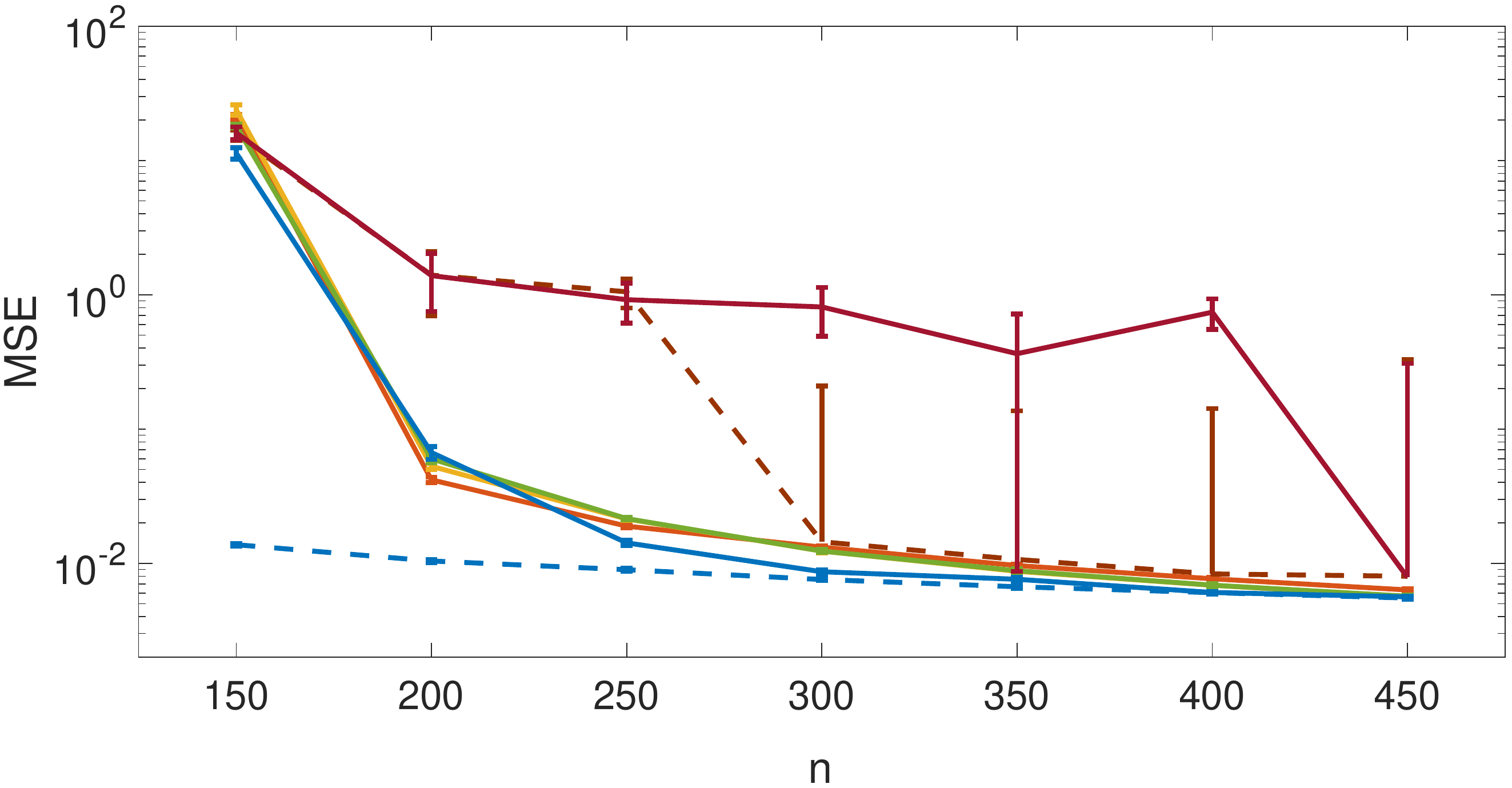}}
	\caption{MSE for varying covariance graph structures and values of $n$ and $p$. Median of 100 trials are shown, and error bars denote
the standard deviation of the median estimated using the bootstrap method with 500 resamplings on the 100 mean-squared errors.  GTV-Esti yields lower MSEs than other methods for a broad range of $n,p$.}
	\label{figure:n}
\end{figure}

We want to investigate how the mean-squared error (MSE) changes with number of observations $n$ and the number of covariates $p$.
The results are summarized in Figure~\ref{figure:n}.  We show the median MSE of 100 trials and we add error bars with the standard deviation (of the median) estimated using the bootstrap method with 500 resamplings on the 100 MSEs.  We see that over the different graph structures and values of $p,n$, GTV-Esti usually has lower MSE than CRL-Esti, OWL, Elastic Net and LASSO; if we have additional side information we can achieve better results by using GTV-Indep or CRL-Indep. We can also see that the MSE decreases as $n$ increases and MSE increases as $p$ or $s$ increases, which is consistent with our theoretical results. 

\begin{figure}
\centering     
\subfloat[Robustness to increases in $\sigma$.  An increase in $\sigma$ causes an increase in $\|\Gamma \beta^\ast\|_1$ while holding $\|\Gamma \beta^\ast\|_0$ constant. ]{\label{fig:l1}\includegraphics[width=.5\linewidth]{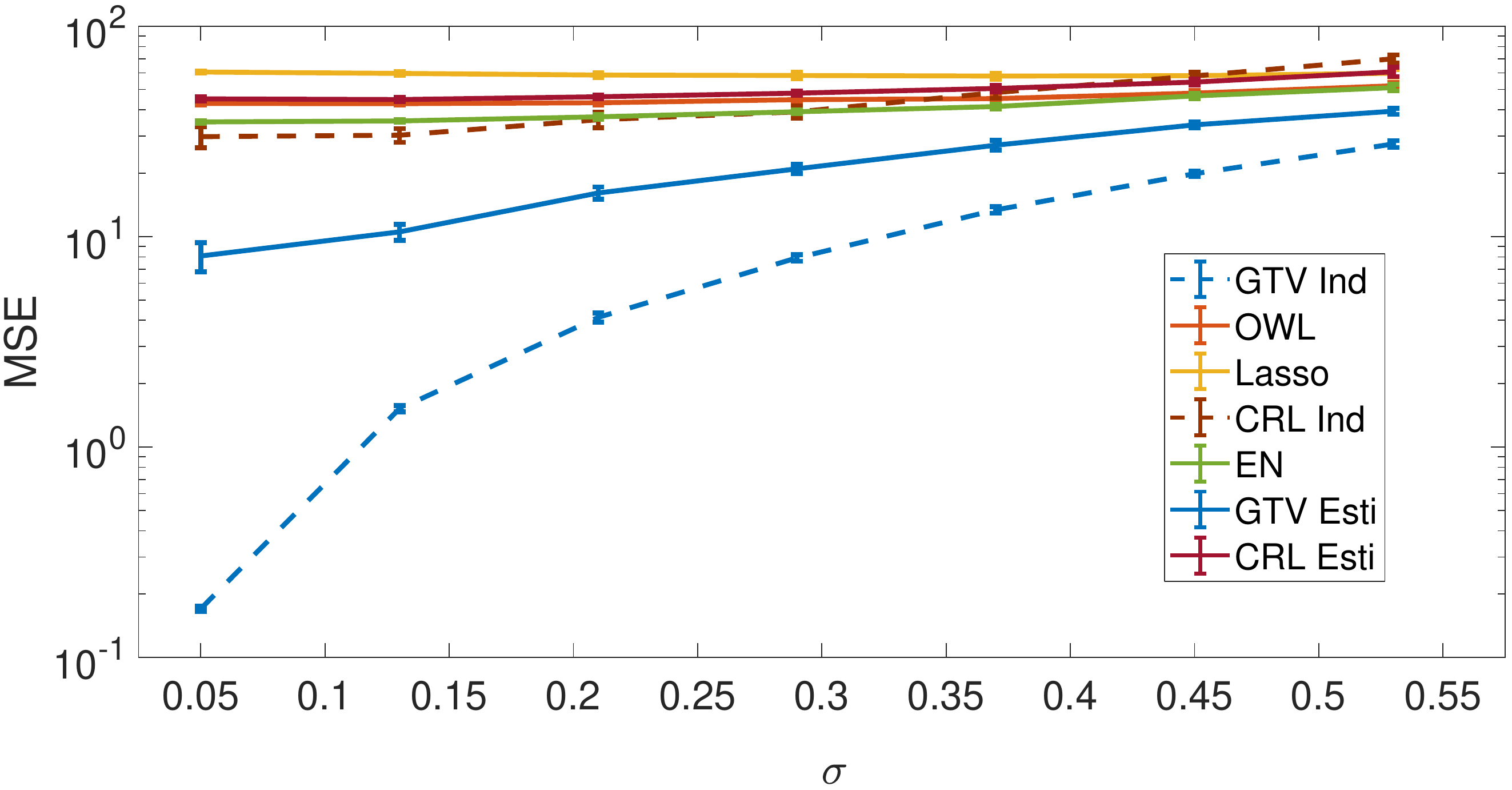}}
	\subfloat[Robustness to increases in $\|\Gamma \beta^\ast\|_0$. ]{\label{fig:l0}\includegraphics[width=.48\linewidth]{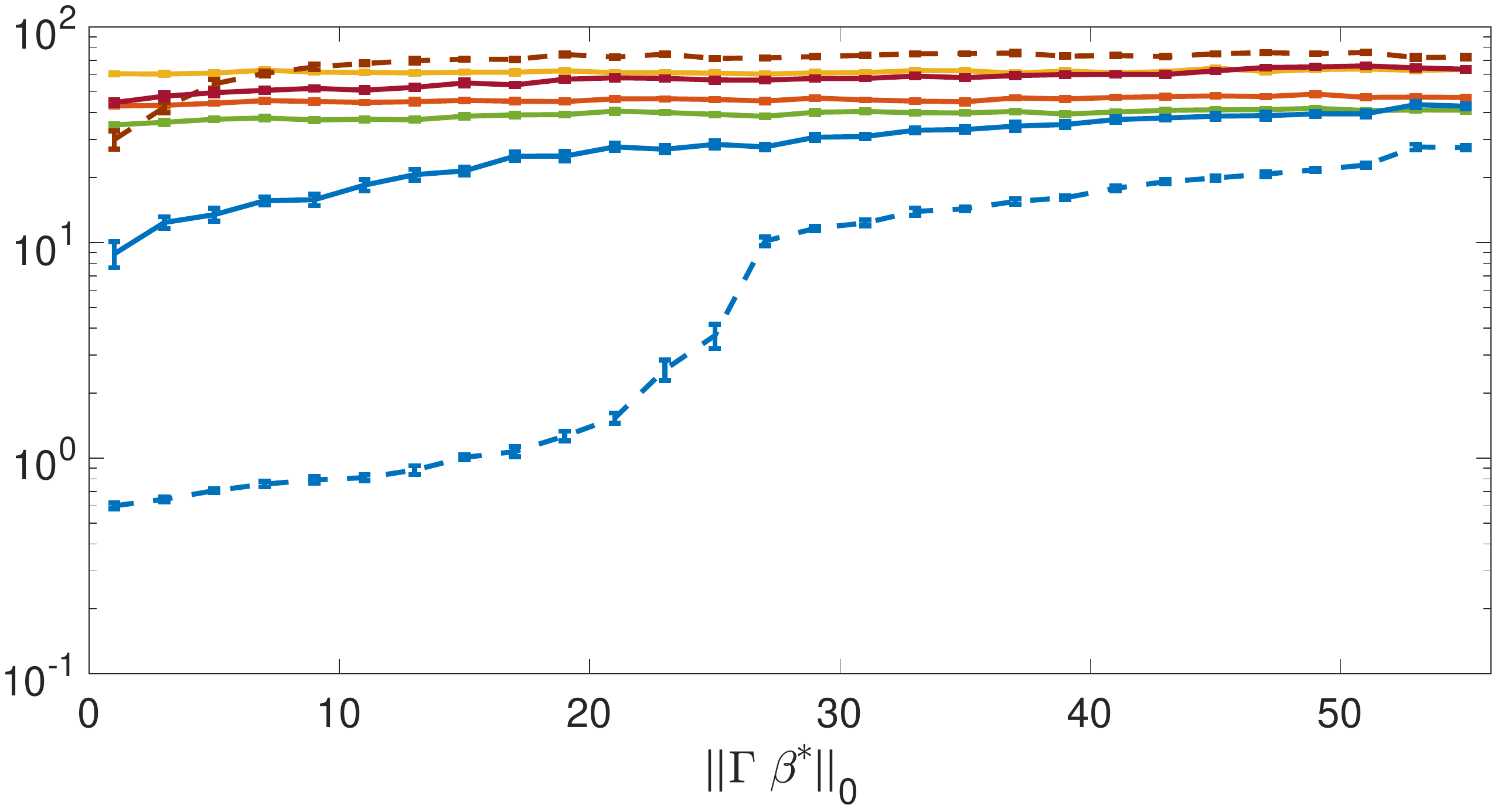}}
\caption{Chain graph (p=280, n=100, s=80 and r=.4).  On left, all active nodes are contained in one continuous block, and active nodes are chosen from ${\cal N}(1,\sigma^2)$.  On right, active nodes are separated into an increasing number of distinct block, and all active nodes are chosen from ${\cal N}(1,.01^2)$.  Plots demonstrate that GTV performs well with moderate amounts of misalignment between the graph and $\beta^\ast$.  Medians of 100 trials are shown, and error bars denote
the standard deviation of the median estimated using the bootstrap method with 500 resamplings on the 100 mean-squared errors.}  
\label{figure:robust} 
\end{figure}

We next test how the error scales with $\| \Gamma \beta^* \|_0$ and $\| \Gamma \beta^* \|_1$.  In Figure \ref{fig:l1} we take a chain graph with $p=280$ nodes and let the first $s=80$ nodes be active.  For the active nodes we set $\beta^\ast_j \sim {\cal N}(1,\sigma^2)$ for varying values of $\sigma$.  In other words we change the value of $\|\Gamma \beta^*\|_1$ while holding $\| \Gamma \beta^* \|_0$ constant.  We see that GTV is reasonably robust to increases in $\|\Gamma \beta^*\|_1$ and still performs well with high levels of noise within the active block.  

In Figure \ref{fig:l0} we again look at a chain graph with $p=280$ nodes and $s=80$ active nodes, but this time we break up the active nodes into distinct blocks.  Each active node is chosen from ${\cal N}(1,.01^2)$.  We measure MSE a function of the number of distinct blocks the active nodes are divided into.  In other words, this setting measures robustness to $l_0$ misalignment as opposed to $l_1$ misalignment.  We see that GTV performs well even when $\|\Gamma \beta^*\|_0$ is reasonably large, again suggesting that our methods are robust to moderate amounts of misalignment between the graph and $\beta^*$.

\section{Biochemistry application: Cytochrome P450 enzymes}
\label{sec:biochem}
In this section we describe an application of the proposed GTV methodology to protein thermostability data. As described in Section~\ref{sec:applications}, the thermostability data we use was provided by the Romero Lab at UW-Madison. The data contains thermostability measurements for 242 proteins in the P450 protein family. For each protein, 50 structure features were simulated via RosettaCommons \citep{Alford2017-pt} and the goal is to understand the relationship between the $50$ structural features and thermostability. Hence the design matrix $X \in \mathbb{R}^{242 \times 50}$ consists of the structural features.  The response variable $y \in \mathbb{R}^{242}$ contains the thermostability measurements. Additionally, we have side information in the form of the amino acid sequences that make each of the 242 proteins; this is used to estimate the covariance matrix amongst the structural features.

\subsection{Estimation of the covariance matrix with side information} \label{sec:biochem_esti}
One advantage of our GTV method is that side information can be incorporated to estimate the strength of correlations among features. It is a well known fact that the structure of the protein is a function of its amino acid sequence. We exploit this sequence and structure relationship and model the structural features as linear functions of sequence features. Then we use this model to obtain a better approximation of the covariance of structural features.

\begin{figure}
\centering     
\includegraphics[width=.6\linewidth, height = .35\textwidth]{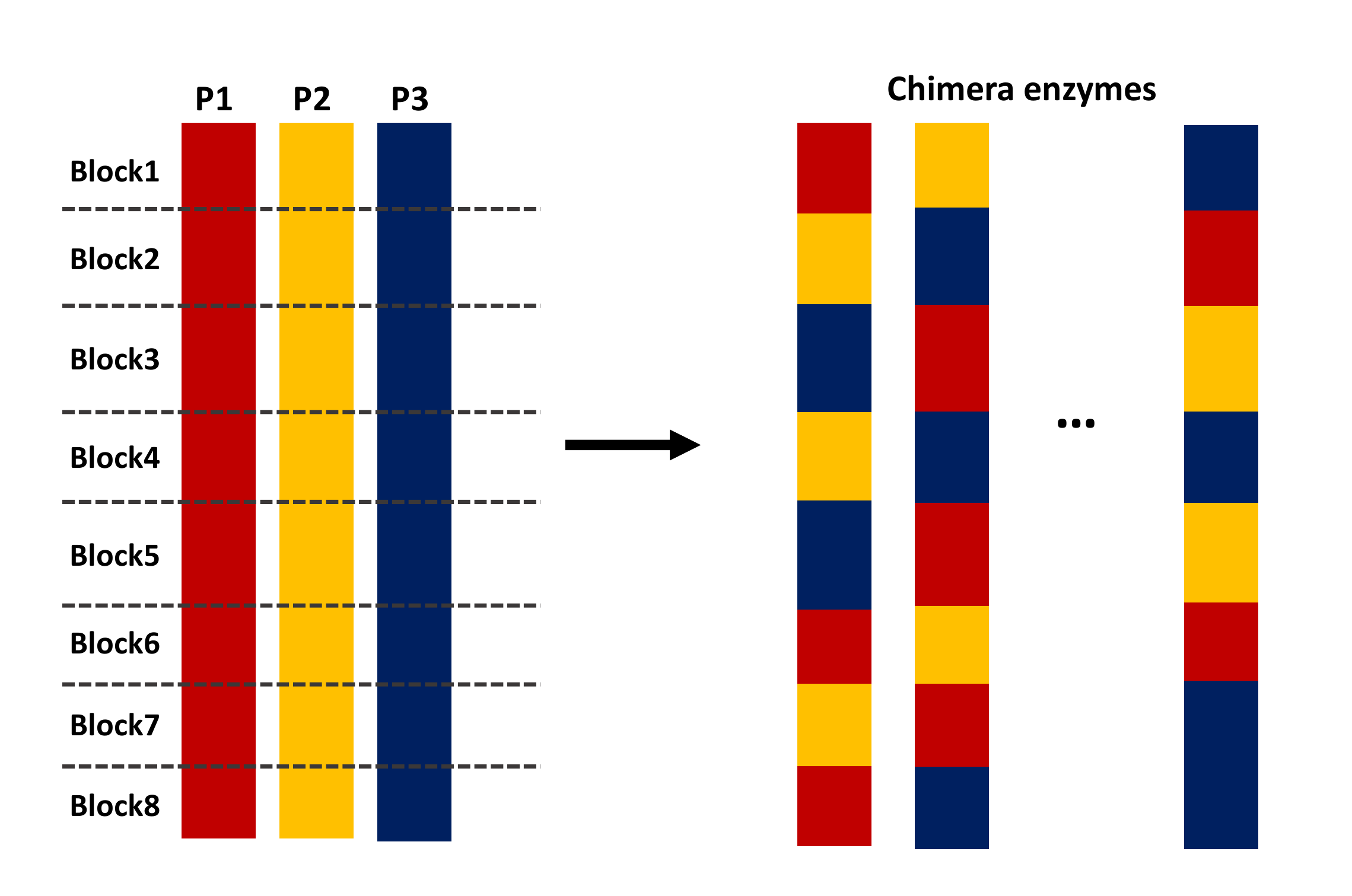}
\caption{A diagram of the process of creating Chimeras enzymes. P1, P2, and P3 are three parent proteins. They are each made up of an amino acid sequence (represented by red, yellow, or blue). There are $8$ pieces/blocks in each sequence. Chimera enzymes are made from recombining blocks from the $3$ parents. The P450 dataset we use consist of Chimeras.}  
\label{figure:protein} 
\end{figure}

The proteins were created by the recombination of $3$ other proteins. Each protein's amino acid sequence can be thought of as having $8$ pieces/blocks where each piece came from one of $3$ parent proteins (Figure \ref{figure:protein}). So the amino acid sequence can be represented as $8$ categorical features, each with $3$ categories. Each feature represents one piece of the sequence and indicates which parent that piece came from. 
We can use the one-hot encoding of these 8 categorical features to obtain $24$ binary features that represent an amino acid sequence for a protein. Because each piece comes from one of three parents, the sum of the $3$ binary features for each piece of the sequence must be $1$. So only $2$ parameters are needed for each piece of the sequence. Hence a model of the amino acid sequence has $K = 16$ parameters. 

 Hence we model $p=50$ structural features as linear functions of $K=16$ binary sequence features via a multivariate linear regression model. More concretely, we assume a linear model
\begin{align}\label{eq:S_to_X}
    X^{(i)} =  A^TS^{(i)} + \delta^{(i)}
\end{align}
where $X^{(i)} \in \mathbb{R}^{p}$ is a vector of the $i$th structural feature and $S^{(i)} \in (0,1)^K$ is the binary sequence features of the $i$th enzyme in the dataset. The matrix $A \in \mathbb{R}^{K \times p}$ is an unknown  parameter matrix which determines the relationship between $X^{(i)}$ and $S^{(i)}$, and we assume Gaussian noise 
 $\delta^{(i)} \sim \mathcal{N}(0,\sigma_\delta^2)$ independent from $S^{(i)}$ and $\epsilon^{(i)}$. 
We note that the model assumption \eqref{eq:S_to_X} amounts to assuming that the thermostability $y$ can be modeled by the sequence matrix $S$ which is of rank $K$. Although modeling $y$ directly via $S$ is possible, the results will not provide an understanding of how structural features contribute to the thermostability of a protein, which is the goal of our analysis.

Exploiting the structure of $X$ in \eqref{eq:S_to_X}, we estimate the covariance matrix of X given sequence $S$ as
$$ \widehat{\Sigma}_{\text{ind}}:= \widehat{\text{Var}}( \mathbb{E}[X^{(i)}|S^{(i)}])  = \widehat{A}^T \widehat{\text{Var}}(S^{(i)})\widehat{A} = \widehat{A}^T \widehat{\Sigma}_s\widehat{A},$$
where $\widehat{\Sigma}_s$ is an empirical covariance matrix of $(S^{(i)})_{i=1}^n$ and  $\widehat{A}$ is the LSE of $A$, i.e. 
$$\widehat{A}= \arg \min_{A \in \mathbb{R}^{K\times p}} \|\mathbf{X} - \mathbf{S}A\|_F^2.$$
We note that the dimensions of $A$ and $\Sigma_s$ are $K$ by $p$ and $K$ by $K$, respectively. Thus we reduce the estimation problem of a $p$ by $p$ matrix to a smaller problem, with $K=16$ being much less than $p=50$.

\subsection{Results}

We compare our GTV method (with and without side information) with Ordered Weighted LASSO (OWL), Cluster Representative LASSO (CRL), and the standard LASSO (LASSO), and the Elastic Net (EN) method. For all models, the tuning parameters were selected via five-fold cross validation on the training set. For OWL, the weights were set corresponding to the OSCAR regularizer. 

To compare the performance of the five methods on the real P450 data, we considered two performance criteria: prediction accuracy and stability of estimated coefficients. To measure stability between estimated coefficients, we considered following two criteria:
\begin{enumerate}
    \item $\text{Cor}(\hat{\beta}_i,\hat{\beta}_j)$ where $\hat{\beta}_i$ and $\hat{\beta}_j$ are estimates from two different fittings for the same model.
    \item  Tanimoto Distance \citep{kalousis2007stability}: 
$$ D(i,j):= 1 - \frac{|\supp(\hat\beta_i)| + |\supp(\hat\beta_j)| - 2|\supp(\hat\beta_i) \cap \supp(\hat\beta_j)|}{|\supp(\hat\beta_i)| + |\supp(\hat\beta_j)| - |\supp(\hat\beta_i) \cap \supp(\hat\beta_j)|}$$ where $\supp$ refers to the support set.
\end{enumerate}
For prediction accuracy, we use $10$-fold cross validation. We trained the six models on each training set and evaluated the prediction performances on the test set. On the other hand, stability measures were calculated by spliting the entire P450 dataset into ten non-overlapping subsamples and fitting the six models using each of the subsamples. 

Table \ref{tab:tab1} summarizes prediction accuracy. The result for EN is excluded since the tuning parameter for the $l_2$ penalty $\lambda_S$ was chosen to be $0$ in all cross-validation folds, and the result for EN is the same as LASSO. From the Table \ref{tab:tab1}, we see that GTV Esti has the highest accuracy. GTV Ind (GTV with side information) is the next most accurate. CRL Ind and CRL Esti show very bad prediction performance. CRL is expected to perform badly in the case where variables are not grouped into tight clusters or coefficients within a group have opposite signs and their sum is close to zero \cite{buhlmann2013correlated}. In our application, in most cross-validation folds Algorithm 1 in \cite{buhlmann2013correlated} resulted in one huge cluster in the case of CRL Esti, whose member features do not have similar effects on the response variable. We observed similar phenomenon in the case of CRL Ind, although to a lesser extent than the CRL Esti, where we observed one cluster with nine features with opposite effects and the remaining clusters are of size 1. As a result, both CRL methods demonstrated very poor prediction results.

\begin{table}[H]
    \centering
   \begin{tabular}{rrrrrrr}
  \hline
 & GTV Ind & GTV Esti & LASSO & CRL Ind & CRL Esti & OWL \\ 
  \hline
MSE & 5.10 & 5.08 & 5.11 & 13.78 & 31.21 & 5.35 \\ 
   \hline
\end{tabular}
    \caption{The average Mean Squared Error for each model on the P450 dataset.}
    \label{tab:tab1}
\end{table}

\begin{figure}[H]
\centering
\subfloat[Correlation stability]{\label{fig:biochem1}\includegraphics[width=.5\linewidth]{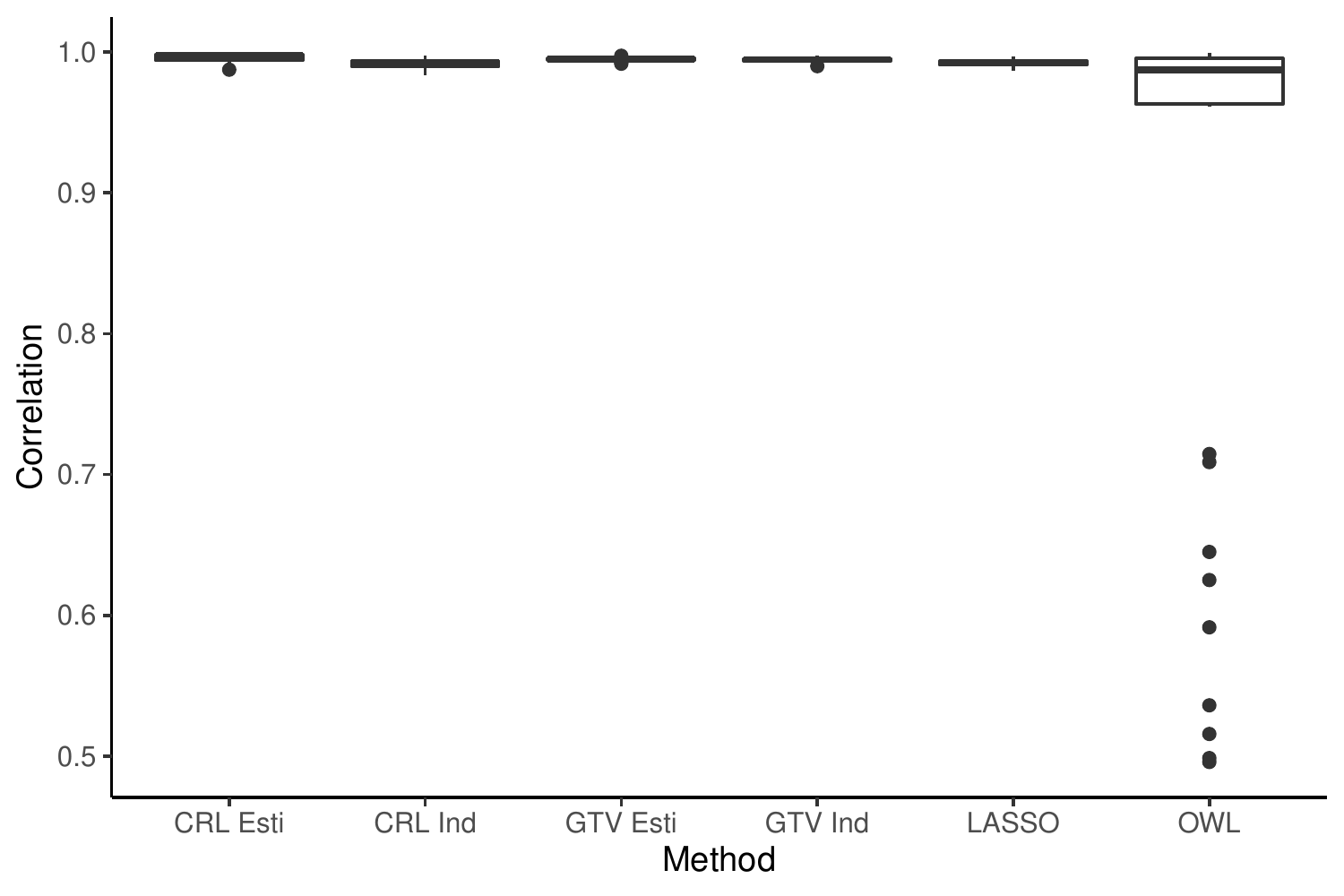}}
	\subfloat[Tanimoto distance]{\label{fig:biochem2}\includegraphics[width=.48\linewidth]{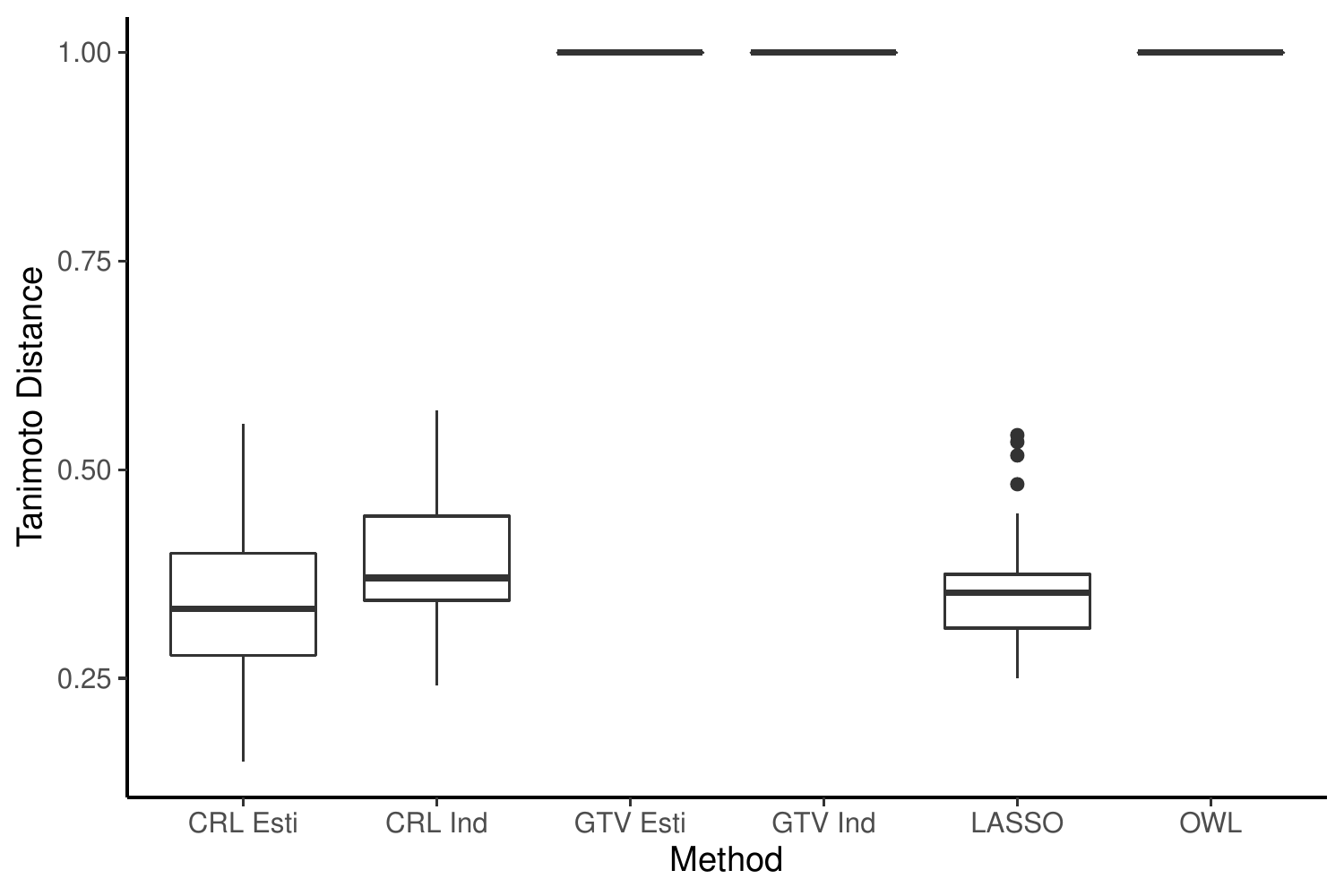}}
\caption{Box Plots of the two stability measures of each model on the P450 dataset. Correlation and Tanimoto distance were calculated between 10 different fittings for each model, leading to 45 measurements per model for each kind of the stability measrue.}  
\label{figure:biochem_plots} 
\end{figure}

Figure \ref{figure:biochem_plots} demonstrates the correlation and variable selection stability. GTV Ind and GTV Esti show the most stable performances overall. In terms of correlations, all five methods generated highly correlated coefficients across different fits, except OWL which had a few outliers. For variable selection stability, both GTV methods and OWL produced the same support sets in all fits. On the contrary, the support sets from LASSO and both CRL methods greatly varied across fits. Only about 30\% of the support sets overlap between any pair of fits. It appears that relatively strong correlation in the design but the lack of tightly grouped clusters contributed to the instability of clustering and support recovery in LASSO and CRL methods.

\section{Conclusion}
\label{sec:conclusion}
This paper describes a new graph-based regularization method for high-dimensional regression with highly-correlated designs and alignment between the covariance and regression coefficients. The structure of the estimator leverages ideas behind the Elastic Net \citep{Zou05}, the Fused LASSO \citep{tibshirani2005sparsity}, the edge LASSO \citep{sharpnack2012sparsistency}, trend filtering on graphs \citep{wang2016trend}, and graph total variation \citep{shuman2013emerging,hutter2016optimal}. Under our model, the graph corresponding to the covariance structure of the covariates also provides prior information about the similarities among elements in the regression weights. Thus this graph allows us to effectively pre-condition our design matrix and regularize regression weights to promote alignment with the covariance structure of the problem. We are able to provide mean-squared error bounds in settings where covariates are highly dependent, provided there is alignment between the $ \beta^{\ast}$ and graph. We also demonstrate in both simulations and a biochemistry application superior performance of our method compared to LASSO, Elastic Net and CRL. The proposed framework allows us to leverage correlation structure jointly with the response variable $y$, in contrast to previous work that depended upon clustering covariates independent of the responses. In settings where there exist very strong clusters (like the block graph studied above), clustering with and without responses yield similar results. However, when correlations are too weak to reveal strong clusters and yet too strong for the LASSO alone to be effective (like with the chain and lattice graphs studied above), the implicit response-based clustering associated with our method can yield significant performance benefits. The results in this paper suggest several exciting avenues for future exploration, including more refined performance bounds for additional classes of graphs and more extensive evaluations on real-world data.

\section{Acknowledgement}
The authors would like to thank \href{http://stat.columbia.edu/department-directory/name/ian-kinsella/}{Ian Kinsella} for helpful discussions, suggestions, and preliminary experiments. The authors would also like to thank \href{http://http://www.romerolab.org/}{Philip Romero} and {Jerry Duan} for creating and providing access to the biochemistry example.

\bibliographystyle{chicago}

\bibliography{Biblio_GroupLasso}
\newpage 
\appendix

\section{Covariance estimation}
\label{app:cov}
Assume we observe  a collection of $m$ i.i.d.\ unlabeled feature vectors $(\check X^{(i)})_{i=1}^m$ that may be independent of the design features $(X^{(i)})_{i=1}^n$ with $\check X^{(i)} \sim \cN(\textbf{0},\Sigma_{p \times p})$. 
In this case, we need to estimate $\Sigma$ based on $(\check X^{(i)})_{i=1}^m$, and there is a large literature on high-dimensional covariance estimation in high dimensions under different structural assumptions (see \citet{BicLev06,BicLev07,cai2011adaptive,cai2016,donoho2013optimal,BaiSil06}). As an example, we consider estimators based on thresholding the sample covariance matrix under block and sparsity assumptions developed by~\citet{BicLev07}. 
\subsection{Sparse covariance matrix}
To be specific, suppose the true covariance matrix $\Sigma$ belongs to the following class:
$$\Omega(q,c_0(p),M)=\left\{\Sigma:\Sigma_{j,j}\leq M, \sum_{k=1}^p|\Sigma_{j,k}|^q\leq c_0(p), \text{for all}~j\right\},$$
where $0\leq q<1$, $c_0(p)$ is a constant that depends on $p$ and $M$ is an absolute constant. Then \citet[Theorem 1]{BicLev07} show that if we define the thresholded covariance matrix $\hSigma_{j,k}=S_{j,k}\mathbbm{1}(|S_{j,k}|\geq t)$ for all $1 \leq j,k\leq p$ where $S$ is the sample covariance matrix and $t=O\left(\sqrt{\frac{\log p}{m}}\right)$, then
$$\|\hSigma-\Sigma\|_{1,1}=O_P\left(c_0(p)M\left(\frac{\log p}{m}\right)^{\frac{1-q}{2}}\right).$$
Though the original error bound result for $\hSigma-\Sigma$ in \citet{BicLev07} was shown in operator norm,  they bounded $\|\hSigma-\Sigma\|_{1,1}$ in the proof. In particular if $q=0$ and $c_0(p)\leq s$ denotes the sparsity level, $$\|\hSigma-\Sigma\|_{1,1}=O_P\left(s\sqrt{\frac{\log p}{m}}\right),$$ meaning if $m=O(s^2\log p),$ Assumption \ref{as:estimatecov} is satisfied.
\subsection{Block covariance matrix}
On the other hand, if the covariance matrix $\Sigma$ is not sparse but rather block-structured, we can use an alternative bound developed in \citet{BicLev07}. If $\Sigma$ has $K$ identical blocks where each block has $p/K$ elements, we can ensure Assumptions~\ref{as:tvlambdasigma} and~\ref{as:sigmanormalization} are satisfied if $\Sigma_{j,k}=O(\frac{K}{p})$ for each non-zero $\Sigma_{j,k}$. Then if we choose $\hSigma$ to be the sample covariance matrix, \citet{BicLev07} prove that 
\begin{eqnarray}
\label{eq:OPmax}
\max_{j,k}\left|\frac{\hSigma_{j,k}-\Sigma_{j,k}}{K/p}\right| = O_P\left(\sqrt{\frac{\log p}{m}}\right)
\end{eqnarray}
since now we have $\frac{\Sigma_{j,j}}{K/p}=O(1)$ for $1\leq j\leq p$. Thus by (\ref{eq:OPmax}) we know that 
\begin{eqnarray}
\label{eq:OPmax2}
\max_{j,k}|\hSigma_{j,k}-\Sigma_{j,k}| = O_P\left(\frac{K}{p}\sqrt{\frac{\log p}{m}}\right)
\end{eqnarray}
and
$$\|\hSigma-\Sigma\|_{1,1}=\max_{1\leq j\leq p}\sum_{k=1}^p|\hSigma_{j,k}-\Sigma_{j,k}|=O_P\left(K\sqrt{\frac{\log p}{m}}\right)$$
by (\ref{eq:OPmax2}), so that when $m=O(K^2\log p)$  Assumption \ref{as:estimatecov} is satisfied.

\section{Proof of Theorem \ref{theo:tvMain}}
\label{sec:Proof}
Much of our analysis follows standard steps for analysis of regularized M-estimators (see~\citet{BiRiTsy08,Neg10,vandeGeer}), but we face two additional challenges not present in these works. First, since the regularization penalty in Equation \eqref{eq:est2} is $\|\tilde{\Gamma}\beta\|_1$ rather than $\|\beta\|_1$ we need to deal with error terms involving $\tX\tilde{\Gamma}^{\dagger}$ instead of $\tX$.  To  address this we incorporate techniques from~\citet{hutter2016optimal} and \citet{raskutti2015convex}.  Second, we need to establish a restricted eigenvalue condition for $\tX$ rather than $X$.  We incorporate techniques from~\citet{RasWaiYu10b} in order to accomplish this.

Based on the optimization problem~\eqref{eq:est2}, by the definition of
$\hat{\beta}$ and the basic inequality,
\begin{eqnarray*}
\frac{1}{n}\|\ty-\tX\hat{\beta}\|_2^2 + \lone \|\tPen\hat{\beta}\|_1\leq\frac{1}{n}\|\ty-\tX\bstar\|_2^2 + \lone \|\tPen \bstar\|_1.
\end{eqnarray*}
By simple re-arrangement,
\begin{eqnarray*}
\frac{1}{n}\|\tX(\hat{\beta}-\beta^{\ast})\|_2^2\leq\frac{2}{n}(\ty-\tX\bstar)^{\top} \tX(\hat{\beta}-\bstar)+\lone (\|\tPen\bstar\|_1-\|\tPen\hat{\beta}\|_1).
\end{eqnarray*}
For the remainder of the proof let $\Delta := \hat{\beta}-\bstar$. Then
\begin{eqnarray*}
\frac{1}{n}\|\tX \Delta\|_2^2\leq\frac{2}{n}(\ty-\tX\bstar)^{\top} \tX \Delta+\lone (\|\tPen\bstar\|_1-\|\tPen\hat{\beta}\|_1).
\end{eqnarray*}

First we control the term $(\ty-\tX\bstar)^{\top} \tX \Delta$. Using basic algebra,
\begin{eqnarray*}
(\ty-\tX\bstar)^{\top} \tX \Delta &=& \epsilon^{\top} X \Delta -n\lrid{\bstar}^{\top} \Gamma^{\top} \Gamma \Delta.
\end{eqnarray*}
Since $\tPen^{\dagger} \tPen = I_{p \times p}$,
where $\tPen^{\dagger}$ is the pseudo-inverse of $\tPen$. Therefore
\begin{eqnarray*}
\epsilon^{\top} X \Delta &=& \epsilon^{\top} X\tPen^{\dagger}\tPen\Delta\\
&\leq& \|(X\tPen^{\dagger})^{\top} \epsilon\|_{\infty}\|\tPen\Delta\|_1.
\end{eqnarray*}
We next bound $n\lrid{\bstar}^{\top} \Gamma^{\top} \Gamma \Delta$ by
\begin{eqnarray*}
n\lrid{\bstar}^{\top} \Gamma^{\top} \Gamma\Delta&\leq& n\lrid\|\Gamma^{\top} \Gamma\bstar\|_{\infty}\|\Delta \|_1\\
& \leq & n\lrid\|L\bstar\|_{\infty}\|\tPen \Delta \|_1\\
&\leq & n \frac{\lone}{8} \|\tPen \Delta\|_1,
\end{eqnarray*}
where the last inequality follows from the constraint that $\lone\geq 8 \lrid \|L\btrue\|_\infty$.  
Now recall the constraint $\lone\geq48\rho \sigma \sqrt{\frac{{c_u}\log p}{n}}$, the following lemma shows that with high probability we have $\lambda_1\geq\frac{8}{n}\|(X\tilde{\Gamma}^{\dagger})^{\top}\epsilon\|_{\infty}$.

\blems
\label{lemma:gwidthcompare}
Suppose we have $\lambda_1\geq48\rho \sigma \sqrt{\frac{{c_u}\log p}{n}}$. Then with probability at least $1-\frac{C_1}{p}$,
\begin{eqnarray*}
 \lambda_1\geq\frac{8}{n}\|(X\tilde{\Gamma}^{\dagger})^{\top}\epsilon\|_{\infty}
 \end{eqnarray*}
for absolute constant $C_1>0$.
\elems
Combining the constraints for $\lambda_1$ with the inequalities above,
\begin{eqnarray*}
\frac{2}{n}(\ty-\tX\bstar)^{\top} \tX \Delta &\leq&\frac{\lone}{4}\|\tPen \Delta\|_1 + \frac{\lone}{4}\|\tPen \Delta\|_1 
= \frac{\lone}{2}\|\tPen \Delta\|_1.
\end{eqnarray*}
Putting these pieces together we have
\begin{eqnarray}
\label{eq:tv4}
\frac{1}{n}\|\tX \Delta\|_2^2 &\leq& \frac{\lone}{2}(\|\tPen \Delta \|_1+2\|\tPen\bstar\|_1-2\|\tPen\hat{\beta}\|_1).
\end{eqnarray}
Furthermore by the triangle inequality and the fact that $\frac{1}{n}\|\tX \Delta\|_2^2 \geq 0$ we have
\begin{eqnarray*}
0 \leq \|\tPen(\hat{\beta}-\bstar)\|_1+2\|\tPen\bstar\|_1-2\|\tPen\hat{\beta}\|_1&\leq&3\|(\tPen \Delta)_T\|_1-\|(\tPen \Delta)_{T^c}\|_1
+ 4\|(\tPen\bstar)_{T^c}\|_1.
\end{eqnarray*}
Therefore $\Delta$ lies in the translated cone
\begin{eqnarray}
\label{eq:tvcone}
\mathcal{C} := \{v: \|(\tPen v)_{T^c}\|_1 &\leq& 3\|(\tPen v)_T\|_1+4\|(\tPen\bstar)_{T^c}\|_1 \}.
\end{eqnarray}
Moreover by the definition of $k_T$ we have
\begin{eqnarray*}
\|(\tPen \Delta)_T\|_1\leq\frac{\sqrt{|T|}\|\Delta\|_2}{k_T};
\end{eqnarray*}
 from (\ref{eq:tv4}) we have
\begin{eqnarray}
\label{eq:tv5}
\frac{1}{2n}\|\tX \Delta\|_2^2\leq\lone \|(\tPen\bstar)_{T^c}\|_1 +\frac{3\lone }{4k_T}\sqrt{|T|}\|\Delta\|_2.
\end{eqnarray}

\subsection{Restricted Eigenvalue Condition}
From (\ref{eq:tv4}) and (\ref{eq:tvcone}) we need to lower bound 
\begin{eqnarray*}
\frac{\|\tX \Delta\|_2^2}{n}=\Delta^{\top} \left(\frac{X^{\top} X}{n}+\lrid L\right)\Delta,
\end{eqnarray*}
for all $\Delta$ belonging to the cone $\mathcal{C}$ defined in (\ref{eq:tvcone}). The result is stated in the following lemma:

\blems
\label{lemma:tvrec}
For all $\Delta$ belonging to the cone defined in (\ref{eq:tvcone})
if we have
\begin{eqnarray}
\label{eq:ltvupper}
\lone\leq c_2\sqrt{\frac{\lambda_{\min}(\Sigma+\lrid L)}{|T|}}k_T,
\end{eqnarray}
then 
\begin{eqnarray}
\label{eq:tv6}
\Delta^{\top} (\frac{X^{\top} X}{n}+\lrid L)\Delta\geq c_1\lambda_{\min}(\Sigma+\lrid L)\|\Delta\|_2^2-c_3\lone^2\|(\tPen\bstar)_{T^c}\|_1^2
\end{eqnarray}
holds with probability at least $1-c_4\exp(-c_5n)$, where $c_i>0$ for $i=1,...,5$ are positive constants.
\elems
The proof for this lemma is is based on a technique used in~\citet{RasWaiYu10b}.
\subsection{Final Part for Proof}
From (\ref{eq:tv5}) and (\ref{eq:tv6}),
\begin{eqnarray*}
c_1\lambda_{\min}(\Sigma+\lrid L)\|\Delta\|_2^2&-&c_3\lone^2\|(\tPen\bstar)_{T^c}\|_1^2\leq2\lone \|(\tPen\bstar)_{T^c}\|_1 +\frac{3\lone }{2k_T}\sqrt{|T|}\|\Delta\|_2,
\end{eqnarray*}
which is a quadratic inequality involving $\|\Delta\|_2$ as follows:
\begin{eqnarray*}
a\|\Delta\|_2^2-b\|\Delta\|_2-c\leq0
\end{eqnarray*}
with
\begin{eqnarray*}
a &=& 1,\\
b &=& \frac{3\lone\sqrt{|T|}}{2c_1k_T\lambda_{\min}(\Sigma+\lrid L)},\\
c &=& \frac{1}{c_1\lambda_{\min}(\Sigma+\lrid L)}(2\lone\|(\tPen\bstar)_{T^c}\|_1+c_3\lone^2\|(\tPen\bstar)_{T^c}\|_1^2).
\end{eqnarray*}
By solving this quadratic inequality,
\begin{eqnarray*}
\|\Delta\|_2^2\leq4\max\{b^2,|c|\}.
\end{eqnarray*}
Therefore these exists a positive constant $C_u$ such that 
\begin{eqnarray*}	
	\|\hat{\beta}-\bstar\|_2^2\leq C_u\max\left\{\frac{\lone^2|T|}{k_T^2\lambda_{\min}^2(\Sigma+\lrid L)},\frac{\lone \|(\tPen\bstar)_{T^c}\|_1+\lone^2 \|(\tPen\bstar)_{T^c}\|^2_1}{\lambda_{\min}(\Sigma+\lrid L)}\right\}.
\end{eqnarray*}
Note that the above inequality is true for all $T$, thus
\begin{eqnarray*}	
	\|\hat{\beta}-\bstar\|_2^2\leq C_u\min_{T}\max\left\{\frac{\lone^2|T|}{k_T^2\lambda_{\min}^2(\Sigma+\lrid L)},\frac{\lone \|(\tPen\bstar)_{T^c}\|_1+\lone^2 \|(\tPen\bstar)_{T^c}\|^2_1}{\lambda_{\min}(\Sigma+\lrid L)}\right\}.
\end{eqnarray*}
This completes the proof.

\section{Proof of Theorem \ref{theo:tvMain2}}
The upper bound result $\|\hat{\beta}-\beta^{\ast}\|_2^2$ stated in Theorem \ref{theo:tvMain} holds for all choices of $T$. If we choose $T=\text{supp}(\tPen\beta^{\ast})$ then $\|(\tPen\beta^{\ast})_{T^c}\|_1=0$ and by Lemma \ref{lemma:kT},
\begin{eqnarray*}
k_T^{-1}\leq\frac{\ltv\sqrt{2\|\hSigma\|_{1,1}\|\Gamma\beta^{\ast}\|_0}+\sqrt{\|\beta^{\ast}\|_0}}{\sqrt{\|\Gamma\beta^{\ast}\|_0+\|\beta^{\ast}\|_0}}.
\end{eqnarray*}
Then by Theorem \ref{theo:tvMain} we have 
\begin{eqnarray}
\label{eq:Tchoice1}
\|\hat{\beta}-\beta^{\ast}\|_2^2\leq\frac{2C_u}{\lambda_{\min}^2(\Sigma+\lrid L)}(\lambda_1^2\|\beta^{\ast}\|_0+2\lambda_1^2\ltv^2\|\hSigma\|_{1,1}\|\Gamma\beta^{\ast}\|_0).
\end{eqnarray}
On the other hand if we choose $T=\text{supp}(\beta^{\ast})$, $\|(\tPen\beta^{\ast})_{T^c}\|_1=\ltv\|\Gamma\beta^{\ast}\|_1$ and by Lemma \ref{lemma:kT}, $k_T^{-1}\leq1$. Thus if $\lambda_1\ltv\|\Gamma\beta^{\ast}\|_1\leq1$ by Theorem \ref{theo:tvMain}
\begin{eqnarray}
\label{eq:Tchoice2}
\|\hat{\beta}-\beta^{\ast}\|_2^2\leq C_u\left(\frac{\lambda_1^2\|\beta^{\ast}\|_0}{\lambda_{\min}^2(\Sigma+\lrid L)}+\frac{2\lambda_1\ltv\|\Gamma\beta^{\ast}\|_1}{\lambda_{\min}(\Sigma+\lrid L)}\right).
\end{eqnarray}
Theorem \ref{theo:tvMain2} follows by combining (\ref{eq:Tchoice1}) and (\ref{eq:Tchoice2}) and taking the minimum over these two choices of $T$.

\section{Prediction Error Bounds}
\label{sec:prediction_error}
In this section we observe that the proofs of Theorems \ref{theo:tvMain} and \ref{theo:tvMain2} also give rise to bounds on the prediction error of our estimator.  Starting from Equation \eqref{eq:tv5} in the proof of Theorem \ref{theo:tvMain}, 
\begin{eqnarray}
\frac{1}{2n}\|\tX \Delta\|_2^2\leq\lone \|(\tPen\bstar)_{T^c}\|_1 +\frac{3\lone }{4k_T}\sqrt{|T|}\|\Delta\|_2.
\end{eqnarray}
The Restricted Eigenvalue condition in Lemma \ref{lemma:tvrec} gives that 
$$\| \Delta \|_2 \leq c\sqrt{\lambda_{\min}(\Sigma+\lambda_SL)} \frac{1}{\sqrt{n}}\|X\Delta\|_2.$$
Thus 
\begin{eqnarray}
\frac{1}{2n}\|\tX \Delta\|_2^2-\frac{3c\lone }{4k_T\sqrt{n}}\sqrt{|T|\lambda_{\min}(\Sigma+\lambda_SL)}\|X\Delta\|_2-\lone \|(\tPen\bstar)_{T^c}\|_1 \leq 0 .
\end{eqnarray}
As in the proof of Theorem \ref{theo:tvMain}, we can solve this quadratic inequality to conclude
\btheos[Theorem \ref{theo:tvMain} for Prediction Error] Suppose the conditions of Theorem \ref{theo:tvMain} hold.  Then with probability at least $1-\frac{C_1}{p}$ we have
$$\frac{1}{n}||X\hat{\beta}-X\beta^\ast||_2^2 \preceq \min_T \max \left(\frac{\lambda_1^2 |T|}{k_T^2 \lambda_{\min}(\Sigma + \lambda_S L)},\lambda_1 ||(\tilde{\Gamma}\beta^\ast)_{T^c}||_1\right)$$
\etheos

This result holds for all choices of $T$.  Choosing $T$ to be $\text{supp}(\tPen\beta^{\ast})$ and $\text{supp}(\beta^{\ast})$ as in the proof of Theorem $\ref{theo:tvMain2}$ gives an analogous result for prediction error.
\btheos[Theorem \ref{theo:tvMain2} for Prediction Error] Suppose the conditions of Theorem \ref{theo:tvMain2} hold.  Then with probability at least $1-\frac{C_1}{p}$ we have
$$\frac{1}{n}||X\hat{\beta}-X\beta^\ast||_2^2 \preceq \frac{\lambda_1^2 ||\beta^\ast||_0}{\lambda_{\min}(\Sigma+\lambda_S L)}+\min\left(2\lambda_1^2\lambda_{TV}^2 ||\hat{\Sigma}||_{1,1}||\Gamma \beta^\ast||_0, 2\lambda_!\lambda_{TV}||\Gamma \beta^\ast||_1\right)$$
\etheos

\section{Proof of Lemma~\ref{lemma:eigenvalue}}
First note that 
\begin{eqnarray*}
\lambda_{\min}(\Sigma+\lrid L) & = &\lambda_{\min}((1-\lrid)\Sigma+\lrid (\Sigma + L))\\
& \geq & (1-\lrid)\lambda_{\min}(\Sigma) + \lrid \lambda_{\min}(\Sigma + L)
\end{eqnarray*}
where the second inequality follows from Weyl's inequality. For the remainder of the proof, we bound $\lambda_{\min}(\Sigma + L)$.
Recall that 
\begin{eqnarray}
\label{eq:SigmaplusGamma}
\Sigma+L = \Sigma - \hSigma + D
\end{eqnarray}
where $D\in\mathbb{R}^{p\times p}$ is a diagonal matrix with
$$D_{jj}=\sum_{k=1}^p|\hSigma_{j,k}|,~1\leq j\leq p.$$
Then
\begin{eqnarray*}
\lambda_{\min}(\Sigma+L)=\lambda_{\min}(\Sigma-\hSigma+D)\geq\lambda_{\min}(\Sigma-\hSigma)+\lambda_{\min}(D)
\end{eqnarray*}
by Weyl's inequality. Since
\begin{eqnarray*}
\lambda_{\min}(\Sigma-\hSigma)=-\lambda_{\max}(\hSigma-\Sigma)\geq-\|\Sigma-\hSigma\|_{op}\geq-\|\Sigma-\hSigma\|_{1,1}.
\end{eqnarray*}
Hence
\begin{eqnarray*}
\lambda_{\min}(\Sigma+L)&\geq&\lambda_{\min}(D)-\|\Sigma-\hSigma\|_{1,1}\\
&\geq&\min_{j}\sum_{k=1}^p|\hSigma_{j,k}|-\frac{c_{\ell}}{4}~(\text{by Assumption \ref{as:estimatecov}})\\
&\geq&\min_j \left[\sum_{k=1}^p |\Sigma_{j,k}| - \sum_{k=1}^p|\Sigma_{j,k}-\hSigma_{j,k}|\right]-\frac{c_{\ell}}{4}\\
&\geq&\min_{j}\sum_{k=1}^p|\Sigma_{j,k}|-\max_{j}\sum_{k=1}^p|\Sigma_{j,k}-\hSigma_{j,k}|-\frac{c_{\ell}}{4}\\
&\geq&c_{\ell}-\frac{c_{\ell}}{4}-\frac{c_{\ell}}{4}=\frac{c_{\ell}}{2}~(\text{by Assumptions \ref{as:sigmanormalization} and \ref{as:estimatecov}}).
\end{eqnarray*}

\section{Proof of Lemma \ref{lemma:kT}}
\label{sec:kT}
By the definition of $k_T$ we have
\begin{align*}
\sqrt{|T|}k_T^{-1} =& \sup_\beta \frac{\|(\tPen
                      \beta)_T\|_1}{\|\beta\|_2}\\
=&  \sup_{\beta: \|\beta\|_2=1} \|(\tPen\beta)_T\|_1\\
=&  \sup_{\beta: \|\beta\|_2=1} \ltv\|(\Pen\beta)_{T_1}\|_1+\|\beta_{T_2}\|_1\\
\leq&  \sup_{\beta: \|\beta\|_2=1} \ltv \|(\Pen\beta)_{T_1}\|_1+\sqrt{|T_2|}\|\beta\|_2\\
\leq&  \sup_{\beta: \|\beta\|_2=1} \ltv \|(\Pen\beta)_{T_1}\|_1+\sqrt{|T_2|}.
\end{align*}
Next we will bound the term $\|(\Pen\beta)_{T_1}\|_1$. First note that
\begin{eqnarray*}
\|(\Pen\beta)_{T_1}\|_1 \leq&  \sqrt{|T_1|}\|(\Pen\beta)_{T_1}\|_2\\
\leq& \sqrt{|T_1|\sum_{(j,k) \in E \bigcap T_1}
                             |\hSigma_{j,k}| |\beta_j -
                             \sign(\hSigma_{j,k}) \beta_k|^2}\\
\leq&  \sqrt{|T_1|\sum_{(j,k) \in E \bigcap T_1}
     |\hSigma_{j,k}| (2|\beta_j|^2+2|\beta_k|^2)}\\
\leq&  \sqrt{|T_1|\sum_{j=1}^p \left(\sum_{k: (j,k) \in E \bigcap T_1}
     2|\hSigma_{j,k}|\right) |\beta_j|^2} \\
\leq&  \sqrt{|T_1|}\sqrt{\max_{1\leq j\leq p} \left[\left(\sum_{k: (j,k) \in E \bigcap T_1}
     2|\hSigma_{j,k}|\right)\right]} \sqrt{\sum_{j=1}^p |\beta_j|^2} \\
\leq&  \sqrt{|T_1|}\sqrt{\max_{1\leq j\leq p} \left[\left(\sum_{k: (j,k) \in E \bigcap T_1}
     2|\hSigma_{j,k}|\right)\right]}  \\
\leq&  \sqrt{|T_1|} \sqrt{2\|\hSigma\|_{1,1}}.
\end{eqnarray*}
Thus
$$
k_T^{-1} \leq\frac{\ltv\sqrt{2\|\hSigma\|_{1,1}|T_1|}+\sqrt{|T_2|}}{\sqrt{|T_1|+|T_2|}},
$$
which completes the proof.

\section{Proof of Lemma \ref{lemma:rho}}
Note that $\Pen$ is the edge incidence matrix and $L = \Pen^\top \Pen$ is the weighted graph Laplacian matrix. Let the singular value decomposition for $\Gamma$ to be $\Gamma = U_{m\times p} D_{p\times p} V^\top_{p\times p}$. Next recall that
$\tPen = \begin{bmatrix}\ltv \Pen \\ I \end{bmatrix}$, then we have

\begin{align*}
\ptPen =& (\ltv^2 \Gamma^\top \Gamma + I)^{-1} \begin{bmatrix} \ltv \Pen^\top
  && I \end{bmatrix}\\
=&(\ltv^2 V D^2 V^\top + I)^{-1} \begin{bmatrix} \ltv V D U^\top && I\end{bmatrix}\\
=&V(\ltv^2  D^2 + I)^{-1} V^\top \begin{bmatrix} \ltv V D U^\top && I \end{bmatrix}\\
=&\begin{bmatrix}\underbrace{V(\ltv^2  D^2 + I)^{-1}\ltv D U^\top}_{=:A} 
&& \underbrace{V(\ltv^2  D^2 +I)^{-1} V^\top}_{=:B}\end{bmatrix}.
\end{align*}
From the definition of $\rho$ we can see that the maximum diagonal
entry of $(\ptPen)^\top \ptPen$ will just be $\rho^2$. Since
\begin{align*}
(\ptPen)^\top \ptPen =& \begin{bmatrix}A^\top A & A^\top B \\ B^\top A & B^\top B\end{bmatrix},
\end{align*}
we need to find the maximum diagonal values for matrices $A^\top
A$ and $B^\top B$.\\ 

Suppose there are $K$ connected components in the associated graph $G$. Thus the weighted graph Laplacian matrix $L$ is block diagonal, as is the matrix
$V$  (after appropriate permutation of rows
and columns), with each block corresponding to a different connected components. That is, each of the $K$ connected
components of the graph has its own weighted graph Laplacian $L_k = V_k D_k^2
V_k^\top,$ for $k=1,\ldots,K$ and the diagonal blocks of $V$ are the $V_k$s. Let $\mu_k$ be the minimum
nonzero eigenvalue of $L_k$. Let $B_k$
be the subset of vertices in the $k$-th connected component and $|B_k|$ be the
number of vertices in that component, and let $k(i)$ denote which
block contains vertex $i$. Now let $v_i^\top$ be the \ith row of $V$,
$u_i^\top$ be the \ith row of $U$, and note that $v_i$ is only
supported on $B_{k(i)}$. Further note that the first (upper left)
element of the $k$-th
diagonal block of $V$ is $1/\sqrt{|B_{k}|}$ if the minimum eigenvalue of $L_k$ is $0$.
Then we have:
\begin{align*}
B^\top B =& V(\ltv^2 D^2 + I)^{-2} V^\top,
\end{align*}
and then the maximum diagonal element for $B^\top B$ can be upper bounded as: 
\begin{eqnarray}
\text{max diag}(B^\top B) =& \max_{i \in \{1,\ldots,p\}} v_i^\top
                             (\ltv^2 D^2 + I)^{-2} v_i\notag\\
=& \max_{i \in \{1,\ldots,p\}} \sum_{j=1}^p \frac{v_{j,i}^2}{(\ltv^2 D_{jj}^2+1)^2} \notag\\
=& \max_{i \in \{1,\ldots,p\}} \sum_{j\in B_{k(i)}}\frac{v_{j,i}^2}{(\ltv^2
   D_{jj}^2+1)^2}\notag\\
\le& \max_{i \in \{1,\ldots,p\}} \left\{\frac{1}{|B_{k(i)}|} + \sum_{\substack{j \in
   B_{k(i)}: \\ D_{jj}^2 > 0}} \frac{v_{j,i}^2}{(\ltv^2 D_{jj}^2 + 1)^2}\right\}\\
\le& \max_{i \in \{1,\ldots,p\}} \left\{\frac{1}{|B_{k(i)}|} + \frac{1}{(\ltv^2 \mu_{k(i)} + 1)^2}\sum_{\substack{j \in
   B_{k(i)}: \\ D_{jj}^2 > 0}} v_{j,i}^2\right\}\notag\\
\le&\max_{i \in \{1,\ldots,p\}} \left\{\frac{1}{|B_{k(i)}|} + \frac{1}{(\ltv^2 \mu_{k(i)} + 1)^2}\right\}\notag\\
\le&\max_{k \in \{1,\ldots,K\}} \left\{\frac{1}{|B_{k}|} + \frac{1}{(\ltv^2 \mu_{k} + 1)^2}\right\}\notag. \label{eq:rsv}
\end{eqnarray}
On the other hand we note that 
\begin{align*}
A^\top A =& U\ltv^2 D^2 (\ltv^2 D^2 + I)^{-2} U^\top,
\end{align*}
similarly the maximum diagonal element for $A^\top A$ can be upper bounded as:
\begin{eqnarray}
\text{max diag}(A^\top A) =& \max_{i \in \{1,\ldots,m\}} u_i^\top
\ltv^2 D^2
                             (\ltv^2 D^2 + I)^{-2} u_i\notag\\
=& \max_{i \in \{1,\ldots,m\}} \sum_{j=1}^p \frac{\ltv^2 D^2_{jj} u_{j,i}^2}{(\ltv^2 D^2_{jj}+1)^2}\notag\\
=& \max_{i \in \{1,\ldots,m\}} \sum_{j=1}^p \frac{(\ltv^2 D^2_{jj} +1 - 1) u_{j,i}^2}{(\ltv^2 D^2_{jj}+1)^2}\notag\\
=& \max_{i \in \{1,\ldots,m\}} \sum_{j=1}^p \left\{\frac{u_{j,i}^2}{\ltv^2 D^2_{jj}+1}
   -\frac{u_{j,i}^2}{(\ltv^2 D^2_{jj}+1)^2}\right\} \notag\\
\le& \max_{i \in \{1,\ldots,m\}} \sum_{j \in \{1,\ldots,p\}:~D^2_{jj}>0} \left\{\frac{u_{j,i}^2}{\ltv^2 D^2_{jj}+1}\right\} \\
\le& \max_{i \in \{1,\ldots,m\}} \max_{j \in \{1,\ldots,p\}:~D^2_{jj}>0}\frac{1}{\ltv^2 D^2_{jj}+1}\sum_{j=1}^p u_{j,i}^2\notag \\
\le& \max_{i \in \{1,\ldots,m\}} \max_{j \in \{1,\ldots,p\}:~D^2_{jj}>0}\frac{1}{\ltv^2 D^2_{jj}+1}\notag\\
\le& \max_{k \in \{1,\ldots,K\}} \frac{1}{\ltv^2 \mu_k+1}.\notag \label{eq:lsv}
\end{eqnarray}
Then by combining the results above we have
\begin{eqnarray*}
\rho^2 &\le& \max_{1\leq k\leq K} \left\{ \frac{1}{|B_{k}|} + \frac{1}{(\ltv^2
    \mu_{k} + 1)^2} + \frac{1}{\ltv^2 \mu_{k}+1}\right\}\\
    &\leq& \max_{1\leq k\leq K} \left\{ \frac{1}{|B_{k}|} + \frac{2}{\ltv^2 \mu_{k}+1}\right\}.
\end{eqnarray*}
This completes the proof of Lemma \ref{lemma:rho}.

\section{Proof of Lemma \ref{lemma:blockcomplete}}\label{sec:blockCompleteProof}
By the definition of the block complete graph in Section \ref{sec:blockcomplete} we can see that $|B_k|=\frac{p}{K}$ for $1\leq k\leq K$ thus we have $\max_{1\leq k\leq K}\frac{1}{|B_k|}=\frac{K}{p}$. Note that $\mu_k$ is defined  to be the smallest non-zero eigenvalue of weighted graph Laplacian matrix for the $k^{\text{th}}$ complete graph. It is known that the smallest non-zero eigenvalue for un-weighted Laplacian matrix for complete graph is the number of nodes (see \citet[Section 4.1]{hutter2016optimal}). Thus, applying appropriate normalization $\mu_k=ar|B_k|=ar\frac{p}{K}=r$ since $a=\frac{K}{p}$. Hence $\mu_k=r$ for $1\leq k\leq K.$ Also note that $\lambda_{\min}(\Sigma)=a(1-r)$, so we have 
\begin{eqnarray*}
\lambda_{\min}(\Sigma+\lambda_SL)&=&\lambda_{\min}[(1-\lambda_S)\Sigma+\lambda_S(\Sigma+L)]\\
&\geq&(1-\lambda_S)\lambda_{\min}(\Sigma)+\lambda_S\lambda_{\min}(\Sigma+L)\\
&=&(1-\lambda_S)a(1-r)+\lambda_S[a+ar(\frac{p}{K}-1)]~(\text{by}~(\eqref{eq:SigmaplusGamma})~\text{and}~\hSigma=\Sigma)\\
&\geq&(1-\lambda_S)(1-r)\frac{K}{p}+\lambda_Sr~(\text{by using}~a=\frac{K}{p}).
\end{eqnarray*}
This completes the proof of Lemma \ref{lemma:blockcomplete}.

\section{Proof of Lemma \ref{lemma:chain}}
Let $\Gamma$ be the edge incidence matrix for the chain graph and let $\Gamma=UDV^T$ denote the SVD of $\Gamma$.  Note that the chain graph has one connected component, so in the language of Lemma \ref{lemma:rho} we have $|B_1|=p$.  From Equations \eqref{eq:rsv} and \eqref{eq:lsv} in the proof of Lemma \ref{lemma:rho} it follows that
$$\rho^2 \leq \max\left(\max_{i \in \{1,\ldots,p\}} \frac{1}{p}+\sum_{j: D_{j,j}^2>0} \frac{v_{j,i}^2}{(\lambda_{TV}^2D_{j,j}^2+1)^2}, \max_{i \in \{1,\ldots,p-1\}} \sum_{j: D_{j,j}^2>0} \frac{u_{j,i}^2}{\lambda_{TV}^2D_{j,j}^2+1}\right).$$
First note that if $\lambda_{TV}=0$ then $\rho^2 \leq \frac{1}{p}+1$ and our bound is satisfied, so for the remainder of the proof we assume $\lambda_{TV}>0$.

\textbf{Right singular vectors}: We first bound \begin{equation} \frac{1}{p}+\sum_{j: D_{j,j}^2>0} \frac{v_{j,i}^2}{(\lambda_{TV}^2D_{j,j}^2+1)^2}.\label{eq:first}\end{equation} The right singular vectors corresponding to the nonzero singular values are the normalized eigenvectors of the Laplacian matrix which  (see \citet[Section B.2]{hutter2016optimal})  are of the form 
$$v_{j,i} = \sqrt{\frac{2}{p}} \cos\left(\frac{(i+1/2)j\pi}{p}\right)$$ so in particular, $v_{j,i}^2 \leq \frac{2}{p}$ for all $i,j$.  Thus Equation \eqref{eq:first} is $$\leq \frac{1}{p}+\frac{2}{p} \sum_{j: D_{j,j}^2>0} \frac{1}{(\lambda_{TV}^2D_{j,j}^2+1)^2}.$$ The $\frac{D_{j,j}^2}{r^2}$ are the nonzero eigenvalues of the unweighted Laplacian matrix for the path graph which are also given in \citet[Section B.2]{hutter2016optimal} as $\sigma_{j}=2-2\cos(\frac{j\pi}{p})$ for $j=1,\ldots,p-1$.  We have $2-2\cos(\frac{j\pi}{p}) \geq \frac{j^2}{p^2} $ for $1 \leq k \leq p-1$ so this is 
\begin{eqnarray*} &\leq& \frac{1}{p}+\frac{2}{p} \sum_{j=1}^{p-1} \frac{1}{(\frac{r^2\lambda_{TV}^2j^2}{p^2}+1)^2}\\&=& \frac{1}{p} +2p^3 \sum_{j=1}^{p-1} \frac{1}{(r^2\lambda_{TV}^2j^2 +p^2)^2}\\&=&\frac{1}{p}+\frac{2p^3}{r^4\lambda_{TV}^4} \sum_{j=1}^{p-1} \frac{1}{(j^2+(\frac{p}{r\lambda_{TV}})^2)^2}.\end{eqnarray*} Because $f(j)=\frac{1}{(j^2+(\frac{p}{r\lambda_{TV}})^2)^2}$ is monotonically decreasing on $\mathbb{R}^+$ we get that this is 
\begin{eqnarray}&\leq& \frac{1}{p}+\frac{2p^3}{r^4\lambda_{TV}^4} \int_{x=0}^\infty \frac{1}{(x^2+(\frac{p}{r\lambda_{TV}})^2)^2} dx \notag\\ &=&\frac{1}{p}+ \frac{2p^3}{r^4\lambda_{TV}^4} \frac{\pi}{4(\frac{p}{r\lambda_{TV}})^3}=\frac{1}{p}+\frac{\pi} {2r\lambda_{TV}} \label{eq:fourth}.\end{eqnarray}
\textbf{Left singular vectors}
We next focus on bounding 
 \begin{equation}\sum_{j: D_{j,j}^2>0} \frac{u_{j,i}^2}{\lambda_{TV}^2D_{j,j}^2+1}.\label{eq:second} \end{equation}
 The $u_j$ are the normalized eigenvectors of $\Gamma \Gamma^T$.  A computation shows that 
 $$\Gamma \Gamma^T_{i,j}=\begin{cases} 2 \text{ if } i=j \\ -1 \text{ if } |i-j|=1 \\ 0 \text{ otherwise} \end{cases} $$
\citet[Section 1.5]{Strang}, gives $p-1$ orthonormal eigenvectors $u_j$ of $\Gamma \Gamma^T$ which are of the form $u_{j,i} = \sqrt{\frac{2}{p}}\sin(\frac{\pi ij}{p}))$.  In particular $u_{j,i}^2 \leq \frac{2}{p}$ so Equation \eqref{eq:second} is 
 \begin{equation} \leq \frac{2}{p} \sum_{j: D_{j,j}^2>0} \frac{1}{\lambda_{TV}^2D_{j,j}^2+1}.\label{eq:third}\end{equation}
 As before, we have that the $\frac{D_{j,j}^2}{r^2}$ are the nonzero eigenvalues of the unweighted Laplacian of the path graph, so they are of the form $2-2\cos(\frac{\pi j}{p})$ for $j=1,\ldots p-1$ and since $2-2\cos(\frac{\pi j}{p}) \geq \frac{j^2}{p^2}$ for $j=1,\ldots, p-1$ we get that this is \begin{eqnarray*}&\leq& \frac{2}{p} \sum_{j=1}^{p-1} \frac{1}{\frac{r^2\lambda_{TV}^2j^2}{p^2}+1}\\&=&2p \sum_{j=1}^{p-1} \frac{1}{p^2+r^2\lambda_{TV}^2j^2}\\ &=&\frac{2p}{r^2\lambda_{TV}^2} \sum_{j=1}^{p-1} \frac{1}{j^2+(\frac{p}{r\lambda_{TV}})^2}.\end{eqnarray*} Since $ f(j)=\frac{1}{j^2+(\frac{p}{r\lambda_{TV}})^2}$ is montonically decreasing on $\mathbb{R}^+$ we have that this is 
 \begin{eqnarray} &\leq& \frac{2p}{r^2\lambda_{TV}^2} \int_{x=0}^\infty \frac{1}{x^2+(\frac{p}{r\lambda_{TV}})^2} dx \notag\\ &=& \frac{2p}{r^2\lambda_{TV}^2} \frac{r\lambda_{TV}}{p} \arctan\left(\frac{r\lambda_{TV}x}{p}\right) \Big|_{x=0}^{x=\infty} =\frac{\pi}{r\lambda_{TV}}\label{eq:fifth}.\end{eqnarray}
 Moreover, since $u_{i,j}^2$ and $v_{i,j}^2$ are bounded by $\frac{2}{p}$ we immediately have the bound $$\rho^2 \leq \frac{1}{p}+2.$$
 Combining this with Equations \eqref{eq:fourth} and \eqref{eq:fifth} we conclude that 
 $$\rho^2 \leq \min\left(\frac{1}{p}+2,\max(\frac{\pi}{r\lambda_{TV}},\frac{1}{p}+\frac{\pi}{2r\lambda_{TV}})\right) \leq \frac{1}{p}+\frac{2\pi}{r\lambda_{TV}+1}$$ as claimed.

For the final part of the proof, note that $\lambda_{\min}(\Sigma)=a[1+2r\text{cos}(\frac{p}{p+1}\pi)]$ (see \citet[Section 2]{noschese2013tridiagonal}), so we have
\begin{eqnarray*}
\lambda_{\min}(\Sigma+\lambda_SL)&=&\lambda_{\min}[(1-\lambda_S)\Sigma+\lambda_S(\Sigma+L)]\\
&\geq&(1-\lambda_S)\lambda_{\min}(\Sigma)+\lambda_S\lambda_{\min}(\Sigma+L)\\
&=&(1-\lambda_S)a[1+2r\cos(\frac{p}{p+1}\pi)]+\lambda_Sa(1+r)~(\text{by}~(\ref{eq:SigmaplusGamma})~\text{and}~\hSigma=\Sigma)\\
&\geq&(1-\lambda_S)[1+2r\cos(\frac{p}{p+1}\pi)]+\lambda_S~(\text{by using}~a=1)\\
&\geq&(1-\lambda_S)(1-2r)+\lambda_S.
\end{eqnarray*}
This completes the proof of Lemma \ref{lemma:chain}.

\section{Proof of Lemma~\ref{lemma:lattice}}
Note that the lattice graph has one connected component, so in the language of Lemma 3 we have $|B_1|=p$.  Then from Equations \eqref{eq:rsv} and \eqref{eq:lsv} in the proof of Lemma \ref{lemma:rho} it follows that 
$$\rho^2 \leq \max\left(\max_{i \in \{1,\ldots,p\}} \frac{1}{p}+\sum_{j: D_{j,j}^2>0} \frac{v_{j,i}^2}{(\lambda_{TV}^2D_{j,j}^2+1)^2}, \max_{i \in \{1,\ldots,p-1\}} \sum_{j: D_{j,j}^2>0} \frac{u_{j,i}^2}{\lambda_{TV}^2D_{j,j}^2+1}\right).$$
First note that if $\lambda_{TV}=0$ then $\rho^2 \leq \frac{1}{p}+1$ and our bound is satisfied, so for the remainder of the proof we assume $\lambda_{TV}>0$.

\textbf{Right singular vectors} We first bound \begin{equation} \frac{1}{p}+\sum_{j: D_{j,j}^2>0} \frac{v_{j,i}^2}{(\lambda_{TV}^2D_{j,j}^2+1)^2}.\label{eq:lattice1}\end{equation}
The $v_j$ correspond to the normalized eigenvectors of the unweighted Laplacian for the Lattice graph.  We denote the Laplacian $L_{Lat}$.  Let $L_{\sqrt{p}}$ denote the unweighted Laplacian for the path graph with $\sqrt{p}$ nodes.  Since the Lattice graph is the direct product of two copies of the path graph, we have $$L_{Lat}=L_{\sqrt{p}} \otimes I_{\sqrt{p}} + I_{\sqrt{p}} \otimes L_{\sqrt{p}}.$$
Let $\{(w_k\}_{j=1,\ldots,\sqrt{p}}$ denote the normalized eigenvectors of $L_{\sqrt{p}}$ and $\sigma_k$ the corresponding eigenvalues.  Then
\begin{eqnarray*}L_{Lat}(w_k \otimes w_l)&=&L_{\sqrt{p}}w_k \otimes I_{\sqrt{p}}w_l + I_{\sqrt{p}}w_k \otimes L_{\sqrt{p}}w_l \\ &=&\sigma_kw_k \otimes w_l+w_k \otimes \sigma_lw_l=(\sigma_k+\sigma_l)(w_k \otimes w_l).\end{eqnarray*} 
The tensor product of unit vectors is also a unit vector, so $\|w_k \otimes w_l\|_2=1$ and $\{w_k \otimes w_l\}_{k,l=1,\ldots ,\sqrt{p}\}}$ are the normalized eigenvectors $v_j$ of $L_{Lat}$.  The $w_k$ were given in the proof of the path graph case as $w_{k,m} = \sqrt{\frac{2}{\sqrt{p}}}\cos\left(\frac{(m+1/2)k\pi}{\sqrt{p}}\right)$ so in particular we have $v_{j,i}^2 \leq \frac{4}{p}$.  Therefore Equation \eqref{eq:lattice1} is $$ \leq \frac{1}{p} +\frac{4}{p}\sum_{j:D_{j,j}^2>0} \frac{1}{(\lambda_{TV}^2D_{j,j}^2+1)^2}.$$
We have $\frac{D_{j,j}^2}{r^2}=\lambda_j$ where $\lambda_j$ denotes the $j$th eigenvalue of $L_{Lat}$.  We concluded above that the eigenvalues of $L_{Lat}$ are of the form $\sigma_k+\sigma_l$ where $\{\sigma_k\}_{k=0,\ldots,\sqrt{p}-1\}}$ are the eigenvalues of $L_{\sqrt{p}}$.  From the path graph proof, we know these are of the form $$\sigma_k+\sigma_l=4-2\cos(\frac{\pi k}{\sqrt{p}})-2\cos(\frac{\pi l}{\sqrt{p}}) \geq \frac{k^2+l^2}{p}$$ for $k,l=0,\ldots,\sqrt{p}-1$.  
Thus \begin{eqnarray*}\frac{1}{p} +\frac{4}{p}\sum_{j:D_{j,j}^2>0} \frac{1}{(\lambda_{TV}^2D_{j,j}^2+1)^2} \leq \frac{1}{p}+\frac{4}{p}\sum_{k=0}^{\sqrt{p}}\sum_{l=0}^{\sqrt{p}} \frac{\mathbbm{1}_{(k,l) \not =(0,0)}}{(r^2\lambda_{TV}^2\frac{k^2+l^2}{p}+1)^2}.\end{eqnarray*}
Algebraic rearrangement gives that this is 
\begin{eqnarray*}&=&\frac{1}{p}+4p\sum_{k=1}^{\sqrt{p}}\sum_{l=1}^{\sqrt{p}} \frac{1}{(r^2\lambda_{TV}^2(k^2+l^2)+p)^2}+8p \sum_{k=1}^{\sqrt{p}} \frac{1}{(r^2\lambda_{TV}^2k^2+p)^2} \\ &=&\frac{1}{p}+ \frac{4p}{r^4\lambda_{TV}^4} \sum_{k=1}^{\sqrt{p}}\sum_{l=1}^{\sqrt{p}} \frac{1}{(k^2+l^2+\frac{p}{r^2\lambda_{TV}^2})^2}+\frac{8p}{r^4\lambda_{TV}^4} \sum_{k=1}^{\sqrt{p}} \frac{1}{(k^2+\frac{p}{r^2\lambda_{TV}^2})^2}.\end{eqnarray*}
The above functions are monotonically decreasing in $k$ and $l$ for $k,l \geq 0$ and so we can say this is 
\begin{eqnarray}&\leq& \frac{1}{p} + \frac{4p}{r^4\lambda_{TV}^4}\int_{x=0}^\infty \int_{y=0}^\infty \frac{1}{(x^2+y^2+\frac{p}{r^2\lambda_{TV}^2})^2} dydx + \frac{8p}{r^4\lambda_{TV}^4} \int_{x=0}^\infty \frac{1}{(x^2+\frac{p}{r^2\lambda_{TV}^2})^2}dx\notag \\ &=&\frac{1}{p}+\frac{4p}{r^4\lambda_{TV}^4} \frac{\pi r^2\lambda_{TV}^2}{4p}+\frac{8p}{r^4\lambda_{TV}^4}\frac{\pi r^3\lambda_{TV}^3}{4p^{3/2}} =\frac{1}{p}+\frac{4\pi}{r^2\lambda_{TV}^2}+\frac{8\pi}{r\lambda_{TV}\sqrt{p}}\label{eq:lattice5}.\end{eqnarray} \textbf{Left singular vectors} We next focus on bounding \begin{equation}\sum_{j: D_{j,j}^2>0} \frac{u_{j,i}^2}{\lambda_{TV}^2D_{j,j}^2+1}\label{eq:lattice2}.\end{equation}
 The $u_j$ are the normalized eigenvectors of $\Gamma \Gamma^T$.  The eigenvectors of this matrix are nontrivial to derive, but \cite{wang2016trend} finds them in their proof of Corollary 8.  Moreover, they show that after normalizing the eigenvectors, each entry is bounded by $\sqrt{\frac{4}{p}}$.  In particular, we have $u_{j,i}^2 \leq \frac{4}{p}$ for all $i,j$ and so Equation \eqref{eq:lattice2} is 
$$\leq \frac{4}{p}\sum_{j: D_{j,j}^2>0} \frac{1}{\lambda_{TV}^2D_{j,j}^2+1}.$$
As in the right singular vector case, the $\frac{D_{j,j}^2}{r^2}$ are the eigenvalues of the unweighted Laplacian for the lattice graph, so they are of the form $$\sigma_k+\sigma_l=4-2\cos(\frac{\pi k}{\sqrt{p}})-2\cos(\frac{\pi l}{\sqrt{p}}) \geq \frac{k^2+l^2}{p}$$ for $k,l=0,\ldots,\sqrt{p}-1$.  Thus
$$\frac{4}{p}\sum_{j:D_{j,j}^2>0} \frac{1}{\lambda_{TV}^2D_{j,j}^2+1} \leq \frac{4}{p}\sum_{k=0}^{\sqrt{p}}\sum_{l=0}^{\sqrt{p}} \frac{\mathbbm{1}_{(k,l) \not =(0,0)}}{r^2\lambda_{TV}^2\frac{k^2+l^2}{p}+1}.$$
Algebraic manipulation gives that this is
\begin{eqnarray*}&=&4\sum_{k=1}^{\sqrt{p}}\sum_{l=1}^{\sqrt{p}} \frac{1}{r^2\lambda_{TV}^2(k^2+l^2)+p}+8 \sum_{k=1}^{\sqrt{p}} \frac{1}{r^2\lambda_{TV}^2k^2+p}\\
&=& \frac{4}{r^2\lambda_{TV}^2} \sum_{k=1}^{\sqrt{p}}\sum_{l=1}^{\sqrt{p}} \frac{1}{k^2+l^2+\frac{p}{r^2\lambda_{TV}^2}}+\frac{8}{r^2\lambda_{TV}^2} \sum_{k=1}^{\sqrt{p}} \frac{1}{k^2+\frac{p}{r^2\lambda_{TV}^2}}.\end{eqnarray*}
And now we use an integral comparison as before to conclude that this is 
\begin{eqnarray}&\leq& \frac{4}{r^2\lambda_{TV}^2} \int_{x=0}^{\sqrt{p}} \int_{y=0}^\infty \frac{1}{x^2+y^2+\frac{p}{r^2\lambda_{TV}^2}}dydx + \frac{8}{r^2\lambda_{TV}^2} \int_{x=0}^\infty \frac{1}{x^2+\frac{p}{r^2\lambda_{TV}^2}}dx \notag \\ &=&\frac{2\pi}{r^2\lambda_{TV}^2} \int_{x=0}^{\sqrt{p}} \frac{1}{\sqrt{x^2+\frac{p}{r^2\lambda_{TV}^2}}} dx+ \frac{8}{r^2\lambda_{TV}^2} \frac{\pi r\lambda_{TV}}{2\sqrt{p}}\label{eq:lattice3}.\end{eqnarray}
We compute this last integral explicitly as 
\begin{eqnarray*}\int_{x=0}^{\sqrt{p}} \frac{1}{\sqrt{x^2+\frac{p}{r^2\lambda_{TV}^2}}}dx&=&\frac{\pi}{2} \log(\sqrt{\frac{p}{r^2\lambda_{TV}^2}+x^2}+x) \Big|_{x=0}^{x=\sqrt{p}}\\ &=& \frac{\pi}{2} \log\left(\frac{\sqrt{\frac{p}{r^2\lambda_{TV}^2}+p}+\sqrt{p}}{\sqrt{\frac{p}{r^2\lambda_{TV}^2}}}\right).\end{eqnarray*}
Some additional algebra, along with the fact that $\sqrt{a+b} \leq \sqrt{a}+\sqrt{b}$ for $a,b \geq 0$ gives that this is  $$ \leq \frac{\pi}{2} \log(2+r\lambda_{TV}).$$  Overall we've concluded that Equation \eqref{eq:lattice3} is 
\begin{equation} \leq \frac{\pi^2\log(2+r\lambda_{TV})}{r^2\lambda_{TV}^2}+\frac{8\pi}{r\lambda_{TV}\sqrt{p}}\label{eq:lattice4}.\end{equation}
Moreover, since $u_{i,j}^2$ and $v_{i,j}^2$ are bounded by $\frac{4}{p}$ we immediately have the bound $$\rho^2 \leq 5.$$
Combining this with Equations \eqref{eq:lattice5} and \eqref{eq:lattice4} we conclude that 
$$\rho^2 \leq \min\left(5,\frac{1}{p}+\frac{4\pi \log(2+r\lambda_{TV})}{r^2\lambda_{TV}^2}+\frac{8\pi}{r\lambda_{TV}\sqrt{p}}\right) \leq \frac{1}{p}+\frac{5\pi\log(2+r\lambda_{TV})}{r^2\lambda_{TV}^2+1}+\frac{10\pi}{r\lambda_{TV}\sqrt{p}+1}$$ as claimed.

For the final part of the proof recall that $r \in (0,\frac{1}{4})$.  Thus $\Sigma$ is diagonally dominant with $\Sigma_{i,i}-\sum_{j \not =i} \Sigma_{i,j} \geq 1-4r>0$ for all $i$ and therefore $\lambda_{\min}(\Sigma) \geq 1-4r$.  This implies that
\begin{eqnarray*}
\lambda_{\min}(\Sigma+\lambda_SL)&=&\lambda_{\min}[(1-\lambda_S)\Sigma+\lambda_S(\Sigma+L)]\\
&\geq&(1-\lambda_S)\lambda_{\min}(\Sigma)+\lambda_S\lambda_{\min}(\Sigma+L)\\
&=&(1-\lambda_S)(1-4r)+\lambda_S(1+2r)~(\text{by}~(\ref{eq:SigmaplusGamma})~\text{and}~\hSigma=\Sigma)
\\&\geq&(1-\lambda_S)(1-4r)+\lambda_S.
\end{eqnarray*}
This completes the proof of Lemma \ref{lemma:lattice}.

\section{Extension to Logistic Regression}
\label{sec:logistic}
In the main body of the paper we consider only a linear model in the interest of simplicity.  However, it is straightforward to extend the theory in this paper to generalized linear models.  In this section we need to assume that $||\beta^\ast||_1 \leq u$ for a universal constant $u$.  We will informally sketch an extension to logistic regression.  Consider a logistic model where
$$y_i \sim \mbox{Bernoulli}(\lambda_i)$$
$$\lambda_i=\frac{1}{1+\exp(-\langle \beta^\ast,X_i \rangle)}.$$
Instead of using squared loss, we want to use the logistic loss function
$$L(\beta;X,y)=\sum_{i=1}^n \log(1+\exp(\langle \beta^\ast,X_i \rangle))-y_i \langle \beta^\ast,X_i \rangle.$$
The GTV estimator for the logistic model takes the form
\begin{align}
\hat \beta = &~\argmin_{\beta: ||\beta||_1 \leq u} \frac{1}{n} L(\beta)+ \lrid
\sum_{j,k} |\hSigma_{j,k}| (\beta_j - \hat s_{j,k} \beta_k)^2\nonumber
\\
& \qquad + \lone (\ltv
\sum_{j,k} |\hSigma_{j,k}|^{1/2} |\beta_j - \hat s_{j,k} \beta_k|+\|\beta\|_1).\label{eq:logistic_estimator}
\end{align}
For convenience define 
$$R(\beta):=\lrid
\sum_{j,k} |\hSigma_{j,k}| (\beta_j - \hat s_{j,k} \beta_k)^2
 + \lone (\ltv
\sum_{j,k} |\hSigma_{j,k}|^{1/2} |\beta_j - \hat s_{j,k} \beta_k|+\|\beta\|_1).$$
To derive similar theoretical bounds in this setting, first note that by definition
$$\frac{1}{n} L(\hat \beta) \leq \frac{1}{n} L(\beta^\ast) +(R(\beta^\ast)-R(\hat \beta)).$$
We now use standard steps for the analysis of generalized models in order to reduce our problem to the linear setting in the proof of Theorem \ref{theo:tvMain}.  For the remainder of the section, we use the shorthand $f(x)=\log(1+\exp(x))$.  Using the definition of $L(\beta)$ and rearranging terms yields
$$\sum_{i=1}^n\frac{1}{n}  f(\langle \hat \beta,X_i \rangle)-f(\langle \beta^\ast,X_i \rangle)-y_i \langle \triangle, X_i \rangle \leq R(\beta^\ast)-R(\hat \beta).$$
Define $\epsilon_i:=y_i-\mathbb{E}[y_i|X_i]=y_i-f'(\langle \beta^\ast,X_i \rangle)$ and then
\begin{equation}\frac{1}{n}  f(\langle \hat \beta,X_i \rangle)-f(\langle \beta^\ast,X_i \rangle)- f'(\langle \beta^\ast,X_i \rangle) \langle \triangle, X_i \rangle \leq \frac{1}{n}\sum_{i=1}^n \epsilon_i\langle \triangle, X_i \rangle + R(\beta^\ast)-R(\hat \beta).\label{eq:convexity}\end{equation}
For $x,y$ contained in an interval $[-d,d]$, $f$ is a strongly convex function so that
$$f(x)-f(y)-f'(y)(x-y) \geq \psi \|x-y\|_2^2$$ for a strong convexity parameter $\psi$ which depends on $d$.  
Applying this to Equation \eqref{eq:convexity}, 
\begin{equation}\frac{\psi}{n}  \langle \triangle, X_i \rangle^2 \leq \frac{1}{n}\sum_{i=1}^n \epsilon_i\langle \triangle, X_i \rangle+ R(\beta^\ast)-R(\hat \beta).\label{eq:second_convexity}\end{equation}
Assuming $||\beta^\ast||_1,||\hat{\beta}||_1 \leq u$ for a universal constant $u$, the convexity parameter $\psi$ is also bounded by a universal constant.  Rearranging terms and ignoring the factor of $\psi$, the inequality in Equation \eqref{eq:second_convexity} is exactly the inequality at the beginning of the proof of Theorem \ref{theo:tvMain}.  Thus all the bounds derived in the linear setting also apply in the logistic regression setting up to a factor of the convexity parameter $\psi$.

\section{Proof of Lemma \ref{lemma:gwidthcompare}}
We will use two classical Lemmas for Gaussian processes~\citet{AndersonStat,Slepian62} to prove our results.
\blems[Anderson's comparison inequality]
\label{lemma:anderson}
Let $X$ and $Y$ be zero-mean Gaussian random vectors with covariance $\Sigma_X$ and $\Sigma_Y$ respectively. If $\Sigma_Y-\Sigma_X$ is positive semi-definite then for any convex symmetric set $C$,
\begin{eqnarray*}
	P(X\in C)\leq P(Y\in C).
\end{eqnarray*}
\elems

\blems[Slepian's Lemma]
\label{lemma:slepian}
Let $\{G_s,s\in S\}$ and $\{H_s,s\in S\}$ be two centered Gaussian processes defined over the same index set $S$. Suppose that both processes are almost surely bounded. For each $s,t\in S$, if $\mathbb{E}(G_s-G_t)^2\leq \mathbb{E}(H_s-H_t)^2$, then $\mathbb{E}[\sup_{s\in S}G_s]\leq \mathbb{E}[\sup_{s\in S}H_s].$ Further if $\mathbb{E}(G_s^2)=\mathbb{E}(H_s^2)$ for all $s\in S$, then
\begin{eqnarray*}
	P\{\sup_{s\in S}G_s>x\}\leq P\{\sup_{s\in S}H_s>x\},
\end{eqnarray*}
for all $x>0$.
\elems

$(X\tPen^{\dagger})^{\top}\epsilon=\sum_{i=1}^n\langle\tPen^{\dagger},\epsilon_iX^{(i)}\rangle$ and $Cov(X)=\Sigma\preceq\lambda_{\max}(\Sigma)I_{p\times p}$. Then by Assumption~\ref{as:tvlambdasigma} $\lambda_{\max}(\Sigma)\leq{c_u}$, and if we use Lemma~\ref{lemma:anderson}, for any $x>0$ 
\begin{eqnarray*}
P\{\sup\sum_{i=1}^n\langle\tPen^{\dagger},\epsilon_iX^{(i)} \rangle\leq x\}\geq P\{\sup\sqrt{{c_u}}\sum_{i=1}^n\langle\tPen^{\dagger},\epsilon_ig_i \rangle\leq x\},
\end{eqnarray*}
where $X^{(i)}$ is the $i^{\text{th}}$ row of matrix $X$ and $\{g_i:~i=1,...,n\}$ is i.i.d. standard normal Gaussian vectors with $g_i\in\mathbb{R}^p$. Now let $G\in\mathbb{R}^p$ be an i.i.d. standard norm Gaussian vector and define the zero-mean Gaussian process $\sqrt{n}\sigma\langle\tPen^{\dagger},G \rangle$, we can see that the conditions in Lemma \ref{lemma:slepian} are satisfied for two centered Gaussian processes $\sum_{i=1}^n\langle\tPen^{\dagger},\epsilon_ig_i\rangle$ and $\sqrt{n}\sigma\langle\tPen^{\dagger},G \rangle$ thus we have
\begin{eqnarray*}
P\{\sup\sqrt{{c_u}}\sum_{i=1}^n\langle\tPen^{\dagger},\epsilon_ig_i\rangle\leq x\}\geq P\{\sup\sigma\sqrt{n{c_u}}\langle\tPen^{\dagger},G \rangle\leq x\}.
\end{eqnarray*}
Further, using known results on Gaussian maxima (\citet[Theorem 2.5]{boucheron2013concentration}), $\sup\langle \tPen^{\dagger},G \rangle\leq3\rho\sqrt{\log (m+p)}$ with probability at least $1-\frac{C_1}{p}$ for some absolute constant $C_1>0$. By choosing $x=3\sigma\rho\sqrt{n{c_u}\log (m+p)}$,
\begin{eqnarray*}
P\{\sup\sum_{i=1}^n\langle\tPen^{\dagger},\epsilon_iX^{(i)}\rangle\leq x\}\geq P\{\sup\sigma\sqrt{n{c_u}}\langle\tPen^{\dagger},G\rangle\leq x\}\geq1-\frac{C_1}{p}.
\end{eqnarray*}
Thus we have shown with high probability that $\|(X\tPen^{\dagger})^{\top}\epsilon\|_{\infty}\leq3\sigma\rho\sqrt{n{c_u}\log (m+p)}.$ Since $m$ is the number of edges, $m\leq \frac{p(p-1)}{2}$, thus with probability at least $1-\frac{C_1}{p}$ we have that $\|(X\tPen^{\dagger})^{\top}\epsilon\|_{\infty}\leq6\sigma\rho\sqrt{n{c_u}\log p}$. This completes the proof.

\section{Proof of Lemma \ref{lemma:tvrec}}
The proof of Lemma \ref{lemma:tvrec} involves two parts.\\

\noindent \textbf{Part 1}: We first show that the following inequality 
\begin{eqnarray}
\label{eq:gaussianrec}
	\frac{\|X\Delta\|_2}{\sqrt{n}}\geq\frac{1}{4}\|\Sigma^{1/2}\Delta\|_2-9\frac{\lone}{\sigma}\|\tPen \Delta\|_1
\end{eqnarray}
holds with probability at least $1-c_4\exp(-c_5n)$ by using similar techniques to those used to prove Theorem 1 in \citet{RasWaiYu10b}.

First note that it is sufficient to show (\ref{eq:gaussianrec}) holds with $\|\Sigma^{1/2}\Delta\|_2=1$. The reason is as follows: if $\|\Sigma^{1/2}\Delta\|_2=0$ we can see that (\ref{eq:gaussianrec}) holds trivially; otherwise when $\|\Sigma^{1/2}\Delta\|_2>0$ we can define $\tilde{\Delta}=\frac{\Delta}{\|\Sigma^{1/2}\Delta\|_2}$ then we have $\|\Sigma^{1/2}\tilde{\Delta}\|_2=1$. Since (\ref{eq:gaussianrec}) is invariant with respect to the scale of $\Delta$, if it holds for $\tilde{\Delta}$, it also holds for $\Delta$. Thus in the following proof we just assume that $\|\Sigma^{1/2}\Delta\|_2=1$. To show (\ref{eq:gaussianrec}) with $\|\Sigma^{1/2}\Delta\|_2=1$ holds there are three main steps:

\noindent (1) Since we want to lower bound $\frac{\|X\Delta\|_2}{\sqrt{n}}$ in terms of $\|\Sigma^{1/2}\Delta\|_2$ and $\|\tPen \Delta\|_1$, we define the set $V(r):=\{\Delta\in\mathbb{R}^p~|~\|\Sigma^{1/2}\Delta\|_2=1,~\|\tPen \Delta\|_1\leq r\}$ for a fixed radius $r$. Note that we are only concerned with choices of $r$ such the set $V(r)$ is non-empty. Our first step is to give an upper bound for $\mathbb{E}[M(r,X)]$, where $M(r,X)$ is defined as:
\begin{eqnarray*}
M(r,X):=1-\inf_{\Delta\in V(r)}\frac{\|X\Delta\|_2}{\sqrt{n}}=\sup_{\Delta\in V(r)}\left\{1-\frac{\|X\Delta\|_2}{\sqrt{n}}\right\}.
\end{eqnarray*}

\noindent (2) The second step is to use concentration inequalities to show that with high probability for each fixed $r>0$, the random quantity $M(r,X)$ is sharply concentrated around $\mathbb{E}[M(r,X)].$

\noindent (3) The third step is to use a peeling argument to show that the analysis holds uniformly over all possible values of $r$ with high probability, then we can show that (\ref{eq:gaussianrec}) holds with high probability.

In the following proof we only provide details for proving step (1) since our proof for step (2) and (3) will be identical to those in \citet{RasWaiYu10b}. For step (1) we prove the following lemma:

\blems
\label{lemma:step1}
For any radius $r>0$ such that $V(r)$ is non-empty, we have
\begin{eqnarray*}
\mathbb{E}[M(r,X)]\leq\frac{1}{4}+3r\frac{\lambda_1}{\sigma}.
\end{eqnarray*}
\elems

\spro
Define the Euclidean sphere of radius 1 to be $S^{n-1}=\{u\in\mathbb{R}^n~|~\|u\|_2=1\}$. Then $\|X\Delta\|_2=\sup_{u\in S^{n-1}}u^{\top}X\Delta$. In order to write the quantity $M(r,X)$ in a form that is easier to analyze, we define $Y_{u,\Delta}:=u^{\top}X\Delta$ for each pair $(u,\Delta)\in S^{n-1}\times V(r)$. Then we have 
\begin{eqnarray*}
-\inf_{\Delta\in V(r)}\|X\Delta\|_2=-\inf_{\Delta\in V(r)}\sup_{u\in S^{n-1}}u^{\top}X\Delta=\sup_{\Delta\in V(r)}\inf_{u\in S^{n-1}}Y_{u,\Delta}.
\end{eqnarray*}
Next we will use a Gaussian comparison inequality to upper bound the expected value of the quantity $\sup_{\Delta\in V(r)}\inf_{u\in S^{n-1}}Y_{u,\Delta}$. Here we use a form of Gordon's inequality that is stated in \citet{DavSza01} for our analysis. Suppose that $\{Y_{u,\Delta},(u,\Delta)\in U\times V\}$ and $\{Z_{u,\Delta},(u,\Delta)\in U\times V\}$ are two zero-mean Gaussian processes on $U\times V$. We denote $\sigma(\cdot)$ to be the standard deviation of a random variable. Using Gordon'e inequality, if
\begin{eqnarray*}
\sigma(Y_{u,\Delta}-Y_{u',\Delta'})\leq\sigma(Z_{u,\Delta}-Z_{u',\Delta'}),~~\forall~(u,\Delta)~\text{and}~(u',\Delta')\in U\times V,
\end{eqnarray*}
and this inequality holds with equality when $\Delta=\Delta'$, then
\begin{eqnarray*}
\mathbb{E}[\sup_{\Delta\in V}\inf_{u\in U}Y_{u,\Delta}]\leq\mathbb{E}[\sup_{\Delta\in V}\inf_{u\in U}Z_{u,\Delta}].
\end{eqnarray*}
Now we consider the zero-mean Gaussian process $Z_{u,\Delta}$ with $(u,\Delta)\in S^{n-1}\times V(r)$ as follows:
\begin{eqnarray*}
Z_{u,\Delta} = g^{\top}u+h^{\top}\Sigma^{1/2}\Delta,
\end{eqnarray*}
where $g\sim {\cal N}(0,I_{n\times n})$ and $h\sim {\cal N}(0,I_{p\times p})$. It follows that (see \citet{RasWaiYu10b} for more details)
\begin{eqnarray*}
\sigma(Y_{u,\Delta}-Y_{u',\Delta'})\leq\sigma(Z_{u,\Delta}-Z_{u',\Delta'}),~~\forall~(u,\Delta)~\text{and}~(u',\Delta')\in S^{n-1}\times V(r),
\end{eqnarray*}
and the equality holds when $\Delta=\Delta'$. Thus we can apply Gordon's inequality to conclude that 
\begin{eqnarray*}
\mathbb{E}[\sup_{\Delta\in V(r)}\inf_{u\in S^{n-1}}Y_{u,\Delta}]&\leq&\mathbb{E}[\sup_{\Delta\in V(r)}\inf_{u\in S^{n-1}}Z_{u,\Delta}]\\
&=&\mathbb{E}[\inf_{u\in S^{n-1}}g^{\top}u]+\mathbb{E}[\sup_{\Delta\in V(r)}h^{\top}\Sigma^{1/2}\Delta]\\
&=&-\mathbb{E}[\|g\|_2]+\mathbb{E}[\sup_{\Delta\in V(r)}h^{\top}\Sigma^{1/2}\Delta].
\end{eqnarray*}
Next we bound the term $\mathbb{E}[\sup_{\Delta\in V(r)}h^{\top}\Sigma^{1/2}\Delta]$ using the following lemma:
\blems
\label{lemma:gwidthcompare2}
Suppose we have $\lambda_1\geq48\rho \sigma \sqrt{\frac{{c_u}\log p}{n}}$, then 
we have that
\begin{eqnarray*}
\lambda_1\geq8\frac{\sigma}{\sqrt{n}}\mathbb{E}[\|(\Sigma^{1/2}\tilde{\Gamma}^{\dagger})^{\top}h\|_{\infty}].
\end{eqnarray*}
with probability at least $1-\frac{c}{p}$ for some absolute constant $c>0$.
\elems
The proof for this lemma will be provided shortly. Thus
\begin{eqnarray*}
\mathbb{E}[\sup_{\Delta\in V(r)} |{h}^{\top} \Sigma^{1/2}\Delta|]&=&\mathbb{E}[\sup_{\Delta\in V(r)} |{h}^{\top} \Sigma^{1/2}\tPen^{\dagger}\tPen \Delta|]\\
&\leq&\mathbb{E}[\sup_{\Delta\in V(r)} \|{h}^{\top} \Sigma^{1/2}\tPen^{\dagger}\|_{\infty}\|\tPen \Delta\|_1]\\
&\leq& \mathbb{E}[\|(\Sigma^{1/2}\tPen^{\dagger})^{\top} h\|_{\infty}]r\\
&\leq& 3r\frac{\lone}{\sigma}\sqrt{n},
\end{eqnarray*}
where the last inequality holds with high probability from Lemma \ref{lemma:gwidthcompare2}. Also by standard $\chi^2$ tail bounds \citep{LedTal91} when $n\geq10$ we have $\mathbb{E}[\|g\|_2]\geq\frac{3}{4}\sqrt{n}$. By combining hese pieces together
\begin{eqnarray*}
\mathbb{E}[-\inf_{\Delta\in V(r)}\|X\Delta\|_2]\leq-\frac{3}{4}\sqrt{n}+3r\frac{\lambda_1}{\sigma}\sqrt{n}.
\end{eqnarray*}
Thus by dividing by $\sqrt{n}$ and adding 1 to both sides we have
\begin{eqnarray*}
\mathbb{E}[M(r,X)]=\mathbb{E}[1-\inf_{\Delta\in V(r)}\frac{\|X\Delta\|_2}{\sqrt{n}}]\leq\frac{1}{4}+3r\frac{\lambda_1}{\sigma}.
\end{eqnarray*}
\fpro
Then by following the rest proof in \citet{RasWaiYu10b} for step (2) and (3), we can show that with probability at least $1-c_4\exp(-c_5n)$,
\begin{eqnarray*}
	\frac{\|X\Delta\|_2}{\sqrt{n}}\geq\frac{1}{4}\|\Sigma^{1/2}\Delta\|_2-9\frac{\lone}{\sigma}\|\tPen \Delta\|_1.
\end{eqnarray*}

\noindent \textbf{Part 2}: Next we can go to second part of the proof. From (\ref{eq:tv4}) and (\ref{eq:tvcone}) we know that 
\begin{eqnarray*}
\|\tPen \Delta\|_1&\leq&4\|(\tPen \Delta)_T\|_1+4\|(\tPen\bstar)_{T^c}\|_1\\
&\leq&\frac{4\sqrt{|T|}\|\Delta\|_2}{k_T}+4\|(\tPen\bstar)_{T^c}\|_1.
\end{eqnarray*}
Then
\begin{eqnarray*}
	\frac{\|X\Delta\|_2}{\sqrt{n}}\geq\frac{1}{4}\|\Sigma^{1/2}\Delta\|_2-9\frac{\lone}{\sigma}\left(4\|(\tPen\bstar)_{T^c}\|_1+\frac{4\sqrt{|T|}\|\Delta\|_2}{k_T}\right).
\end{eqnarray*}
Thus there exist constants $c',c''>0$ such that 
\begin{eqnarray*}
	\Delta^{\top} \left(\frac{X^{\top} X}{n}+\lrid L\right)\Delta\geq c'\Delta^{\top} (\Sigma+\lrid L)\Delta-c''\lone^2\left(\|(\tPen\bstar)_{T^c}\|_1^2+\frac{|T|\|\Delta\|_2^2}{k_T^2}\right).
\end{eqnarray*}
Since 
\begin{eqnarray*}
\Delta^{\top} (\Sigma+\lrid L)\Delta &\geq& \lambda_{\min}(\Sigma+\lrid L)\|\Delta\|_2^2,
\end{eqnarray*}
then when $\lone$ satisfies (\ref{eq:ltvupper}) for some constant
$c_2>0$,
\begin{eqnarray*}
\Delta^{\top} (\frac{X^{\top} X}{n}+\lrid L)\Delta\geq c_1\lambda_{\min}(\Sigma+\lrid L)\|\Delta\|_2^2-c_3\lone^2\|(\tPen\bstar)_{T^c}\|_1^2,
\end{eqnarray*}
for absolute constants $c_1,c_3>0$.

\section{Proof of Lemma \ref{lemma:gwidthcompare2}}
Here we use similar techniques to the proof of Lemma \ref{lemma:gwidthcompare}. First note that $\Sigma\preceq{c_u}I_{p\times p}$ and by using Lemma~\ref{lemma:anderson} we have for any $y>0$ the following inequality
\begin{eqnarray*}
P\{\sup[(\Sigma^{1/2}\tPen^{\dagger})^{\top}h]\leq y\}=P\{\sup\langle\tPen^{\dagger},\Sigma^{1/2}h\rangle\leq y\}\geq P\{\sup\langle\tPen^{\dagger},h\rangle\leq\frac{y}{\sqrt{{c_u}}}\}.
\end{eqnarray*}
Since $h\sim {\cal N}(0,I_{p\times p})$ then also by known results on Gaussian maxima (\citet[Theorem 2.5]{boucheron2013concentration}) we have $\sup\langle\tPen^{\dagger},h\rangle\leq3\rho\sqrt{\log (m+p)}$ with probability at least $1-\frac{c}{p}$ for some constant $c>0$. Then we can choose $y=3\rho\sqrt{{c_u}\log (m+p)}$ and
\begin{eqnarray*}
P\{\sup[(\Sigma^{1/2}\tPen^{\dagger})^{\top}h]\leq y\}\geq P\{\sup\langle\tPen^{\dagger},h\rangle\leq\frac{y}{\sqrt{{c_u}}}\}\geq 1-\frac{c}{p}.
\end{eqnarray*}
Thus with high probability $\|(\Sigma^{1/2}\tPen^{\dagger})^{\top}h\|_{\infty}\leq3\rho\sqrt{{c_u}\log (m+p)},$ then using the fact that $m\leq\frac{p(p-1)}{2}$, $\|(\Sigma^{1/2}\tPen^{\dagger})^{\top}h\|_{\infty}\leq6\rho\sqrt{{c_u}\log p}$ holds with probability at least $1-\frac{c}{p}$. This completes the proof of Lemma \ref{lemma:gwidthcompare2}.

\section{Simulation Details}
\label{sec:sim_details}
The graphs and corresponding covariance structures are constructed as follows:

\paragraph*{Block Complete Graph}$\Sigma$ is block diagonal with $K$ blocks, each of size $\frac{p}{K}\times\frac{p}{K}$. Following the discussion in Section \ref{sec:blockcomplete},  all the diagonal elements are set to $\frac{K}{p}$ and all the  off-diagonal elements in each block are set to $\frac{Kr}{p}$ with $r\in(0,1)$. Here $r$ is the correlation coefficient and will be set to different values in the experiments. Specifically, let
$$B = \frac{K}{p}\left(r\ones_{p/K}\ones_{p/K}^\top + (1-r)I_{p/K}\right) \qquad \mbox{ and } \qquad \Sigma = I_K \otimes B,$$
where $\otimes$ denotes the Kronecker product.
To set the true coefficient vector $\bstar$, we first randomly choose $\ell$ of the $K$ blocks  to be ``active blocks''. Then we set the elements in $\bstar$ that correspond to the $\ell$ active blocks to be $\beta^{\ast}_j\sim {\cal N}(1,0.01^2)$ when $i$ belongs to these $\ell$ active blocks and all other elements in $\beta^{\ast}$ to be 0 (inactive). That is, let $S \in \{0,1\}^p$ indicate the indices in active blocks (and hence the support of $\bstar$); then
$$\bstar \sim {\cal N}(S,0.01^2 \text{diag}(S)).$$
\paragraph*{Chain Graph:}Following the discussion in Section \ref{sec:chain}, we set elements in the main diagonal of $\Sigma$ to be one, the first off-diagonal elements to be $r$ with $r\in(0,\frac{1}{2})$, and all the other elements to be zero; \ie
\begin{eqnarray*}
\Sigma_{j,k}=\begin{cases}
1, & \text{if}~j=k,\\
r, & \text{if}~|j-k|=1,\\
0, & \text{else}.
\end{cases}
\end{eqnarray*}
The corresponding true coefficient vector $\bstar$ is set to have $\beta^{\ast}_j \sim {\cal N}(1,0.01^2)$ for $1\leq j\leq s$ and the remaining elements to be zero. That is, let $S \in \{0,1\}^p$ have its first $s<p$ elements be one and the remaining be zero; then
$$\bstar \sim {\cal N}(S,0.01^2 \text{diag}(S)).$$
\paragraph*{Lattice Graph}Following the discussion in Section \ref{sec:lattice}, we construct $\Sigma$ as follows.
\begin{eqnarray*}
\Sigma_{j,k}=\begin{cases}
1, & \text{if}~j=k,\\
r, & \text{if}~|j-k|=1 \text{ and } \min(j,k) \not =0 \bmod \sqrt{p} ,\\
r, & \text{if}~|j-k|=\sqrt{p} \\
0, & \text{else}.
\end{cases}
\end{eqnarray*}
The corresponding true coefficient vector $\bstar$ with $s$ active elements is set to $\beta^{\ast}_j \sim {\cal N}(1,0.01^2)$ if $j \leq \sqrt{s} \bmod \sqrt{p}$ and $j \leq \sqrt{ps}$ and is set to $\beta_j^\ast=0$ otherwise.  This corresponds to an active $\sqrt{s} \times \sqrt{s}$ sublattice within the $\sqrt{p} \times \sqrt{p}$ lattice.   The remaining elements outside of this sublattice are set to zero.
\newpage
\section{Biochemistry Table}
\LTcapwidth=\linewidth
\begin{longtable}{p{7cm}|p{9cm}}
\hline \multicolumn{1}{c|}{\textbf{Structure Feature}} & \multicolumn{1}{c}{\textbf{Description}} \\ \hline 
\endfirsthead
\hline \multicolumn{1}{c|}{\textbf{Structure Feature}} & \multicolumn{1}{c}{\textbf{Description}} \\ \hline 
\endhead
\hline \hline
\caption{Description of structural features used in the biochemistry data analysis.} \label{tab:biochem_table} 
\endlastfoot

buried\_np\_AFILMVWY\_per\_res & Buried nonpolar surface area on nonpolar amino acids/count buried and core residues. \\ \hline 
  buried\_np\_per\_res & Buried nonpolar surface area of the protein divided by count buried non polar residues. \\ \hline 
  buried\_over\_exposed & buried\_np\_per\_res divided by solvent available surface area (sasa) of hydrophobic residues. \\ \hline 
  buried\_over\_exposed\_AFILMVWY & buried\_np\_AFILMVWY\_per\_res divided by sasa of hydrophobic residues. \\ \hline 
  cbeta & A solvation term intended to correct for the excluded volume effect introduced by the simulation and favor compact structures. It is based on the ratio of probabilities of a residue having a given number of neighbors in a compact structure vs. random coil and summed over all residues. \\ \hline 
  cenpack & A centroid energy term. \\ \hline 
  contact\_all\_per\_res & Count sidechain carbon-carbon contacts among all residues under the given distance cutoff divided by count residues of the sequence modeled. \\ \hline 
  contact\_buried\_core\_boundary\_per\_res & Count sidechain carbon-carbon contacts among the buried and boundary residues under the given distance cutoff divided by count buried and boundary residues. \\ \hline 
  contact\_buried\_core\_per\_res & Count sidechain carbon-carbon contacts among the buried residues under the given distance cutoff divided by count buried residues. \\ \hline 
  degree\_core\_boundary\_per\_res & Count number of residues within a set distance of buried and boundary residues divided by count buried and boundary residues. \\ \hline 
  degree\_core\_per\_res & Count number of residues within a set distance of buried residues divided by count buried residues. \\ \hline 
  degree\_per\_res & Count number of residues within a set distance of other residues divided by count residues of the sequence modeled. \\ \hline 
  env & A context-dependent one-body energy term that describes the solvation of a particular residue (based on the hydrophobic effect). It is based on the probability of a residue having the specified type given its number of neighboring residues. \\ \hline 
  exposed\_hydrophobics\_per\_res & Sasa of hydrophobic residues divided by count residues of the sequence modeled. \\ \hline 
  exposed\_polars\_per\_res & Sasa of polar residues divided by count residues of the sequence modeled. \\ \hline 
  exposed\_total\_per\_res & Sasa of whole protein divided by count residues of the sequence modeled. \\ \hline 
  fa\_atr & Lennard-Jones attractive. \\ \hline 
  fa\_dun & Internal energy of sidechain rotamers as derived from Dunbrack's statistics. \\ \hline 
  fa\_elec & Coulombic electrostatic potential with a distance-dependent dielectric.  Supports canonical and noncanonical residue types. \\ \hline 
  fa\_intra\_rep & Lennard-Jones repulsive between atoms in the same residue. \\ \hline 
  fa\_intra\_sol\_xover4 & Intra-residue LK solvation, counted for the atom-pairs beyond torsion-relationship. Supports arbitrary residues types. \\ \hline 
  fa\_rep & Lennard-Jones repulsive. \\ \hline 
  fa\_sol & Lazaridis-Karplus solvation energy. \\ \hline 
  hbond\_bb\_sc & Sidechain-backbone hydrogen bond energy. \\ \hline 
  hbond\_lr\_bb & Backbone-backbone hbonds distant in primary sequence. \\ \hline 
  hbond\_sc & Sidechain-sidechain and sidechain-backbone hydrogen bond energy. \\ \hline 
  hbond\_sr\_bb & Backbone-backbone hbonds close in primary sequence. \\ \hline 
  hs\_pair & Describes packing between strands and helices. It is based on the probability that two pairs of residues (1 pair in the sheet and 1 pair in the helix) will have their current dihedral angles given the separation (in sequence and physical distance) between the helix and the strand. \\ \hline 
  lk\_ball\_wtd & Weighted sum of lk\_ball \& lk\_ball\_iso (w1*lk\_ball + w2*lk\_ball\_iso); w2 is negative so that anisotropic contribution(lk\_ball) replaces some portion of isotropic contribution (fa\_sol=lk\_ball\_iso). Supports arbitrary residue types. \\ \hline 
  n\_charged & Number of charged residues. \\ \hline 
  netcharge & The total charge. \\ \hline 
  omega & Omega angles. \\ \hline 
  one\_core\_each & The fraction of secondary structure elements (helices and strands) with one large hydrophobic residue (FILMVYW) at a position in the core layer of the protein. \\ \hline 
  p\_aa\_pp & Probability of observing an amino acid, given its phi/psi energy method declaration. \\ \hline 
  pack & Packing statistics. Calculated on whole protein. \\ \hline 
  pair & A two-body energy term for residue pair interactions (electrostatics and disulfide bonds). For each pair of residues, it is based on the probability that both of these two residues will have their specified types given their sequence separation and the physical distance between them, normalized by the product of the probabilities that each residue will have its specified type given the same information. \\ \hline 
  polar\_over\_hydrophobic & exposed\_polars\_per\_res divided by exposed\_hydrophobics\_per\_res. \\ \hline 
  pro\_close & Proline ring closure energy. \\ \hline 
  rama\_prepro & Backbone torsion preference term that takes into account of whether preceding amono acid is Proline or not. Currently supports the 20 canonical alpha-amino acids, their mirror-image D-amino acids, oligoureas, and N-methyl amino acids. Arbitrary new building-blocks can also be supported provided that an N-dimensional mainchain potential can be generated somehow. \\ \hline 
  ref & Reference energy for each amino acid. \\ \hline 
  rg & Favors compact structures and is calculated as the root mean square distance between residue centroids. \\ \hline 
  rsigma & Scores strand pairs based on the distance between them and the register of the two strands. \\ \hline 
  sheet & Favors the arrangement of individual beta strands into sheets. It is derived from the probability that a structure with a given number of beta strands will have the current number of beta sheets and lone beta strands. \\ \hline 
  ss\_contributes\_core & The fraction of secondary structure elements (helices and strands) with one large hydrophobic residue (FILMVYW) at a position in either the core or interface layer of the protein. \\ \hline 
  ss\_mis & Generate secondary structure predictions from sequence. Calculated on whole protein. \\ \hline 
  ss\_pair & Describes hydrogen bonding between beta strands. \\ \hline 
  total\_score\_per\_res & The sum of all features, averaged by residue number. \\ \hline 
  two\_core\_each & The fraction of secondary structure elements (helices and strands) with two large hydrophobic residues (FILMVYW) at positions in the core layer of the protein. \\ \hline 
  vdw & Represents only steric repulsion and not attractive van der Waals forces (those are modeled in terms rewarding compact structures, such as the rg term; local interactions are implicitly included from fragments). It is calculated over pairs of atoms only in cases where: 1. the interatomic distance is less than the sum of the atoms' van der Waals radii, and 2. the interatomic distance does not depend on the torsion angles of a single residue. \\ \hline 
  yhh\_planarity & Helps control the alcohol hydrogen in tyrosine.
\end{longtable}	
\end{document}